\documentclass{article}

\usepackage{microtype}
\usepackage{graphicx}
\usepackage{subcaption}
\usepackage{adjustbox}
\usepackage{booktabs} 

\usepackage[hyphens]{url}
\usepackage{hyperref}

\usepackage[accepted]{icml2024}

\usepackage{amsmath}
\usepackage{amssymb}
\usepackage{mathtools}
\usepackage{amsthm}
\usepackage{caption}

\usepackage[capitalize,noabbrev]{cleveref}

\theoremstyle{plain}

\theoremstyle{definition}

\theoremstyle{remark}

\usepackage[textsize=tiny]{todonotes}
\usepackage[T1]{fontenc}
\usepackage[utf8]{inputenc}
\usepackage{tikz}
\usetikzlibrary{shadows}

\newcommand*\keystroke[1]{%
  \tikz[baseline=(key.base)]
    \node[%
      draw,
      fill=white,
      drop shadow={shadow xshift=0.25ex,shadow yshift=-0.25ex,fill=black,opacity=0.75},
      rectangle,
      rounded corners=2pt,
      inner sep=1pt,
      line width=0.5pt,
      font=\scriptsize\sffamily
    ](key) {#1\strut}
  ;
}

\icmltitlerunning{Craftax: A Lightning-Fast Benchmark for Open-Ended Reinforcement Learning}

\begin{document}

\twocolumn[
\icmltitle{Craftax: A Lightning-Fast Benchmark for Open-Ended Reinforcement Learning}



\icmlsetsymbol{equal}{*}

\begin{icmlauthorlist}
\icmlauthor{Michael Matthews}{oxeng}
\icmlauthor{Michael Beukman}{oxeng}
\icmlauthor{Benjamin Ellis}{oxeng}
\icmlauthor{Mikayel Samvelyan}{ucl}
\icmlauthor{Matthew Jackson}{oxeng}
\icmlauthor{Samuel Coward}{oxeng}
\icmlauthor{Jakob Foerster}{oxeng}
\end{icmlauthorlist}

\icmlaffiliation{oxeng}{University of Oxford}
\icmlaffiliation{ucl}{University College London}

\icmlcorrespondingauthor{Michael Matthews}{michael.matthews@eng.ox.ac.uk}

\icmlkeywords{Machine Learning, ICML}

\vskip 0.3in
]



\printAffiliationsAndNotice{}  

\begin{abstract}
Benchmarks play a crucial role in the development and analysis of reinforcement learning (RL) algorithms. We identify that existing benchmarks used for research into open-ended learning fall into one of two categories.  Either they are too slow for meaningful research to be performed without enormous computational resources, like Crafter, NetHack and Minecraft, or they are not complex enough to pose a significant challenge, like Minigrid and Procgen.  
To remedy this, we first present \textit{Craftax-Classic}: a ground-up rewrite of Crafter in JAX that runs up to 250x faster than the Python-native original.  A run of PPO using 1 billion environment interactions finishes in under an hour using only a single GPU and averages 90\% of the optimal reward. 
To provide a more compelling challenge we present the main \textit{Craftax} benchmark, a significant extension of the Crafter mechanics with elements inspired from NetHack\footnote{Code provided at \url{https://github.com/MichaelTMatthews/Craftax}.}.  Solving Craftax requires deep exploration, long term planning and memory, as well as continual adaptation to novel situations as more of the world is discovered.  We show that existing methods including global and episodic exploration, as well as unsupervised environment design fail to make material progress on the benchmark. We believe that Craftax can for the first time allow researchers to experiment in a complex, open-ended environment with limited computational resources.

\end{abstract}

\section{Introduction}

\begin{figure}
    \centering
    \includegraphics[width=0.4\textwidth]{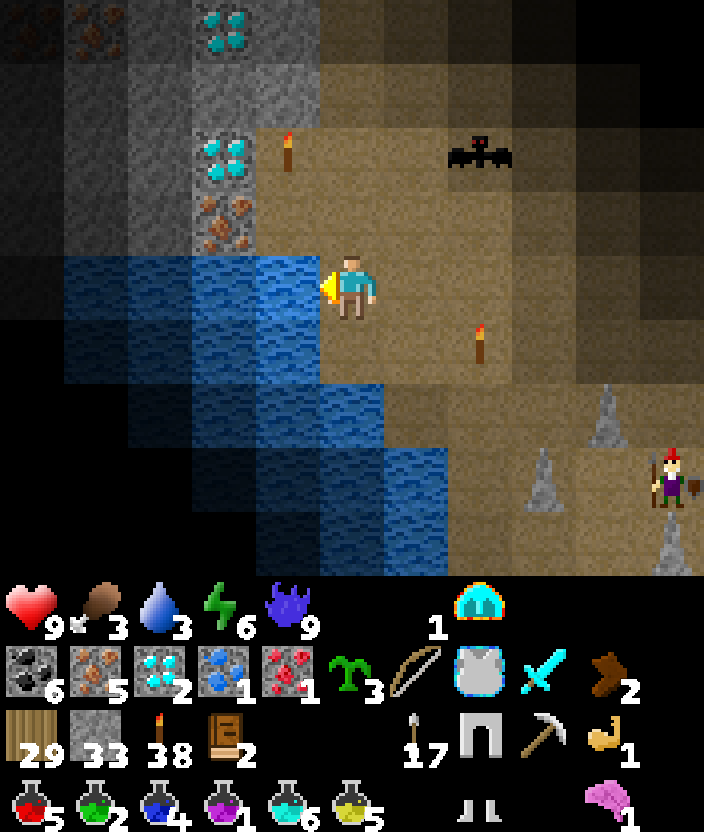}
    \caption{Pixel-based view from \texttt{Craftax}.}
    \label{fig:craftax_extended}
\end{figure}

Progress in reinforcement learning (RL) algorithms is driven in large part by the development and adoption of suitable benchmarks.  Examples include the Arcade Learning Environment \citep{bellemare2013arcade} for value based deep RL algorithms, Mujoco \citep{todorov2012mujoco} for continuous control and the StarCraft Multi-Agent Challenge \citep{samvelyan2019starcraft} for multi-agent RL.  In the effort towards increasingly general agents, there has arisen a community focused on benchmarks that exhibit more open-ended dynamics \citep{stanley2017open}, in the form of procedural world generation, skill acquisition and reuse, long term dependencies and continual learning.  This has motivated the development of environments like MALMO \citep{johnson2016malmo} (Minecraft), The NetHack Learning Environment \citep{nle2020kuttler}, MiniHack \citep{minihack2021samvelyan} and Crafter \citep{hafner2021benchmarking}.  However, slow runtime has rendered them inaccessible to current methods without large-scale computational resources, limiting their practicality to the research community.

Simultaneously, as the speed of running an end-to-end compiled RL pipeline has become fully appreciated \citep{lu2022discovered}, there has been an explosion of RL environments implemented in JAX \citep{brax2021freeman, gymnax2022github, bonnet2023jumanji, flair2023jaxmarl, nikulin2023xlandminigrid, koyamada2023pgx}.  The elimination of CPU-GPU transfer combined with efficient parallelisation and compilation means that experiments that would previously have taken a large compute cluster days to run can finish in minutes on a single GPU \citep{lu2022discovered}.

\begin{figure}[!ht]
    \centering
    \includegraphics[width=0.5\textwidth]{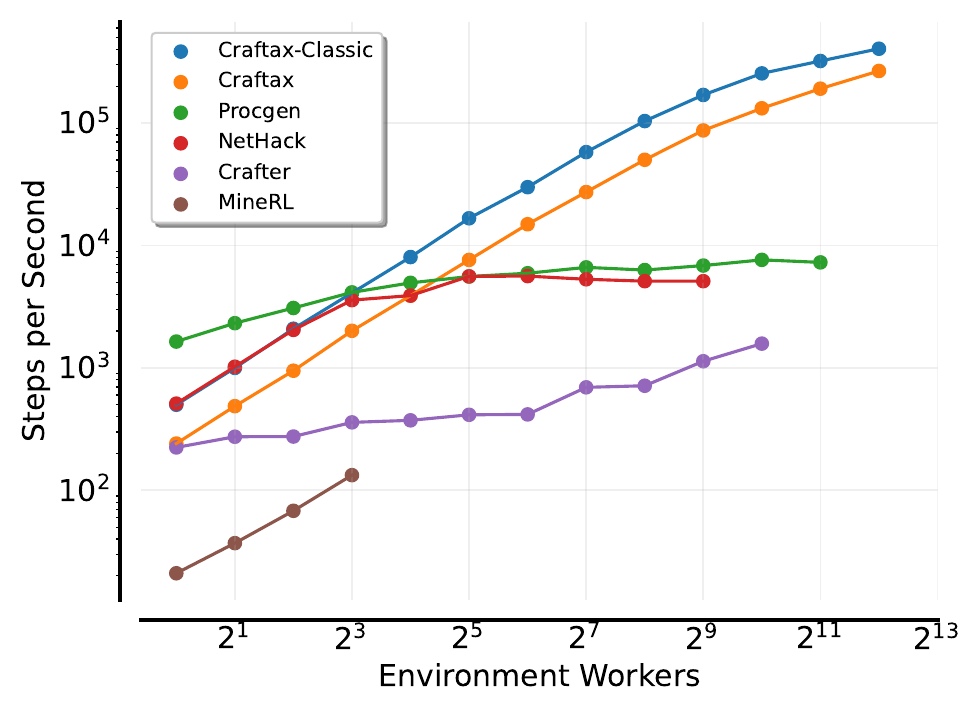}
    \caption{Speed comparison with popular benchmarks for open-ended learning.  \texttt{Craftax-Classic} and \texttt{Craftax} are 257x and 169x faster than Crafter respectively. Details of the speed test are in Appendix \ref{app:speed_comparison} and best case results are in Table \ref{tab:speed_comparison}.}
    \label{fig:speed_comparison}
\end{figure}

To bring these two paradigms together, we present the Craftax benchmark: a JAX-based environment exhibiting complex, open-ended dynamics and running orders of magnitude faster than comparable environments (Figure \ref{fig:speed_comparison}).  Concretely, we first propose \texttt{Craftax-Classic}, a reimplementation of Crafter \citep{hafner2021benchmarking} in JAX that runs 250 times faster than the Python-native original.  We then go on to show that, with easy access to drastically more timesteps, \texttt{Craftax-Classic} is solved (to within 90\% of maximum return) by a simple PPO \citep{schulman2017proximal} agent in 51 minutes (see Appendix \ref{app:classic_experiments} for further details).

We therefore also present \texttt{Craftax}, a significantly more challenging environment that incorporates dynamics inspired from NetHack \citep{nle2020kuttler} and more broadly the Roguelike\footnote{\url{https://en.wikipedia.org/wiki/Roguelike} defines Roguelikes as a class of game characterised by ``procedurally generated levels, turn-based gameplay, grid-based movement and permanent death of the player character"} genre as a whole.  Our experiments show that existing methods make little progress on \texttt{Craftax}.  We therefore hope that it presents a meaningful challenge for future RL research, while allowing experimentation with limited computational resources.

\section{Background} \label{sec:background}

\subsection{JAX-Based Environments}\label{sec:background_jax}
JAX \citep{jax2018github} is a library that allows for code written in Python to be compiled down to Accelerated Linear Algebra\footnote{\url{https://www.tensorflow.org/xla}} and run on hardware accelerators like GPUs and TPUs.  While deep RL training has traditionally been split between collecting trajectories on CPU-based environments and then training policy and value networks on the GPU, the relatively new phenomenon of JAX-based environments allows for the whole RL pipeline to be run on the GPU.  This allows for massive parallelisation of trajectory gathering (we use up to 4096 parallel environment workers), the elimination of the GPU-CPU transfer bottleneck and just-in-time (JIT) compilation of the whole training process.

\subsection{Crafter} \label{sec:background_crafter}

Crafter \citep{hafner2021benchmarking} is a top-down, procedurally generated, open-world survival game that works like a simplified, 2D Minecraft.  The player finds themselves in a world consisting of grassland, forests, lakes and mountains and must survive by gathering materials to craft tools with, maintaining hunger, thirst and energy levels, and fighting enemies.

The Crafter RL environment uses pixel-based observations and has a discrete action space made up of 4 movement directions and 13 interactions (e.g. \texttt{MAKE\_STONE\_PICKAXE}, \texttt{SLEEP}, \texttt{PLACE\_FURNACE}).  It has a set of 22 unique achievements (e.g. \texttt{DEFEAT\_ZOMBIE}, \texttt{EAT\_PLANT}, \texttt{COLLECT\_DIAMOND}), each giving the agent a $+1$ reward the first time it is achieved in an episode.

Crafter was proposed to evaluate ``strong generalization, deep
exploration, and long-term reasoning" \citep{hafner2021benchmarking}.


While Crafter has become a popular benchmark, the evaluation protocol proposed allocates algorithms only 1 million environment interactions, a very limiting constraint when compared to other RL benchmarks.  For instance, the standard protocol for the Atari benchmark (a set of environments that require limited generalisation \citep{cobbe2020leveraging}) is to use 200M environment interactions per game \citep{machado2018revisiting}, while The NetHack Learning Environment suggests using 1 billion \citep{nle2020kuttler}.  In practice, this has made Crafter a benchmark largely focused on testing sample efficiency (similar to other benchmarks like Atari 100k \citep{bellemare2013arcade}).  

While we reuse many of the Crafter dynamics, our aim is to provide a benchmark for investigations into open-endedness rather than sample efficiency. Open-endedness, by its very definition, should not be constrained by a fixed number of samples.  In practice we have to impose some limit, but this should be suitably high as to not impact the emergence of interesting phenomena.


\subsection{The NetHack Learning Environment} \label{sec:background_nle}

NetHack is an infamously difficult video game, adapted into an RL benchmark through The NetHack Learning Environment (NLE) \citep{nle2020kuttler} and MiniHack \citep{minihack2021samvelyan}.  The game involves the player descending through around fifty floors of dungeons, collecting items, fighting enemies and gaining experience before retrieving an amulet and finally beating the game by `ascending'.  The game is known to be incredibly hard, with it typically taking years of practice to achieve an ascension \citep{nle2020kuttler}. Current algorithmic approaches (both rule-based and deep RL) fail to complete the game \citep{nethackchallenge2022hambro}.
While NLE executes NetHack natively in C++ and is faster than Python-based Crafter, it is still significantly slower than GPU based environments (Figure \ref{fig:speed_comparison}).

\subsection{Open-Ended Learning}

Open-Ended Learning \citep{stanley2017open} refers to a broad category of problem domains and algorithms that focus on learning in a perpetual and unguided manner.  We investigate two subfields that are particularly relevant to \texttt{Craftax}.


\textbf{Exploration through Intrinsic Rewards}
Exploration is a key part of RL performance, especially in environments with sparse or deceptive rewards, for which simple methods like max-entropy RL \citep{sac2018haarn} or $\epsilon$-greedy policies are insufficient.  One method to overcome these difficulties is through the idea of an \textit{intrinsic reward} \citep{schmidhuber1991possibility, oudeyer2007intrinsic, barto2013intrinsic, bellemare2016unifying}, in addition to the extrinsic reward received from the environment.  This signal should intuitively reward the agent for visiting states or performing actions that satisfy some novelty-like heuristic \citep{lehman2008novelty}, causing the agent to actively explore the environment by pushing back the frontiers of its knowledge.

\textbf{Unsupervised Environment Design}
Unsupervised environment design (UED) is a paradigm in RL where an \textit{adversary} proposes environments configurations (referred to as \textit{levels}) for an agent to train on~\citep{dennis2020Emergent,jiang2021Replayguided,holder2022Evolving}. The adversary is rewarded for choosing levels that maximise the agent's \textit{regret}, defined as the difference in return between the current and optimal agent.  This has been empirically shown to automatically induce a curriculum of progressively harder levels that aid the performance and generalisation properties of the learned agent.  Different UED algorithms require different levels of access to the underlying environment state, ranging from simply being able to repeat seeds~\citep{jiang2021Replayguided} to directly editing the levels~\citep{holder2022Evolving}.  Due to the functional nature of \texttt{Craftax} necessitated by JAX, the entire environment state is exposed as a single object, making UED methods easy to apply.



\section{Craftax}
\subsection{Craftax-Classic: A Reimplementation of Crafter in JAX} \label{sec:craftax_classic}

We first present \texttt{Craftax-Classic}, a JAX-based reimplementation of the Crafter benchmark.  The aim of \texttt{Craftax-Classic} is to mimic the Crafter dynamics as closely as possible, while allowing orders of magnitude faster RL training (Figure \ref{fig:speed_comparison}).  While most of the dynamics of Crafter were remade exactly in line with the original, some subtle changes were made for performance reasons (See Appendix \ref{app:crafter_diff} for a discussion of all the differences).  Our hope is that \texttt{Craftax-Classic} provides an easy introduction to \texttt{Craftax} for those already familiar with the Crafter benchmark.

While Crafter uses pixel-based observations, many of the aspects that Crafter tests (exploration, memory) are indifferent to the exact nature of the observation and the use of pixels simply adds an extra layer of representation learning to the problem.  For this reason, we provide versions of \texttt{Craftax} with both symbolic and pixel-based observations, with the former running around 10x faster than the latter.

\subsection{Craftax: An Extension of Crafter with NetHack-Like Mechanics in JAX}

As shown in Appendix \ref{app:classic_experiments}, \texttt{Craftax-Classic} is convincingly solved by a PPO agent running for under an hour on a single GPU.  To provide a more compelling challenge we present the main \texttt{Craftax} environment, designed to be significantly harder, while retaining a fast runtime.  \texttt{Craftax} adds a large and diverse range of new game mechanics. This section provides a brief overview, with \cref{app:craftax:details} containing more details.

\textbf{Multiple Floors}  While in Crafter the player is confined to a single 64x64 grid, \texttt{Craftax} contains 9 unique procedurally generated floors, including caves, dungeons, fire and ice floors and a final boss floor.  The player can descend and ascend through the world by finding the ladders that connect adjacent floors.  Each floor contains distinct challenges in the forms of different terrain generation, enemies and required skills, necessitating deep exploration and generalisation.  
While each floor is unique, many game mechanics are shared between them and, on a meta-level, exploration strategies that worked on earlier floors (for instance moving adjacent to a block and trying different actions to figure out its characteristics) will also work on later floors.  In this way we hope to not only facilitate generalisation across different procedurally generated worlds but also generalisation of the exploration strategy through time over the learning process.


{\centering
\includegraphics[width=0.11\textwidth]{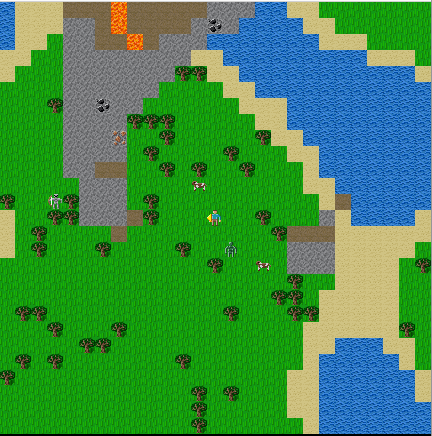}
\includegraphics[width=0.11\textwidth]{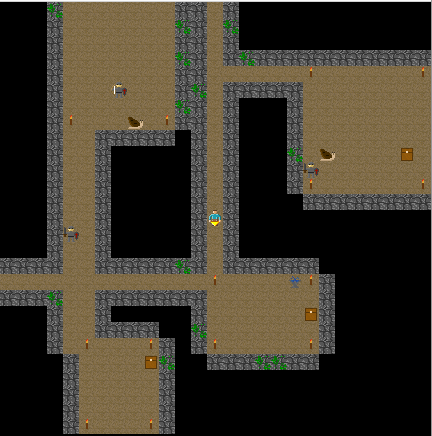}
\includegraphics[width=0.11\textwidth]{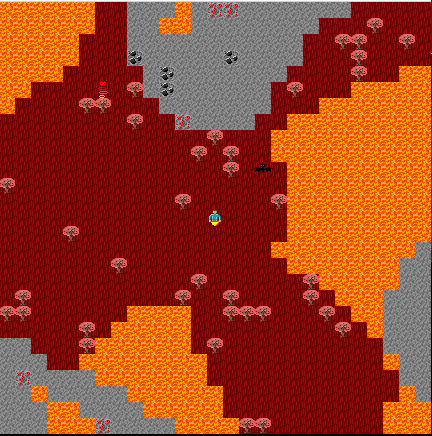}
\includegraphics[width=0.11\textwidth]{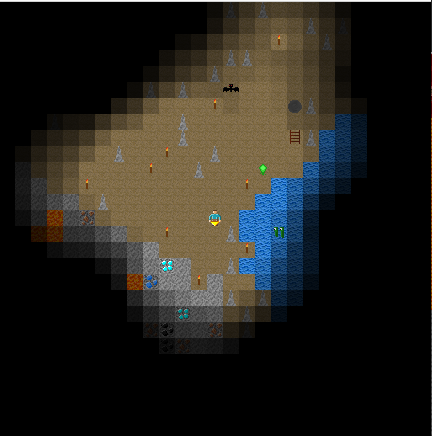}
\par}

\textbf{Combat} We overhaul the combat system with damage and defence split into physical, fire and ice categories.  As well as being able to craft higher-tier weapons, the player can use a bow (found in a random chest on the first dungeon floor) for ranged attacks, and read books (also found in dungeon chests) to learn offensive spells to cast.  The player can also craft armour to protect themselves from damage (essential when descending to the lower floors).  As well as increasing the general complexity of the environment, the diversity in combat furthers the in-context learning element provided by the procedural level generation –- an agent that stumbles upon a strong weapon or armour piece should suitably change its strategy. This further extends the exploration problem as, by design, there should not be one fixed strategy (for instance, always putting experience points into strength and defeating enemies with melee attacks) that works on every level, meaning that an agent will have to explore a diverse range of strategies to achieve consistently high return.


{\centering
\includegraphics[width=0.48\textwidth]{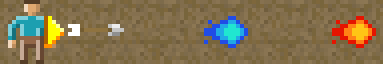}
\par}

\textbf{New Creatures} The 3 creatures present in Crafter (cow, zombie and skeleton) have been increased to 19 different creatures, each with unique effects, behaviours and techniques required to defeat them.  For instance, the Fire Elementals present on floor 7 either require an appropriate enchantment on the player's sword or magic to defeat, while beating the aquatic enemies on floor 6 typically requires ranged weaponry. This diversity of creatures further enhances the exploration problem, as no two are the same and strategies for dealing with one may entirely fail for others.


\includegraphics[width=0.05\textwidth]{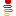}
\includegraphics[width=0.05\textwidth]{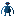}
\includegraphics[width=0.05\textwidth]{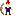}
\includegraphics[width=0.05\textwidth]{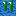}
\includegraphics[width=0.05\textwidth]{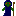}
\includegraphics[width=0.05\textwidth]{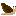}
\includegraphics[width=0.05\textwidth]{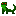}
\includegraphics[width=0.05\textwidth]{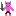}
\includegraphics[width=0.05\textwidth]{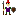}

\textbf{Potions and Enchantments} The player can find potions of varying colours spread over the 9 floors; however the effects of these potions are randomly permuted every episode. This means that an agent will need to discover which potions correspond to which effects through trial and error each episode, further testing in-context learning and memory. Also, by spending mana (magical energy) and sacrificing one of the new gemstones, the player can enchant weapons and armour. This provides them with increased damage or protection against a certain element.

{\centering
\includegraphics[width=0.05\textwidth]{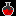}
\includegraphics[width=0.05\textwidth]{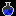}
\includegraphics[width=0.05\textwidth]{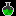}
\includegraphics[width=0.05\textwidth]{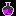}
\includegraphics[width=0.05\textwidth]{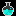}
\includegraphics[width=0.05\textwidth]{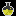}
\includegraphics[width=0.05\textwidth]{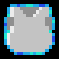}
\includegraphics[width=0.05\textwidth]{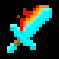}
\par}


\textbf{Attributes} The player is rewarded with an experience point every time they descend to a new floor.  These points can be spent to increase the players dexterity, strength or intelligence.  Dexterity makes the player more capable with a bow, as well as allowing them to go longer between eating, drinking and sleeping.  Strength increases the players health and melee damage, while intelligence increases the players magical powers.  The assignment of experience points can drastically alter the players abilities and have a large influence on the strategy an agent should use, testing the agents long term reasoning capabilities, as successful runs typically take tens of thousands of frames.

{\centering
\includegraphics[width=0.05\textwidth]{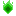}
\includegraphics[width=0.05\textwidth]{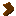}
\includegraphics[width=0.05\textwidth]{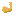}
\includegraphics[width=0.05\textwidth]{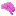}
\par}

\textbf{Boss Floor} The final floor contains a very challenging boss fight, with the player having to defeat waves of every existing enemy in the game.  As well as serving as a final challenge, the boss fight tests how the agent can transfer its prior knowledge of fighting these enemies in different contexts to the unique structure of the boss floor.

\textbf{Difficulty} \texttt{Craftax} strikes a balance of difficulty, being significantly more challenging than Crafter, but not as hard as games like NetHack.  For perspective, it took one of the authors (with extensive knowledge of the game mechanics) roughly 5 hours of gameplay to first achieve a `perfect' run where every achievement was completed.  This was playing in a GUI that allowed for unlimited time to pause and think before taking each action.  We think that \texttt{Craftax} is hard enough to present a significant challenge beyond the capabilities of existing RL methods, while being ultimately achievable without domain knowledge.

\subsection{RL Environment Interface}

\texttt{Craftax} conforms to the Gymnax \citep{gymnax2022github} wrapper for easy integration with existing frameworks.

\textbf{Observation Space} \texttt{Craftax} provides options for both pixel-based and symbolic observations.  The pixel based observation is a downscaled $63\times63\times3$ image for \texttt{Craftax-Classic} (in line with the original Crafter) and $110\times130\times3$ for \texttt{Craftax}.  The symbolic observations uses a one-hot encoding to represent both block types and creatures in the player's visual area, which is then appended to an array containing the rest of the players information (inventory, health, hunger, attributes, etc.), resulting in a flat observation space of 1345 for \texttt{Craftax-Classic} and 8268 for \texttt{Craftax}.  A full description of the observation space is given in Appendix \ref{app:obs_space}.

\textbf{Action Space} \texttt{Craftax} uses a discrete action space of size 17 and 43 for \texttt{Craftax-Classic} and \texttt{Craftax} respectively.  As in Crafter, every action can be taken at any timestep, so attempting an action without its specific prerequisites will effectively cause the agent to execute a no-op action, stepping the environment forward one timestep.  For an exhaustive list of actions see Appendix \ref{app:action_space}.

\textbf{Reward} We follow a similar reward structure to Crafter, defining a set of achievements and giving the agent a reward the first time each achievement is completed each episode.  There are 22 achievements in \texttt{Craftax-Classic} and 65 in \texttt{Craftax}.  We also penalise the agent by 0.1 for every point of damage taken and reward it by 0.1 for every point recovered.  Early experiments on \texttt{Craftax} showed that providing a flat +1 reward for each achievement, did not provide enough incentive for the agent to start exploring the dungeons, as the agent would usually be killed quickly upon descending and learn to avoid them.  To encourage progress towards harder tasks we grouped achievements into four categories: `Basic', `Intermediate', `Advanced' and `Very Advanced', each worth 1, 3, 5 and 8 reward respectively.  An exhaustive listing and description of all the achievements is located in Appendix \ref{app:achievements}.

\begin{figure}
    \centering
    \includegraphics[width=0.5\textwidth]{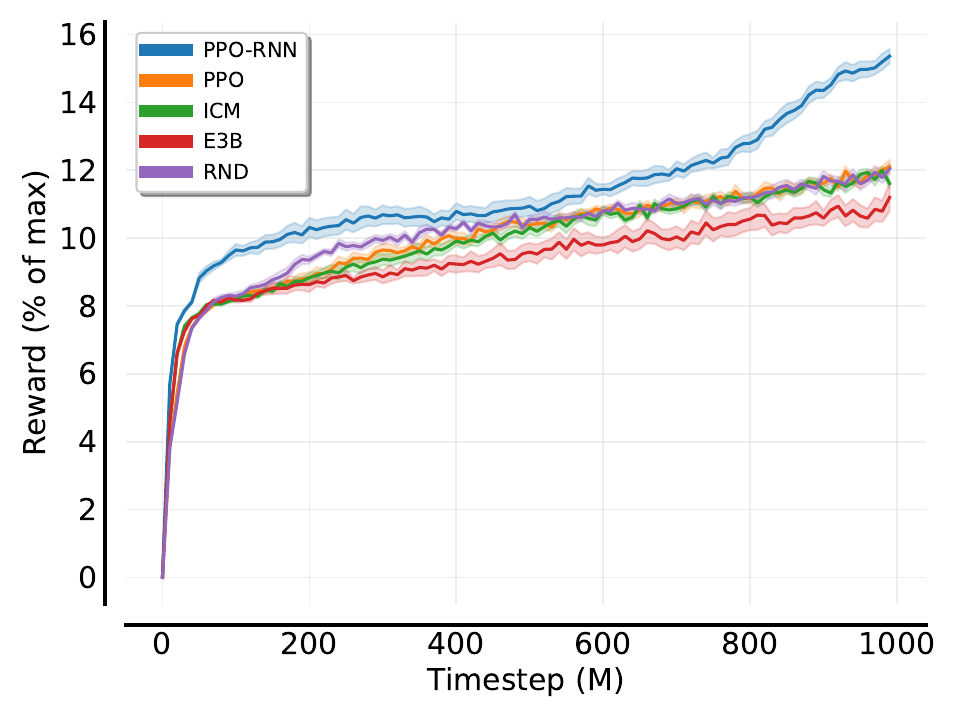}
    \caption{Rewards on Craftax-1B for PPO, PPO-RNN, ICM, E3B and RND.  Each algorithm is run for 1 billion timesteps with 10 seeds.  The shaded area denotes 1 standard error.}
    \label{fig:craftax_1b_return}
\end{figure}

\subsection{Evaluation Framework}

We propose two separate benchmarks to evaluate performance on \texttt{Craftax} (we leave \texttt{Craftax-Classic} mostly as a tool to help researchers iterate quickly on an environment very similar to the original Crafter).  We also limit the benchmarks to considering symbolic observations, as these environments are significantly faster to run, and the pixel-based environments do not contribute anything significant to the core research directions we believe \texttt{Craftax} is useful for investigating.

First, we propose the \textbf{Craftax-1B Challenge}, where a budget of 1 billion environment interactions is permitted on the \texttt{Craftax-Symbolic} environment.  This challenge is meant to provide a chance for algorithms focusing on exploration, continual learning, and long-term planning and reasoning to differentiate themselves on a hard benchmark, without being overly constrained by sample efficiency concerns.  We set the benchmark at 1 billion steps to keep it manageable for researchers with limited computational resources (a run on a single RTX 4090 can finish in an hour), while providing enough timesteps for a substantial amount of meaningful exploration of the environment.  Averaging around 5 steps per second (roughly the rate the authors moved at when playing), 1 billion timesteps corresponds to over 6 years of continual human gameplay.  We use reward as the metric to compare agents.

Secondly, we propose the \textbf{Craftax-1M Challenge}, where a budget of 1 million environment interactions is permitted on the \texttt{Craftax-Symbolic} environment.  This benchmark is meant as a test of sample efficiency, with the benefit that experiments can take only seconds to finish.  We believe the main value of this benchmark is in the almost instantaneous results researchers can obtain, allowing for iteration of methods at an unprecedented speed.  It should be noted that even with such a tight environment interaction budget, we found superior performance with massive parallelism of workers (see Table \ref{tab:craftax_1m_hyp_ppo} in Appendix \ref{app:hyp}).

\section{Experiments}

\subsection{Exploration Baselines}

We use the PPO implementations from PureJaxRL \citep{lu2022discovered} as the foundation for our baselines, allowing us to run fully compiled end-to-end RL pipelines.  We flatten the symbolic observations and use a 4 layer MLP of width 512 for both policy and value networks.  We experimented initially with using convolutions for the map view part of the observation, but consistently found that fully connected networks performed better.

We run PPO with an MLP as well as PPO with memory (using a Gated Recurrent Unit \citep{chung2014empirical}), which we refer to as PPO-RNN.  We also experiment using two state of the art intrinsic rewards for additional exploration:

\textbf{Random Network Distillation} We use the popular exploration baseline Random Network Distillation (RND) \citep{rnd2018burda} as our first method for global intrinsic reward.  RND works by training a network that is given observations to match a network of the same architecture that was randomly initialised.  Intuitively, the error of this network will be high on unseen observations, as the network will not yet have been trained to match the random output.  This error is then used as an additional intrinsic reward that incentivises the agent to explore states with novel observations.

\textbf{Intrinsic Curiosity Module} Our second baseline is the Intrinsic Curiosity Module (ICM) \citep{intrinsic2017sukhbaatar}. ICM learns a world model and provides intrinsic reward proportional to the error of this world model, incentivising the agent to visit areas of the environment which are badly modelled and therefore probably less visited.  The latent space of this world model is constructed in such a way as to ignore aleatoric uncertainty in the environment, in order to focus exploration and  avoid the so called `Noisy TV problem' \citep{rnd2018burda}.

\textbf{Exploration via Elliptical Episodic Bonuses} Our third and final exploration baseline is the E3B algorithm \citep{henaff2022exploration}.  Rather than considering exploration over the entire training process, E3B formulates its intrinsic reward to encourage exploration within each episode.  An ellipse is fitted around the agents current trajectory in latent space and it is rewarded for moving far away from this ellipse.  This approach has been shown to outperform global exploration methods, especially in non-singleton (i.e. procedurally generated) environments \citep{henaff2022exploration, henaff2023study}.

\subsection{UED Baselines}

We assess $\text{PLR}$ and ACCEL and compare them to Domain Randomisation~\citep[DR]{jakobi1997Evolutionary,tobin2017Domain}, which corresponds to uniformly sampling randomly generated levels, as per usual.  We use JaxUED \citep{coward2024JaxUED} as the base implementation for all our UED experiments.

\paragraph{$\text{PLR}$}
Prioritised level replay (PLR) \citep{jiang2021Replayguided,jiang2020Prioritized} is a curation-based UED method which maintains a buffer of training levels.
At every step, one of the two options is randomly chosen: (a) the agent is evaluated on randomly generated levels, and those with high regret are added to the buffer; (b) the agent is trained on a sample of high-regret levels from the buffer.
PLR performs gradient updates on the random levels, whereas robust PLR (denoted as $\text{PLR}^\perp$) only updates the agent on the levels sampled from the buffer. This is theoretically justified and (in some domains) empirically results in higher performance~\citep{jiang2021Replayguided}.

\paragraph{ACCEL}
ACCEL~\citep{holder2022Evolving} is similar to PLR, but also considers randomly mutated levels from the buffer as candidate levels.  ACCEL therefore increases the complexity of levels over time via iterative editing instead of having to rely solely on random generation and curation.

In this section, we provide results for the two proposed benchmarks using the previously discussed algorithms.  We use JaxUED~\citep{coward2024JaxUED} for all UED experiments, and hyperparameter tuning is detailed in Appendix \ref{app:hyp}.

\subsection{Craftax-1B} \label{sec:craftax_1b}

\begin{figure}
    \centering
    \includegraphics[width=1\linewidth]{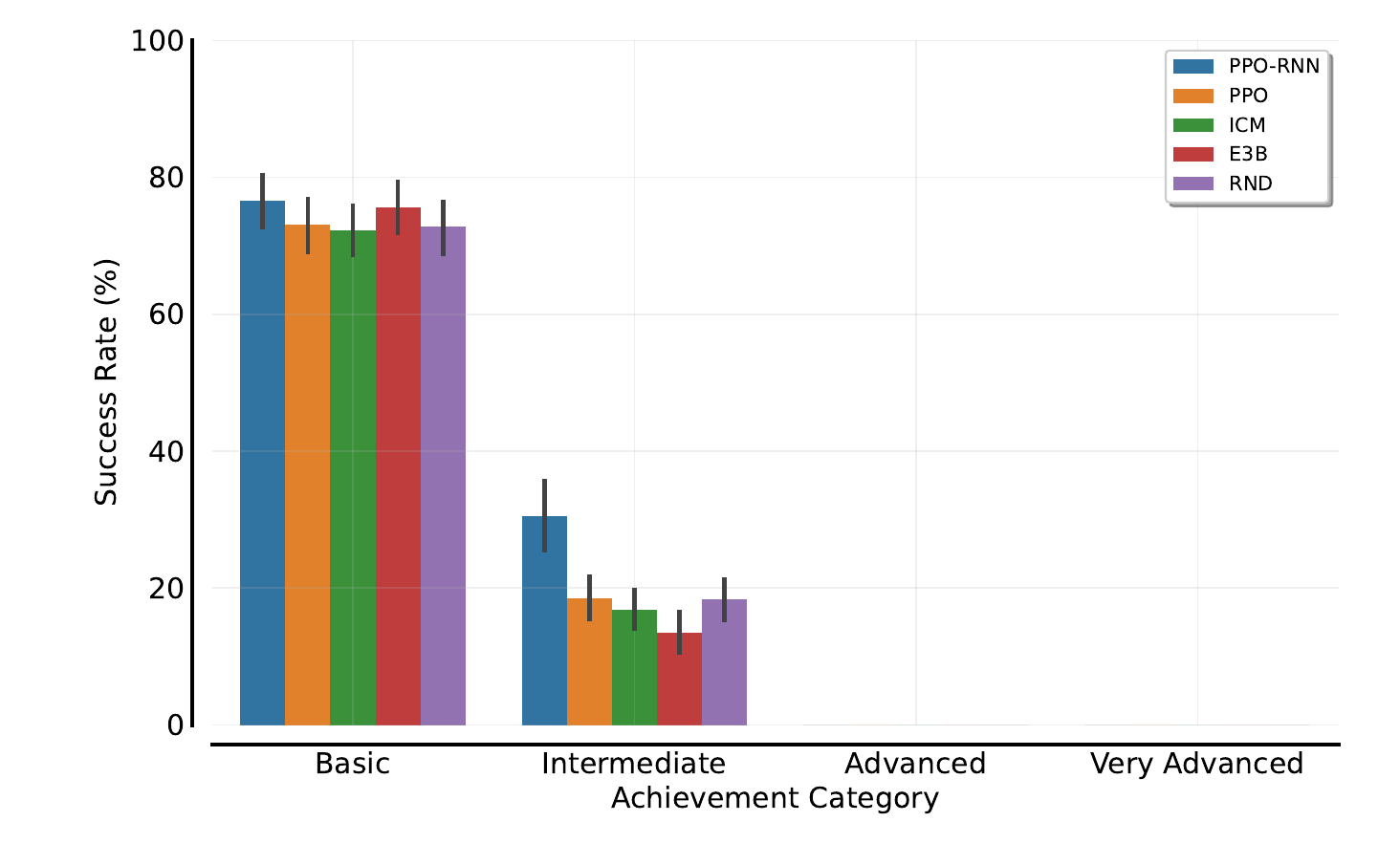}
    \caption{Achievement success rate on Craftax-1B split by achievement difficulty.  Each algorithm is run on 10 seeds, with error bars showing 1 standard error.}
    \label{fig:achievements_1b}
\end{figure}

\begin{figure}
    \centering
    \includegraphics[width=1\linewidth]{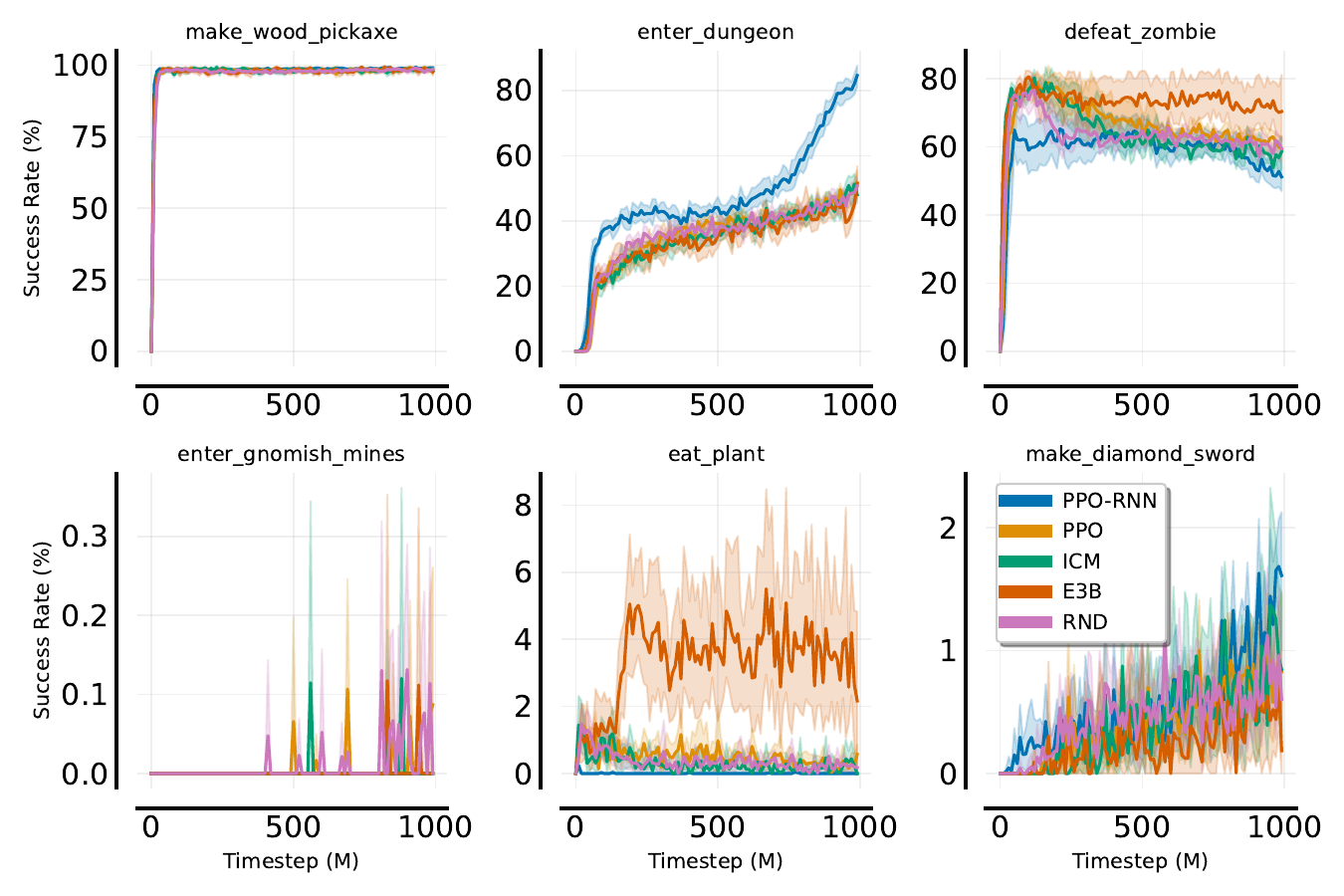}
    \caption{Achievement success rate on Craftax-1B for selected achievements over 10 seeds, with error bars denoting 1 standard error.}
    \label{fig:selected_achievements_1b}
\end{figure}

The returns for the evaluated algorithms on Craftax-1B are summarised in Figure \ref{fig:craftax_1b_return} with the final achievement yields split by difficulty shown in Figure \ref{fig:achievements_1b}.  We report reward as a percentage of the maximum achievable reward (obtained by completing every achievement) of 226 for readability.  Fine-grained achievement results are shown in Appendix \ref{app:achievement_results} with a selected set of achievements highlighted in Figure \ref{fig:selected_achievements_1b}.  We see that PPO and PPO-RNN both quickly master most of the basic achievements, with memory helping to push progress further into the intermediate category. Interestingly, none of the tested exploration methods improved performance and E3B in fact significantly reduced the reward.  We hypothesise this is because the reward structure is dense enough for RL methods without intrinsic reward to learn on, with the intrinsic reward potentially serving to distract the agent.  None of the agents make any progress in the harder achievement categories.

Taking a closer look at the achievements highlighted in Figure \ref{fig:selected_achievements_1b} we first see that all methods quickly and robustly learn basic achievements such as \texttt{MAKE\_WOOD\_PICKAXE}.  While all methods enter the dungeon (the first floor beneath the overworld) an appreciable amount, we see that PPO-RNN does so significantly more than the others, giving it more access to the high-reward achievements and driving its strong performance.  Interestingly, the success rate on some simple achievements like \texttt{DEFEAT\_ZOMBIE} decrease over time, with PPO-RNN doing the worst of all.  This is explained by the fact that zombies only appear in the overworld, so the stronger agents are trading off low-reward achievements in the overworld for high-reward ones underground.  We see that no appreciable progress is made into the second floor (the gnomish mines), with only very rare approaches being made into it.  Interestingly, we also see that the \texttt{EAT\_PLANT} (notable for being perhaps the hardest achievement in the original Crafter) is entirely ignored by PPO-RNN and is actually best achieved by E3B, indicating that while the intrinsic reward may not be helping with overall return, it does incentivise the agent to explore different parts of the environment.

Although it looks like PPO-RNN in particular is still learning, we find that running for 10 billion timesteps barely improves performance (Appendix \ref{app:craftax_10b}), indicating that this is likely an artefact of learning rate decay \citep{you2019does}.

\subsection{Craftax-1M} \label{sec:craftax_1m}

\begin{figure}
    \centering
    \includegraphics[width=0.45\textwidth]{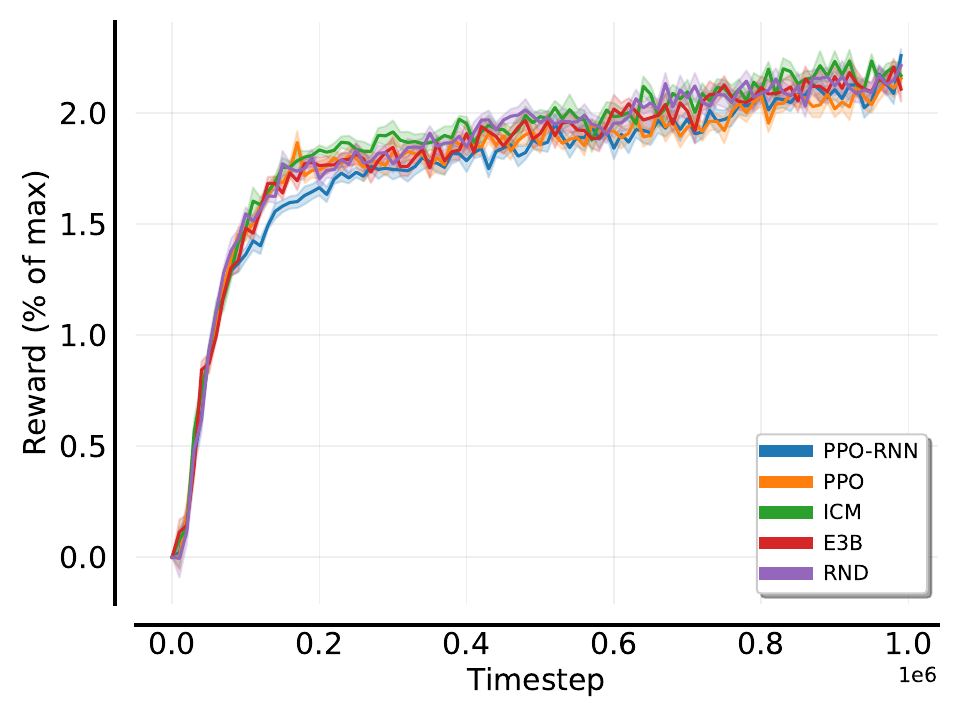}
    \caption{Rewards on Craftax-1M for PPO, PPO-RNN, ICM, E3B and RND.  Each algorithm is run for 1 million timesteps with 10 seeds.  The shaded area denotes 1 standard error.}
    \label{fig:1m_main}
\end{figure}

The results for Craftax-1M are shown in Figure \ref{fig:1m_main}.  Compared to Craftax-1B there is much less separation between the algorithms with all of them performing relatively similarly.  Fine grained achievements are shown in Appendix \ref{app:achievement_results}, showing that progress is made only on the very basic achievements (mostly corresponding to those carried over from Crafter), with agents entering the dungeons only on individual occurrences.

\subsection{Craftax-1B: UED}

We now investigate applying  DR, PLR, Robust PLR and ACCEL to Craftax-1B.  For ACCEL, we evaluate three different mutation operators applied to the overworld:

\textit{Noise} Each level $\theta$ corresponds to the angle vectors that the perlin noise algorithm uses to generate worlds. Mutation adds a uniform random number to each element.

\textit{Swap} We randomly swap two different tiles. To increase relevance, one of these is within the $16\times 16$ central area of the level---close to where the agent starts.

\textit{Restricted Swap (RSwap)} To obtain more coherent levels, we restrict the tiles that can be swapped.  For instance, stone can be swapped with any ore tile, and grass can be swapped with trees, leading to more realistic and coherent levels.


\textbf{Results}
After training for 1B timesteps, we evaluate the saved checkpoints on a fixed set of 20 evaluation levels, which were generated normally. We plot the mean and standard error over 10 independent runs.

\begin{figure}[H]
        \includegraphics[width=1\linewidth]{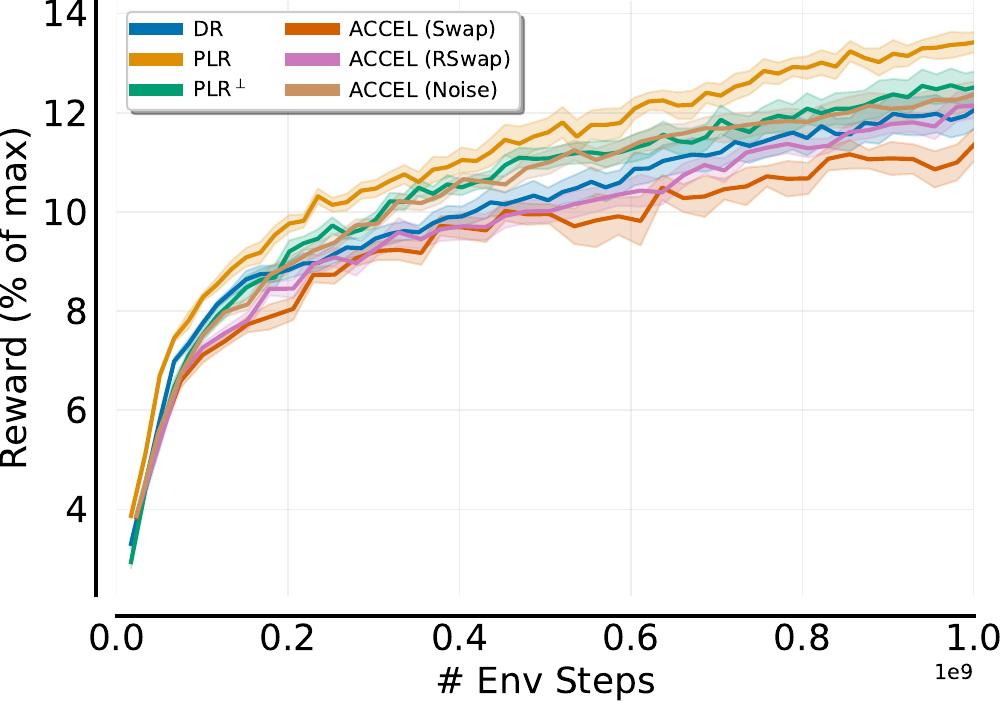}
        \caption{Evaluation reward for various UED methods on Craftax-1B on 10 seeds, with the shaded area denoting 1 standard error.}
        \label{fig:ued:return}
\end{figure}

The primary results are presented in \cref{fig:ued:return}, and success rates on all achievements are provided in \cref{fig:ued:eval:all_achievements}. We find that PLR performs best, followed by DR, $\text{PLR}^\perp$, restricted swap and noise ACCEL, which all perform roughly the same. Unrestricted swap ACCEL performs the worst.




\paragraph{PLR vs Robust PLR}
We find that Robust PLR performs worse than vanilla PLR, and only slightly outperforms DR, despite previous evidence to the contrary~\citep{jiang2021Replayguided}. 
One reason could be that, unlike the gridworld mazes it has previously been tested on, \texttt{Craftax} consistently generates high-quality and solvable levels. Therefore, the agent is not penalised by training on these DR levels. Since we compared each algorithm when using the same number of environment interactions, robust PLR performs fewer gradient updates than DR or PLR.
This result shows how important it is to expand the pool of environments UED is tested on for thorough analysis of the methods.


\paragraph{Distribution Shift}
Since UED alters the training distribution (a phenomenon named curriculum-induced covariate shift (CICS) by \citet{jiang2022Grounding}), we make the distinction between DR levels sampled from the normal generator and \textit{replay} levels sampled from the shifted UED distribution.

We find that ACCEL-based methods perform differently in replay levels compared to DR. For instance, the unrestricted swap method collect diamonds more than 30\% of the time during replay, compared to 7\% on DR levels (see \cref{fig:ued:replay_vs_dr:dr,fig:ued:replay_vs_dr:replay}). In addition, restricted swap ACCEL reaches the gnomish mines 1\% of the time in replay levels, compared to almost never in DR levels. 



\begin{figure}[H]
    \centering
    \includegraphics[width=1\linewidth]{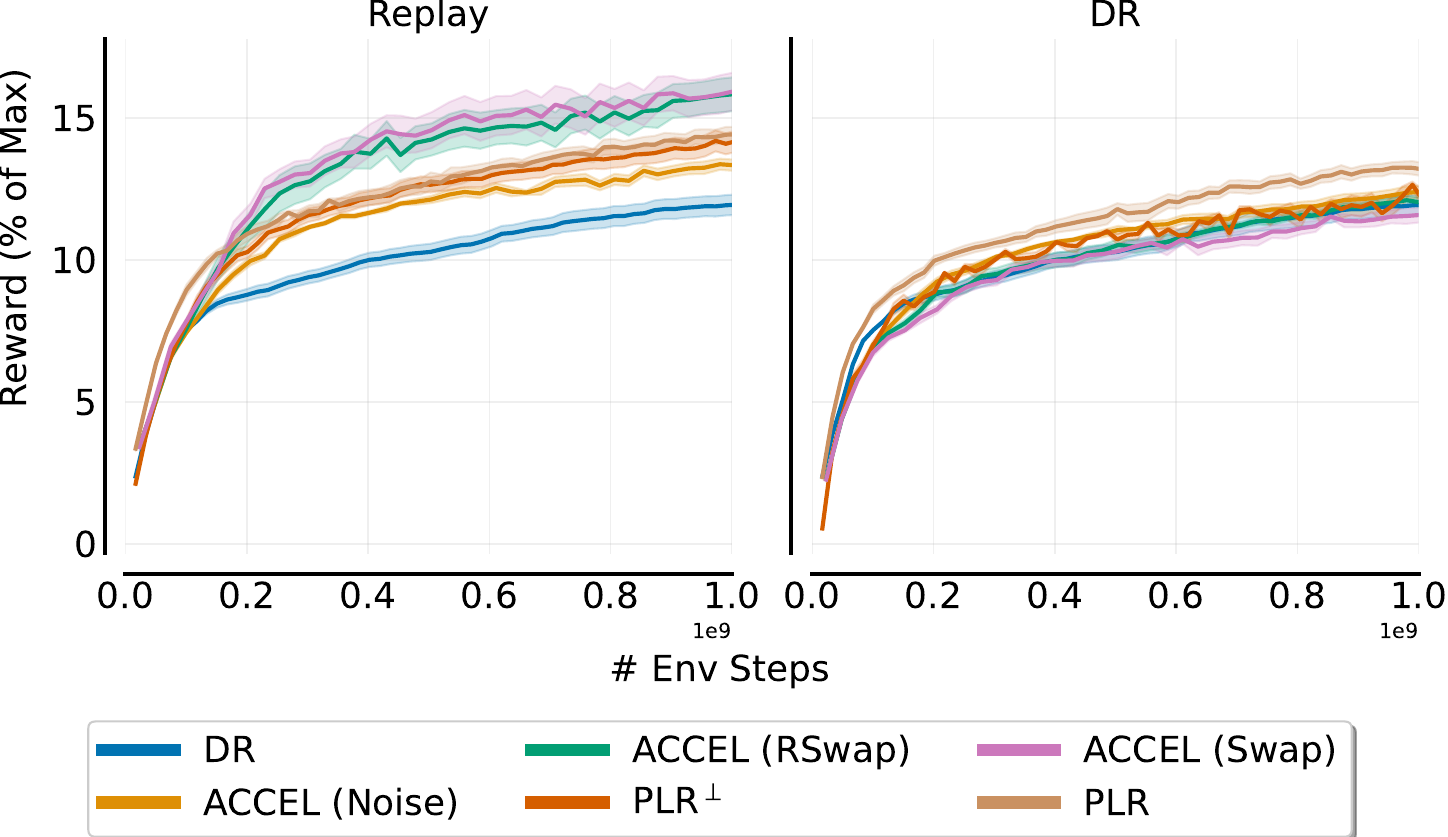}
    \caption{Average episode length for replay and DR levels on Craftax-1B over 10 seeds, with the shaded area denoting 1 standard error.}
    \label{fig:ued:rew:both}
\end{figure}

The average return (shown in \cref{fig:ued:rew:both}) is also different between replay and DR levels. The two swapping-based ACCEL methods have significantly higher returns during replay. This may indicate that the generated levels are easier (e.g., with more resources near the start), and the agent is able to perform more reward-generating activities.
The other methods have similar episode lengths during both phases.

\paragraph{UED Results Analysis}
We find that DR performs competitively to the other UED methods, with only normal PLR significantly outperforming it. We believe this is partly due to \texttt{Craftax} evaluating on in distribution levels, rather than a set of hold-out levels, where UED generally performs better~\citep{jiang2021Replayguided,holder2022Evolving}.

\section{Related Work}

\subsection{Environments for Open-Endedness}

\textbf{Exploration} The most widely known hard-exploration environments are those from the Atari benchmark with sparse rewards, most famously Montezuma's Revenge and Pitfall \citep{bellemare2013arcade}.  While these environments once posed a significant challenge to the community, they have been solved by methods like Go-Explore \citep{ecoffet2021first}, which exhaustively explore the entire game tree using state resets or goal-conditioned policies.  This technique is only possible due to these being singleton environments with mostly deterministic mechanics.  As was similarly argued in \citet{nle2020kuttler}, we believe that these environments are therefore no longer suitable for furthering research into exploration and that focus should instead be applied to more complex environments with randomised levels and stochastic mechanics.

Many recent environments have subscribed to this view and introduced elements like procedural generation and more complex world mechanics.  Procgen \citep{cobbe2020leveraging} addresses the singleton and determinism problems seen in the Arcade Learning Environment.  It randomly samples an initial `context' at the start of each episode, corresponding to concepts like obstacle placement or maze layouts.

This idea is taken even further with environments like Minecraft \citep{johnson2016malmo} which generates entirely unique (and effectively infinite) worlds each episode.  As discussed in Section \ref{sec:background}, both Crafter \citep{hafner2021benchmarking} and NetHack \citep{nle2020kuttler} make use of procedural generation to produce finite worlds each episode.  MiniHack \citep{minihack2021samvelyan} provides a set of small levels that make use of the NetHack mechanics, but are less complicated than attempting to solve the entire NetHack game.
XLand \citep{open2021oelteam} can represent a huge array of 3D games with different goals and emergent strategies.

\textbf{UED}  The most commonly used environments for UED are MiniHack~\citep{minihack2021samvelyan} and MiniGrid~\citep{MinigridMiniworld23}, BipedalWalker \citep{poet2019wang} and Car Racing \citep{jiang2021Replayguided}.  All of these environments are relatively simple and while they were vital for getting the field off the ground, for UED to be used in more advanced settings it needs to be shown to generalise to more complex environments, for which we believe \texttt{Craftax} is a suitable candidate.

\subsection{JAX-Based Environments}

Brax \citep{brax2021freeman} arguably kickstarted the development of JAX-based environments by reimplementing MuJoCo-like mechanics \citep{todorov2012mujoco} and showing that training with large numbers of environment workers could drastically reduce wall clock time without unduly affecting performance.  Gymnax \citep{gymnax2022github} implemented classic control \citep{brockman2016openai}, BSuite \citep{osband2020bsuite} and MinAtar environments \citep{young2019minatar} among others.  JaxMARL \citep{flair2023jaxmarl} remade many popular environments for multi-agent RL like StarCraft \citep{samvelyan2019starcraft, ellis2024smacv2} and Hanabi \citep{bard2020hanabi}.  Pgx \citep{koyamada2023pgx} provides implementations of classic board games like Chess and Go.  Jumanji \citep{bonnet2023jumanji} recreated many combinatorial problems like Sudoku and graph colouring.

The most similar to our own work is that of XLand-Minigrid \citep{nikulin2023xlandminigrid}, which can represent a huge diversity of tasks in a fast gridworld-based environment.  While this provides an incredible tool for open-ended research, we see \texttt{Craftax} as filling a different niche.  Specifically, \texttt{Craftax} can be seen as providing a very difficult exploration task in the form of a distribution of meaningfully similar levels.  Whereas, XLand-Mingrid encompasses a very wide range of semantically unique levels of which each one is significantly simpler than an instance of \texttt{Craftax}.  In other words, \texttt{Craftax} assesses deep exploration whereas XLand-Minigrid assesses wide generalisation.

\section{Conclusion}

We present \texttt{Craftax}, a blazing fast environment filled with complex mechanics that cannot be solved by existing RL algorithms.  We hope that \texttt{Craftax} will facilitate research into areas including exploration, continual learning, generalisation, skill acquisition and long term reasoning.  We believe an agent that could solve \texttt{Craftax} would represent a substantial step forward for the field and look forward to seeing what the community uses the benchmark to develop.

\section*{Acknowledgements}
We would like to thank Alex Goldie and Sebastian Towers for their invaluable contributions to game design, our anonymous reviewers for their useful feedback and to the open-source software community who have already contributed numerous improvements to the codebase.

\section*{Impact Statement}
This paper presents work whose goal is to advance the field of Machine Learning. There are many potential societal consequences of our work, none of which we feel must be specifically highlighted here.

\bibliography{bib,ued_bib}
\bibliographystyle{icml2024}

\newpage
\appendix
\onecolumn
\section{Differences between Craftax-Classic and Crafter} \label{app:crafter_diff}

While \texttt{Craftax-Classic} was designed to be as similar to Crafter as possible, the realities of coding in JAX necessitated some technical deviations.

\subsection{Creatures}

In Crafter, the creatures (zombies, skeletons and cows) are continually spawned around the player and despawned when the player moves far away from them.  This allows for the appearance of a world full of creatures, without having to simulate far-away entities that will have no effect on the player.

We broadly follow this system with \texttt{Craftax-Classic}, with the additional constraint that we have to pre-specify the maximum number of each creature that can exist simultaneously.  This is because JAX requires that when we compile a function, the shapes of every array in the compute graph must be known a priori, preventing the use of arbitrarily sized arrays.  This means that every timestep we have to simulate the steps for a fixed number of creatures.  It should be noted that this has the extra side effect that we actually always have to simulate the maximum number of each creature every timestep, even if less are currently spawned.  In order to allow for a dynamic number of creatures (up to the maximum) we mask out the influences of unspawned creatures.

The maximum number of creatures are 3, 3, 2 and 3 for zombies, cows, skeletons and arrows (which we treat as creatures for purpose of code reuse), respectively.  In practice it is quite rare to observe more than these amounts of each creature simultaneously in Crafter.

A similar reasoning applies to cultivating plants, where only a finite number of growing plants can be tracked.  This was set to 10 in \texttt{Craftax-Classic}.

\subsection{World Generation}

\begin{figure*}
    \centering
    \begin{subfigure}{0.3\linewidth}
        \includegraphics[width=1\linewidth]{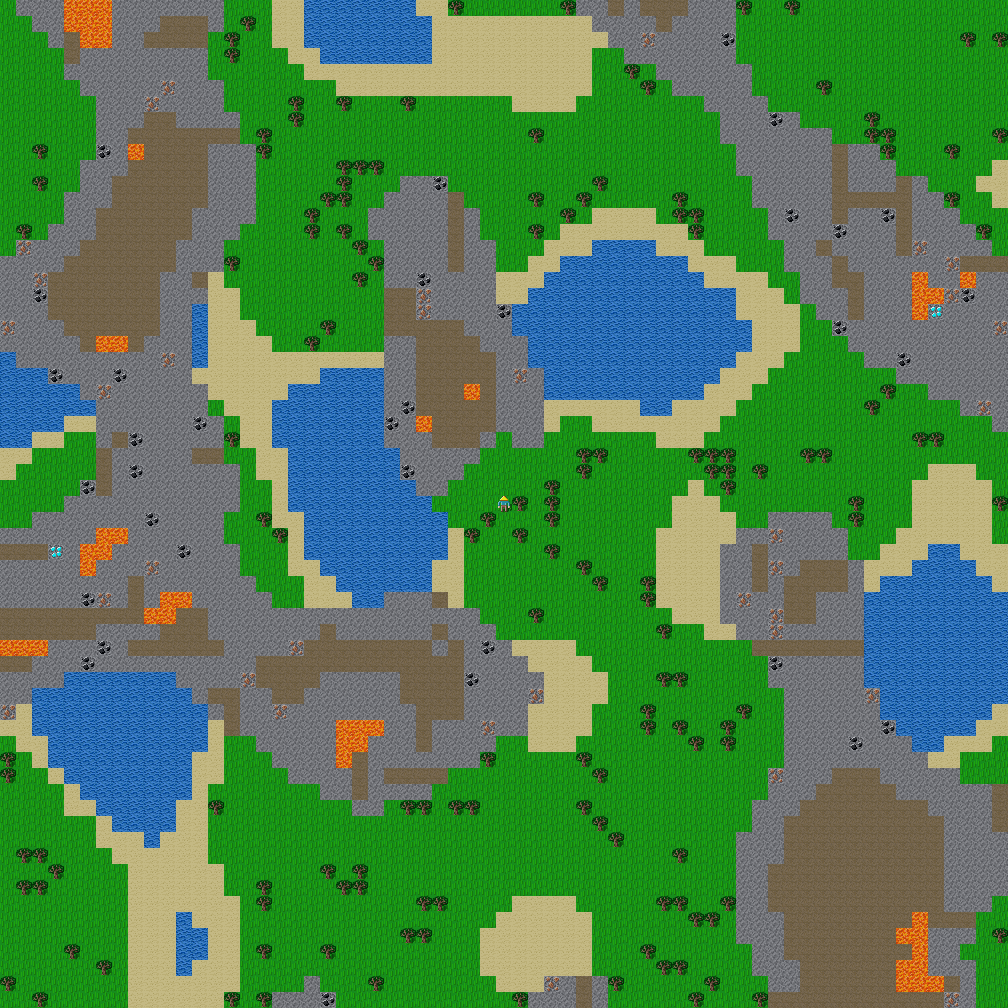}
    \end{subfigure}\hfill%
    \begin{subfigure}{0.3\linewidth}
        \includegraphics[width=1\linewidth]{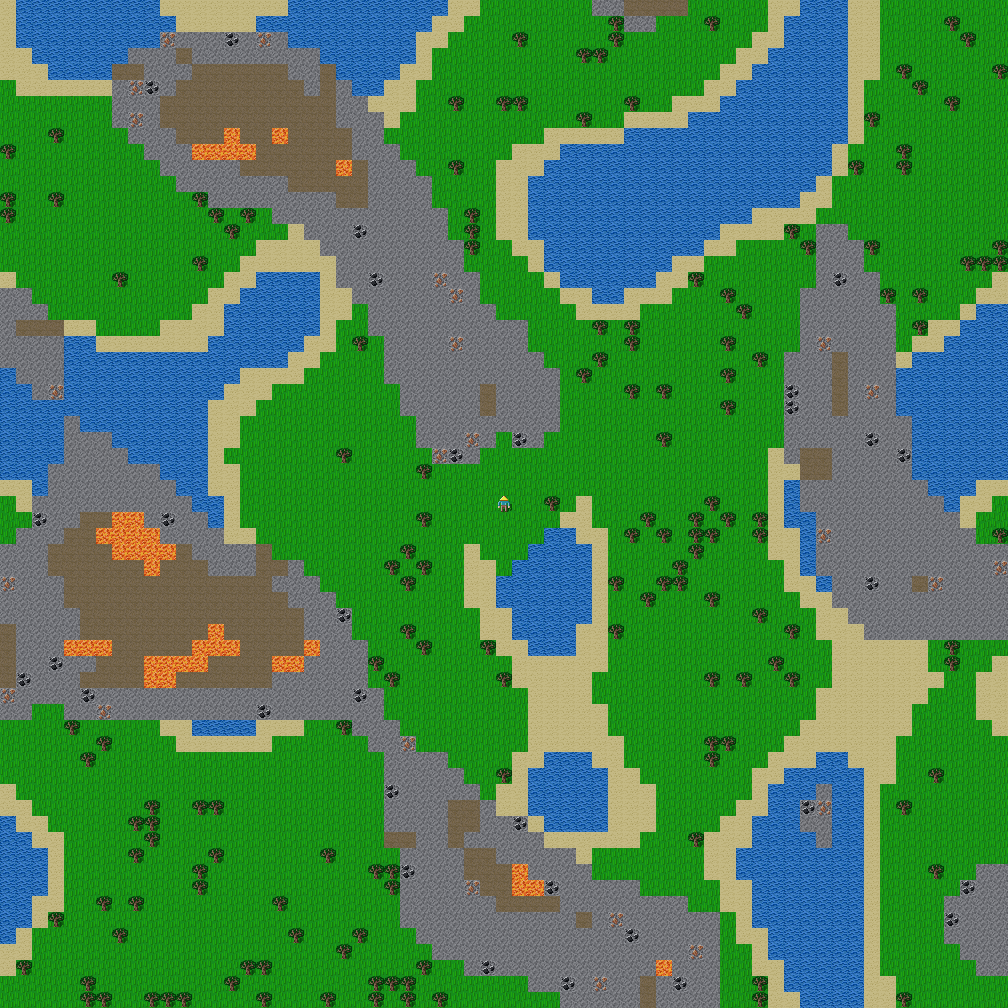}
    \end{subfigure}\hfill%
    \begin{subfigure}{0.3\linewidth}
        \includegraphics[width=1\linewidth]{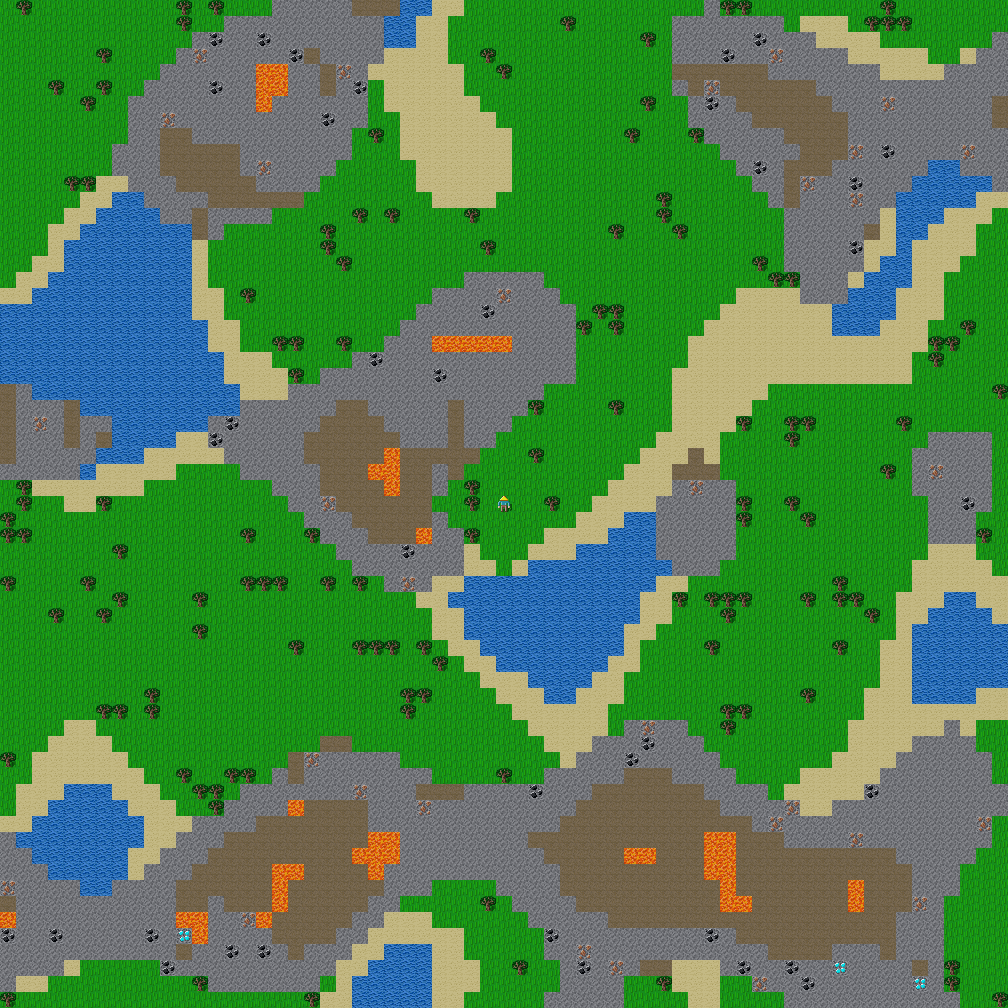}
    \end{subfigure}\hfill%
    \caption{Levels from \texttt{Craftax-Classic}}
    \label{fig:app_craftax_classic_levels}

    \vspace{1cm}
    
    \centering
    \includegraphics[width=1\linewidth]{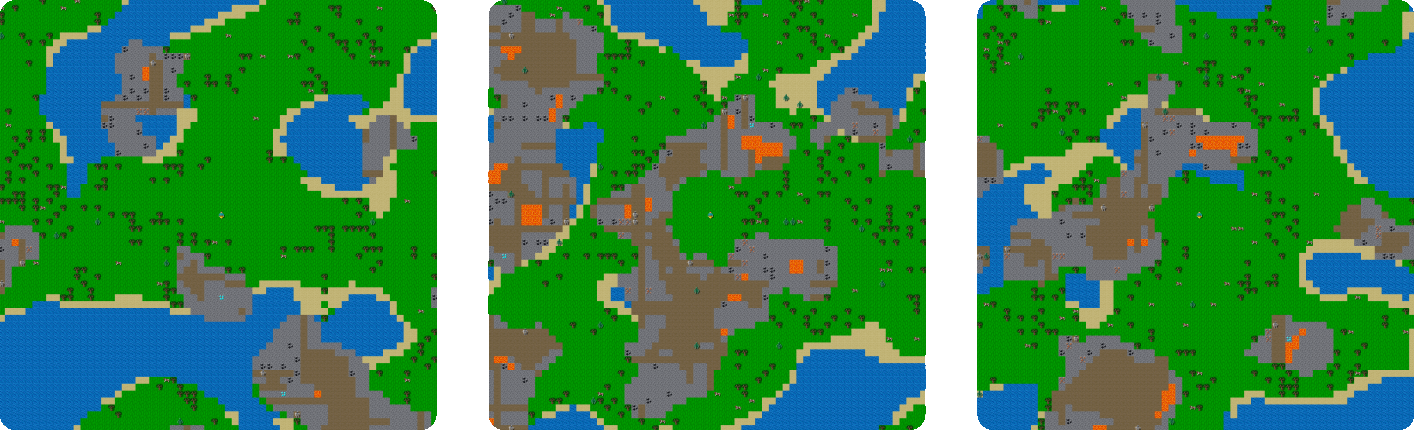}
    
    \caption{Levels from Crafter.  Figure reprinted from \citet{hafner2021benchmarking}.}
    \label{fig:app_crafter_levels}
\end{figure*}

While Crafter uses Simplex Noise as the base of its world generation, we used the very similar Perlin Noise, adapted from the Perlin-Numpy repository\footnote{\url{https://github.com/pvigier/perlin-numpy}}.  This produces qualitatively similar results to Crafter (Figures \ref{fig:app_craftax_classic_levels} and \ref{fig:app_crafter_levels}).

\begin{figure*}
\end{figure*}

\section{Speed Comparison Details} \label{app:speed_comparison}

All experiments were run on a single machine with a GeForce RTX 4090 (24GB of VRAM), i9-13900K (24 cores with 32 threads) and 32GB of RAM.  For each environment we started with a single worker and proceeded to double the number of environment workers until the machine crashed from an out-of-memory error.  We measure the average steps per second over the \textbf{entire RL learning process}, unlike other works which report times from running the environment with a static/random agent.  We believe this provides a more useful indication of the times a researcher could actually expect to see when experimenting on these benchmarks.

For \texttt{Craftax} we tested on the \texttt{Craftax-Classic-Symbolic} and \texttt{Craftax-Symbolic} environments, using an end-to-end compiled RL pipeline based off PureJaxRL \citep{lu2022discovered}.  For NetHack we used the built-in IMPALA baseline.  For Crafter and Procgen we used the stable-baselines3 \citep{sb3} PPO implementation, with the SubprocVecEnv for parallelised workers on multiple threads.  We were unable to get any baselines working on MineRL so, in order to be as generous as possible, we simply ran the environment on multiple threads with random actions.  For this reason the MineRL numbers are likely overestimates of the true expected end-to-end steps-per-second.  Since this was by far the slowest environment anyway, we didn't feel the need to explore this further.

For Craftax, Crafter, Procgen and NetHack we tried to match the magnitude of the value/policy network learning as much as possible by matching number of network parameters (different observation structures made it impossible to completely match network architectures), epochs and minibatch size.  There is no way to perfectly compare all these environments, but we felt that our decisions represent the speeds that a researcher using these environments is likely to see in practice and therefore is a fair and useful metric.

\begin{table*}[t]
\centering
\begin{tabular}{@{}l l l@{}} 
    \toprule
    \textbf{Environment} & \textbf{Steps Per Second (Best case)} & \textbf{Environment Workers (Best Case)} \\
    \midrule
    \texttt{Craftax-Classic} & 405618 & 4096 \\
    \texttt{Craftax} & 266961 & 4096 \\
    Procgen & 7638 & 1024 \\
    NetHack & 5628 & 64 \\
    Crafter & 1580 & 1024 \\
    MineRL & 133 & 8 \\
    \bottomrule
\end{tabular}
\caption{Best case speed comparison for each environment.}
\label{tab:speed_comparison}
\end{table*}

\section{Optimistic Environment Resets} \label{app:env_resets}

One of the features of compilation in JAX is that branching cannot occur inside a function that is parallelised with the \textit{vmap} operator.  In order to achieve branching-like behaviour, both branches must always be evaluated and then the appropriate output can be selected with the branching condition.  The most significant place that this occurs within JAX RL environments is with resets, where applying this constraint involves both resetting the environment and stepping it every timestep, and then selecting the desired next state based on the outputted done flag.  Indeed, this is the standard approach taken in Gymnax \citep{gymnax2022github}.  While this approach is fine when dealing with simple environments, resetting in \texttt{Craftax} is significantly more expensive as it involves generating 9 floors with multiple passes of Perlin Noise and room generation.  Furthermore, episode lengths tend to be in the hundreds of timesteps, so over 99\% of generated worlds are simply discarded.

To overcome this bottleneck, we instead perform what we term `optimistic environment resets'.  When running $N$ parallel environment workers, we only generate $M$ new worlds every timestep, where $M << N$.  Then, for the environments that do need resetting, we sample without replacement from our set of $M$ generated new states.  Typically this will make no tangible difference to the RL training process, except in the exceedingly rare case where more than $M$ environments reset simultaneously.  In this case, some environment workers will be allocated the same new state.  It should be noted that, while this will reduce the diversity of experiences seen by the agent, the stochasticity in the agent and the environment will mean the trajectories will still be different.

In practice, we use a ratio of 1 new state generated for each 16 environment workers.  In the case of 1024 environment workers, this means we generate 64 new states every timestep.  Assuming an average episode length of around 200 (in practice we usually see much longer), we can model the number of done flags as a binomial distribution $X \sim B(1024, \frac{1}{200})$.  The probability that we end up with more resets than generated levels is therefore $P(X > 64)$ which is less than $\frac{1}{10^{10}}$.  This shows that, when running for 1 billion timesteps, it is extremely unlikely we would even see a single duplicated world.

We have shown that this optimisation has essentially no effect on the RL process, while providing a significant speedup (around 2x in \texttt{Craftax}).

\section{Further Environment Details}\label{app:craftax:details}

The maximum episode length is 100,000 at which the episode is truncated.

\subsection{Block Types}

The list of block types in \texttt{Craftax} is shown in Table \ref{tab:blocks_listing}.

\begin{table*}[t]
\centering
\begin{tabular}{@{}l c c@{}} 
    \toprule
    \textbf{ID} & \textbf{Name} & \textbf{Texture}  \\
    \midrule
    2 & Grass & \adjustbox{valign=c}{\includegraphics[width=0.05\linewidth]{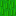}} \\[0.3cm]
    3 & Water & \adjustbox{valign=c}{\includegraphics[width=0.05\linewidth]{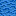}} \\[0.3cm]
    4 & Stone & \adjustbox{valign=c}{\includegraphics[width=0.05\linewidth]{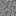}} \\[0.3cm]
    5 & Tree & \adjustbox{valign=c}{\includegraphics[width=0.05\linewidth]{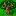}} \\[0.3cm]
    6 & Wood & \adjustbox{valign=c}{\includegraphics[width=0.05\linewidth]{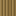}} \\[0.3cm]
    7 & Path & \adjustbox{valign=c}{\includegraphics[width=0.05\linewidth]{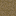}} \\[0.3cm]
    8 & Coal & \adjustbox{valign=c}{\includegraphics[width=0.05\linewidth]{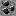}} \\[0.3cm]
    9 & Iron & \adjustbox{valign=c}{\includegraphics[width=0.05\linewidth]{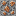}} \\[0.3cm]
    10 & Diamond & \adjustbox{valign=c}{\includegraphics[width=0.05\linewidth]{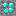}} \\[0.3cm]
    11 & Crafting Table & \adjustbox{valign=c}{\includegraphics[width=0.05\linewidth]{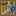}} \\[0.3cm]
    12 & Furnace & \adjustbox{valign=c}{\includegraphics[width=0.05\linewidth]{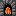}} \\[0.3cm]
    13 & Sand & \adjustbox{valign=c}{\includegraphics[width=0.05\linewidth]{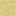}} \\[0.3cm]
    14 & Lava & \adjustbox{valign=c}{\includegraphics[width=0.05\linewidth]{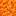}} \\[0.3cm]
    15 & Plant & \adjustbox{valign=c}{\includegraphics[width=0.05\linewidth]{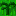}} \\[0.3cm]
    16 & Ripe Plant & \adjustbox{valign=c}{\includegraphics[width=0.05\linewidth]{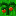}} \\[0.3cm]
    17 & Wall & \adjustbox{valign=c}{\includegraphics[width=0.05\linewidth]{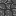}} \\[0.3cm]
    18 & Darkness & \adjustbox{valign=c}{\includegraphics[width=0.05\linewidth]{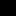}} \\[0.3cm]
    19 & Mossy Wall & \adjustbox{valign=c}{\includegraphics[width=0.05\linewidth]{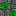}} \\[0.3cm]
    \bottomrule
\end{tabular}
\hspace{5cm}
\begin{tabular}{@{}l c c@{}} 
    \toprule
    \textbf{ID} & \textbf{Name} & \textbf{Texture}  \\
    \midrule
    20 & Stalagmite & \adjustbox{valign=c}{\includegraphics[width=0.05\linewidth]{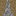}} \\[0.3cm]
    21 & Sapphire & \adjustbox{valign=c}{\includegraphics[width=0.05\linewidth]{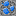}} \\[0.3cm]
    22 & Ruby & \adjustbox{valign=c}{\includegraphics[width=0.05\linewidth]{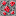}} \\[0.3cm]
    23 & Chest & \adjustbox{valign=c}{\includegraphics[width=0.05\linewidth]{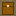}} \\[0.3cm]
    24 & Fountain & \adjustbox{valign=c}{\includegraphics[width=0.05\linewidth]{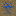}} \\[0.3cm]
    25 & Fire Grass & \adjustbox{valign=c}{\includegraphics[width=0.05\linewidth]{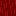}} \\[0.3cm]
    26 & Ice Grass & \adjustbox{valign=c}{\includegraphics[width=0.05\linewidth]{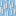}} \\[0.3cm]
    27 & Gravel & \adjustbox{valign=c}{\includegraphics[width=0.05\linewidth]{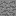}} \\[0.3cm]
    28 & Fire Tree & \adjustbox{valign=c}{\includegraphics[width=0.05\linewidth]{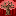}} \\[0.3cm]
    29 & Ice Shrub & \adjustbox{valign=c}{\includegraphics[width=0.05\linewidth]{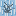}} \\[0.3cm]
    30 & Enchantment Table (Fire) & \adjustbox{valign=c}{\includegraphics[width=0.05\linewidth]{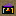}} \\[0.3cm]
    31 & Enchantment Table (Ice) & \adjustbox{valign=c}{\includegraphics[width=0.05\linewidth]{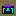}} \\[0.3cm]
    32 & Necromancer & \adjustbox{valign=c}{\includegraphics[width=0.05\linewidth]{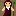}} \\[0.3cm]
    33 & Grave 1 & \adjustbox{valign=c}{\includegraphics[width=0.05\linewidth]{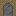}} \\[0.3cm]
    34 & Grave 2 & \adjustbox{valign=c}{\includegraphics[width=0.05\linewidth]{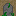}} \\[0.3cm]
    35 & Grave 3 & \adjustbox{valign=c}{\includegraphics[width=0.05\linewidth]{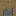}} \\[0.3cm]
    36 & Necromancer (Vulnerable) & \adjustbox{valign=c}{\includegraphics[width=0.05\linewidth]{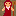}} \\
    \bottomrule
\end{tabular}
\caption{Listing of block types.  Most textures were either taken directly or adapted from those in Crafter.}
\label{tab:blocks_listing}
\end{table*}

\subsection{Creatures}

The list of creatures in \texttt{Craftax} is shown in Table \ref{tab:mob_listing}.

\begin{table*}[t]
\centering
\begin{tabular}{@{}l c c c c c c c@{}} 
    \toprule
    \textbf{Name} & \textbf{Texture} & \textbf{Type} & \textbf{Health} & \textbf{Damage} & \textbf{Defense (\%)} & \textbf{Collision Type} & \textbf{Floor} \\
    \midrule
    Zombie & \adjustbox{valign=c}{\includegraphics[width=0.05\linewidth]{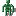}} & Melee & 5 & 2 & 0 & Ground & 0 \\[0.3cm]
    Skeleton & \adjustbox{valign=c}{\includegraphics[width=0.05\linewidth]{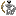}} & Ranged & 3 & 2 & 0 & Ground & 0 \\[0.3cm]
    Cow & \adjustbox{valign=c}{\includegraphics[width=0.05\linewidth]{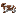}} & Passive & 3 & - & 0 & Ground & 0 \\[0.3cm]
    
    Orc Solider & \adjustbox{valign=c}{\includegraphics[width=0.05\linewidth]{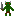}} & Melee & 7 & 3 & 0 & Ground & 1 \\[0.3cm]
    Orc Mage & \adjustbox{valign=c}{\includegraphics[width=0.05\linewidth]{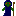}} & Ranged & 5 & 3 & 0 & Ground & 1 \\[0.3cm]
    Snail & \adjustbox{valign=c}{\includegraphics[width=0.05\linewidth]{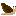}} & Passive & 6 & - & 0 & Ground & 1,3,4 \\[0.3cm]
    
    Gnome Warrior & \adjustbox{valign=c}{\includegraphics[width=0.05\linewidth]{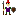}} & Melee & 9 & 4 & 0 & Ground & 2 \\[0.3cm]
    Gnome Archer & \adjustbox{valign=c}{\includegraphics[width=0.05\linewidth]{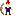}} & Ranged & 6 & 2 & 0 & Ground & 2 \\[0.3cm]
    Bat & \adjustbox{valign=c}{\includegraphics[width=0.05\linewidth]{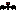}} & Passive & 4 & - & 0 & Flying & 2,5,6 \\[0.3cm]
    
    Lizard & \adjustbox{valign=c}{\includegraphics[width=0.05\linewidth]{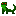}} & Melee & 11 & 5 & 0 & Amphibian & 3 \\[0.3cm]
    Kobold & \adjustbox{valign=c}{\includegraphics[width=0.05\linewidth]{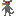}} & Ranged & 8 & 4 & 0 & Ground & 3 \\[0.3cm]
    
    Knight & \adjustbox{valign=c}{\includegraphics[width=0.05\linewidth]{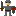}} & Melee & 12 & 6 & 50 & Ground & 4 \\[0.3cm]
    Archer & \adjustbox{valign=c}{\includegraphics[width=0.05\linewidth]{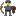}} & Ranged & 12 & 4 & 50 & Ground & 4 \\[0.3cm]
    
    Troll & \adjustbox{valign=c}{\includegraphics[width=0.05\linewidth]{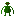}} & Melee & 20 & 6 + \textcolor{red}{1} + \textcolor{blue}{1} & 20 & Ground & 5 \\[0.3cm]
    Deep Thing & \adjustbox{valign=c}{\includegraphics[width=0.05\linewidth]{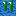}} & Ranged & 6 & 4 + \textcolor{red}{3} + \textcolor{blue}{3} & 0 & Aquatic & 5 \\[0.3cm]
    
    Pig Man & \adjustbox{valign=c}{\includegraphics[width=0.05\linewidth]{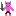}} & Melee & 20 & 3 + \textcolor{red}{5} & 90, \textcolor{red}{100} & Ground & 6 \\[0.3cm]
    Fire Elemental & \adjustbox{valign=c}{\includegraphics[width=0.05\linewidth]{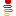}} & Ranged & 14 & 3 + \textcolor{red}{5} & 90, \textcolor{red}{100} & Flying & 6 \\[0.3cm]
    
    Frost Troll & \adjustbox{valign=c}{\includegraphics[width=0.05\linewidth]{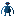}} & Melee & 24 & 4 + \textcolor{blue}{5} & 90, \textcolor{blue}{100} & Ground & 7 \\[0.3cm]
    Ice Elemental & \adjustbox{valign=c}{\includegraphics[width=0.05\linewidth]{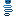}} & Ranged & 16 & 4 + \textcolor{blue}{4} & 90, \textcolor{blue}{100} & Flying & 7 \\[0.3cm]
    \bottomrule
\end{tabular}
\caption{Listing of creatures.  Some textures were taken directly or adapted from Crafter.  Text in \textcolor{red}{red} and \textcolor{blue}{blue} denote \textcolor{red}{fire} and \textcolor{blue}{ice} attack/defence respectively.}
\label{tab:mob_listing}
\end{table*}

\subsection{World Generation}

\begin{figure*}
    \centering
    \begin{subfigure}{0.28\linewidth}
        \includegraphics[width=1\linewidth]{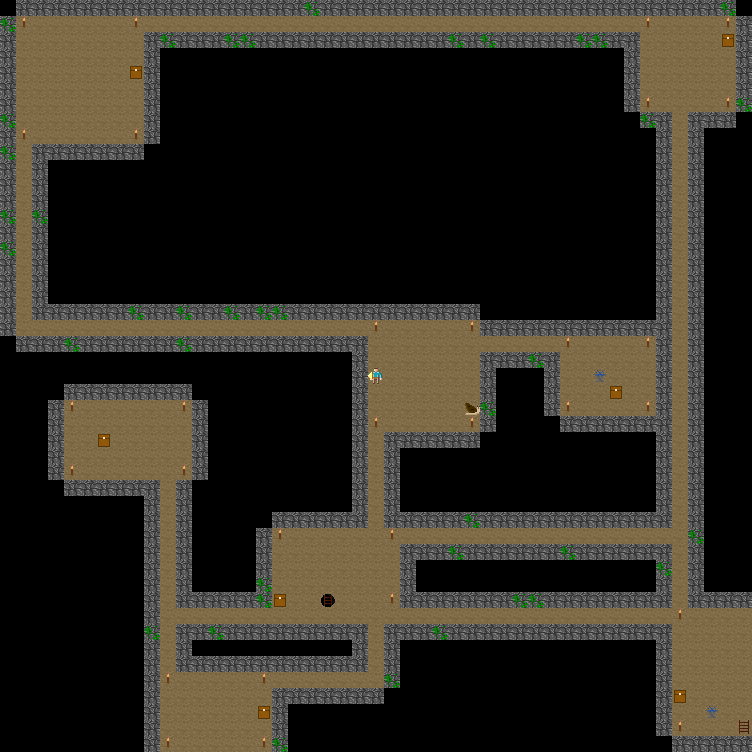}
    \end{subfigure}\hfill%
    \begin{subfigure}{0.28\linewidth}
        \includegraphics[width=1\linewidth]{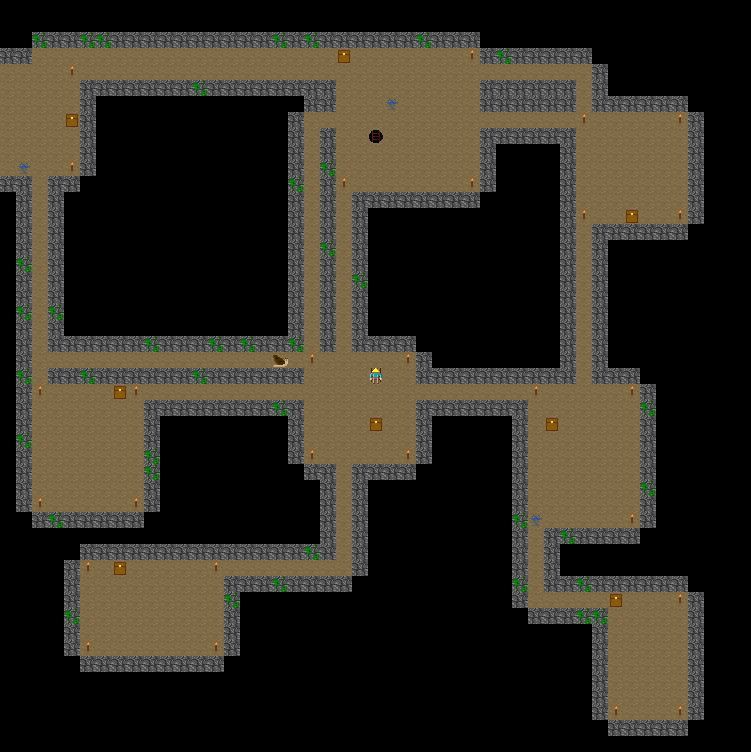}
    \end{subfigure}\hfill%
    \begin{subfigure}{0.28\linewidth}
        \includegraphics[width=1\linewidth]{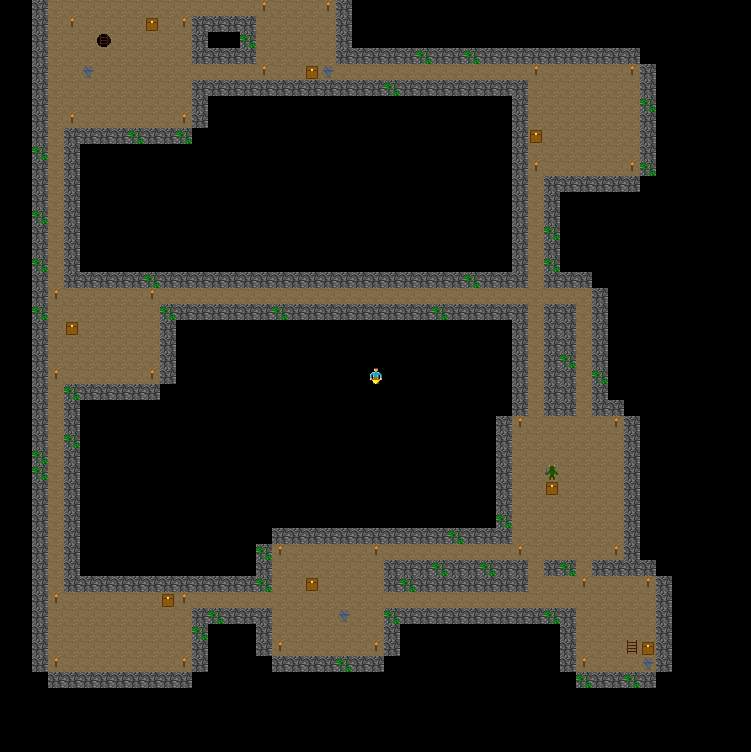}
    \end{subfigure}\hfill%
    \caption{\texttt{Craftax}: Floor 1 - Dungeons}
    \label{fig:app_craftax_classic_level_1}

    \centering
    \begin{subfigure}{0.28\linewidth}
        \includegraphics[width=1\linewidth]{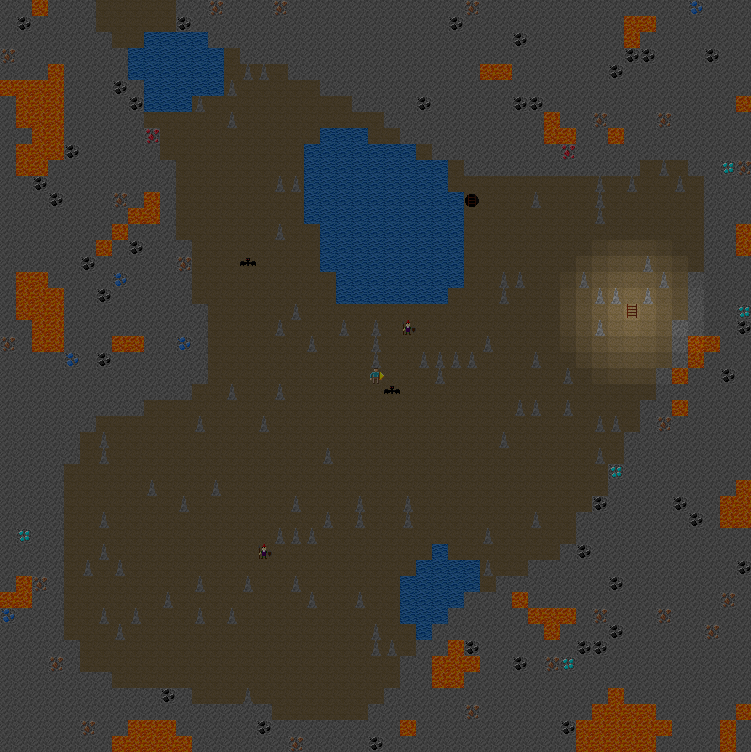}
    \end{subfigure}\hfill%
    \begin{subfigure}{0.28\linewidth}
        \includegraphics[width=1\linewidth]{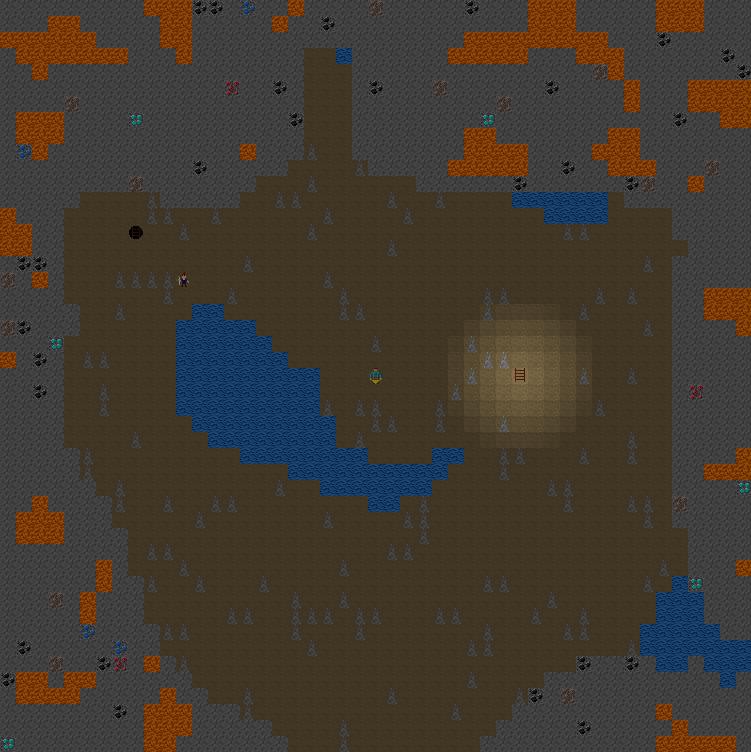}
    \end{subfigure}\hfill%
    \begin{subfigure}{0.28\linewidth}
        \includegraphics[width=1\linewidth]{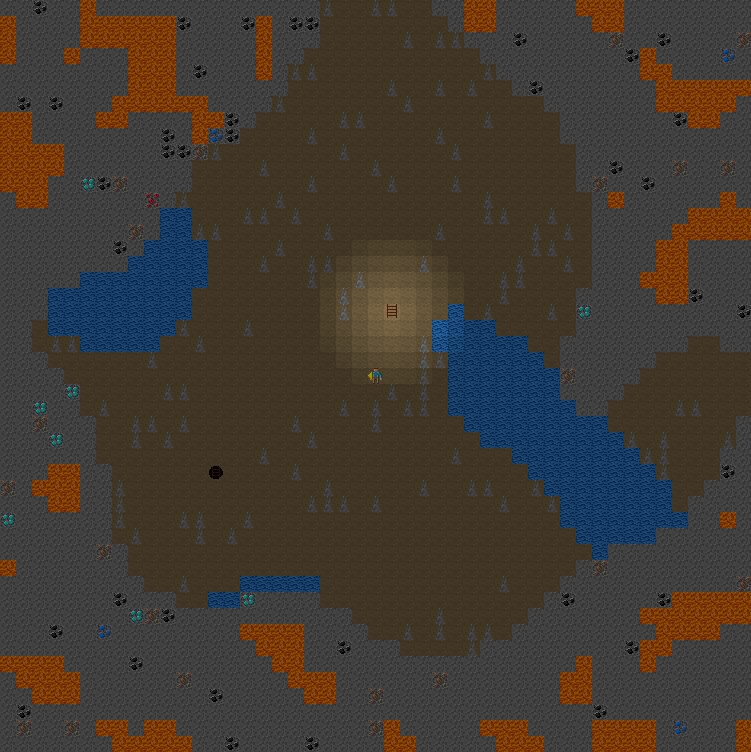}
    \end{subfigure}\hfill%
    \caption{\texttt{Craftax}: Floor 2 - Gnomish Mines}
    \label{fig:app_craftax_classic_level_2}

    \centering
    \begin{subfigure}{0.28\linewidth}
        \includegraphics[width=1\linewidth]{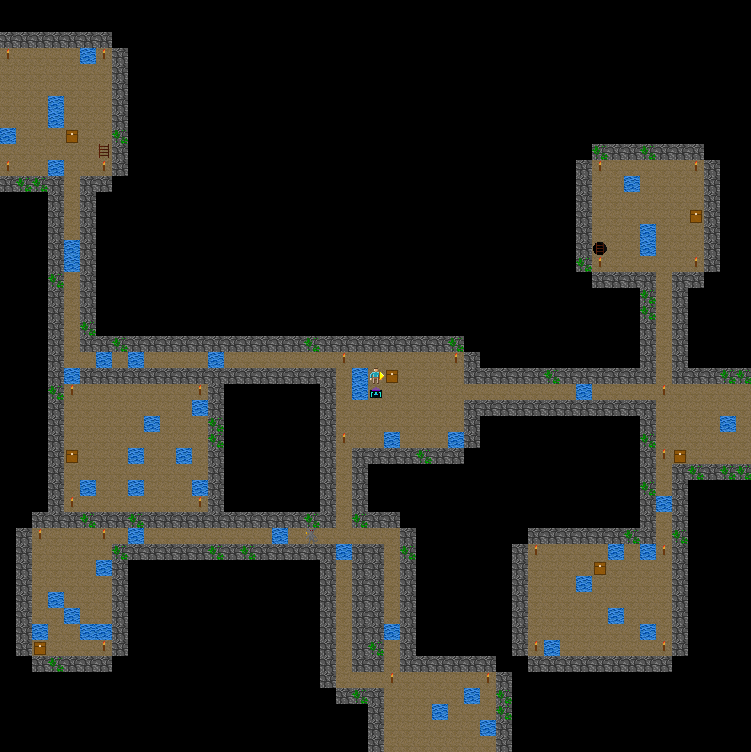}
    \end{subfigure}\hfill%
    \begin{subfigure}{0.28\linewidth}
        \includegraphics[width=1\linewidth]{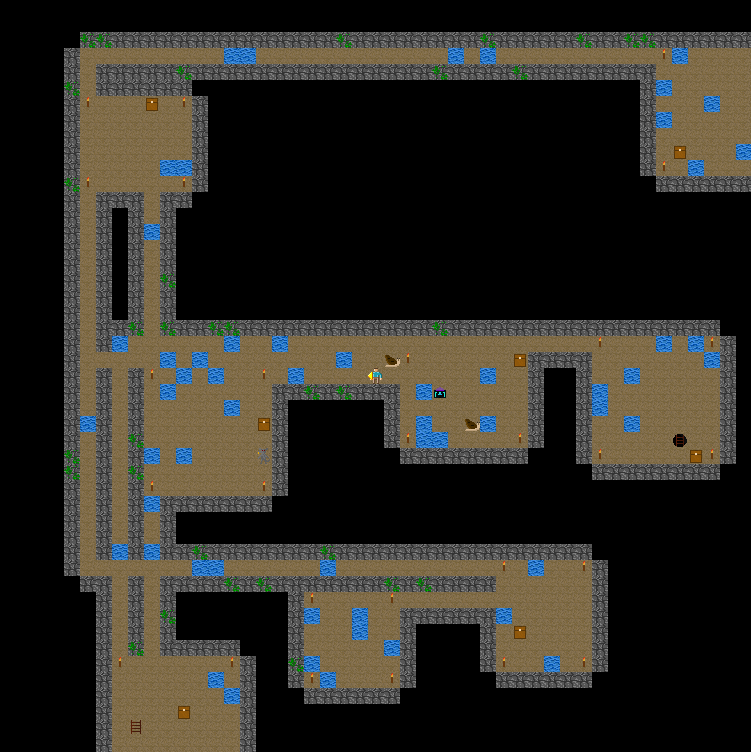}
    \end{subfigure}\hfill%
    \begin{subfigure}{0.28\linewidth}
        \includegraphics[width=1\linewidth]{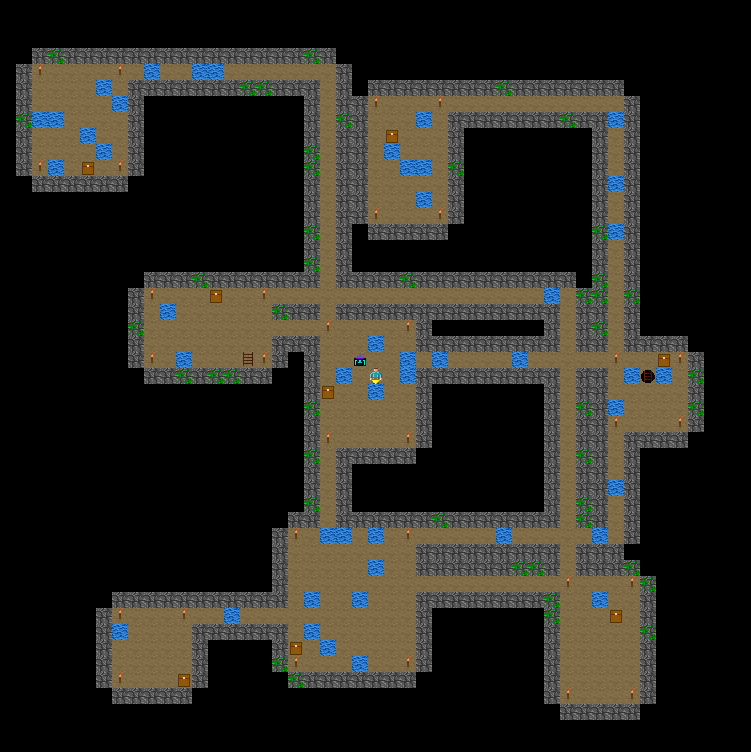}
    \end{subfigure}\hfill%
    \caption{\texttt{Craftax}: Floor 3 - Sewers}
    \label{fig:app_craftax_classic_level_3}

    \centering
    \begin{subfigure}{0.28\linewidth}
        \includegraphics[width=1\linewidth]{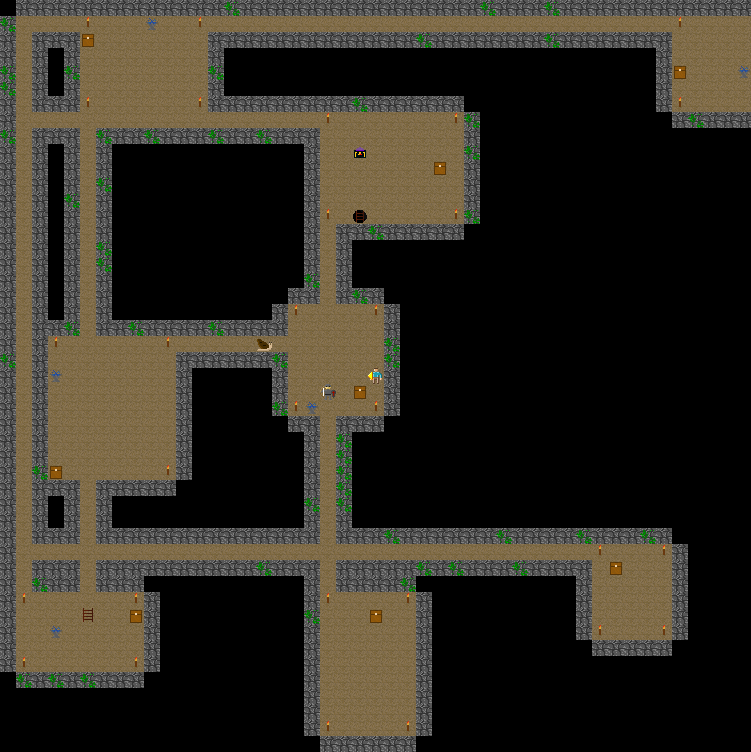}
    \end{subfigure}\hfill%
    \begin{subfigure}{0.28\linewidth}
        \includegraphics[width=1\linewidth]{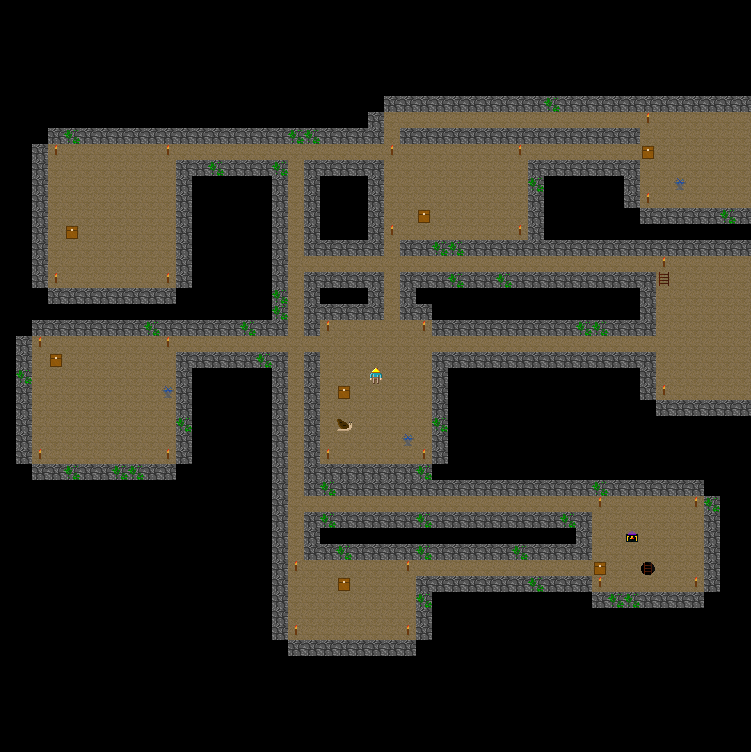}
    \end{subfigure}\hfill%
    \begin{subfigure}{0.28\linewidth}
        \includegraphics[width=1\linewidth]{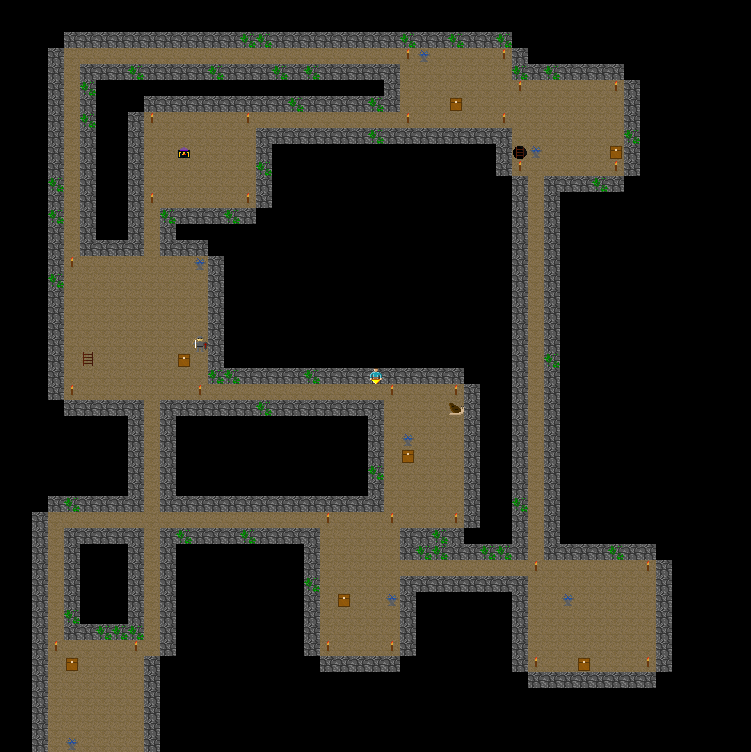}
    \end{subfigure}\hfill%
    \caption{\texttt{Craftax}: Floor 4 - Vaults}
    \label{fig:app_craftax_classic_level_4}

\end{figure*}

\begin{figure*}
    \centering
    \begin{subfigure}{0.28\linewidth}
        \includegraphics[width=1\linewidth]{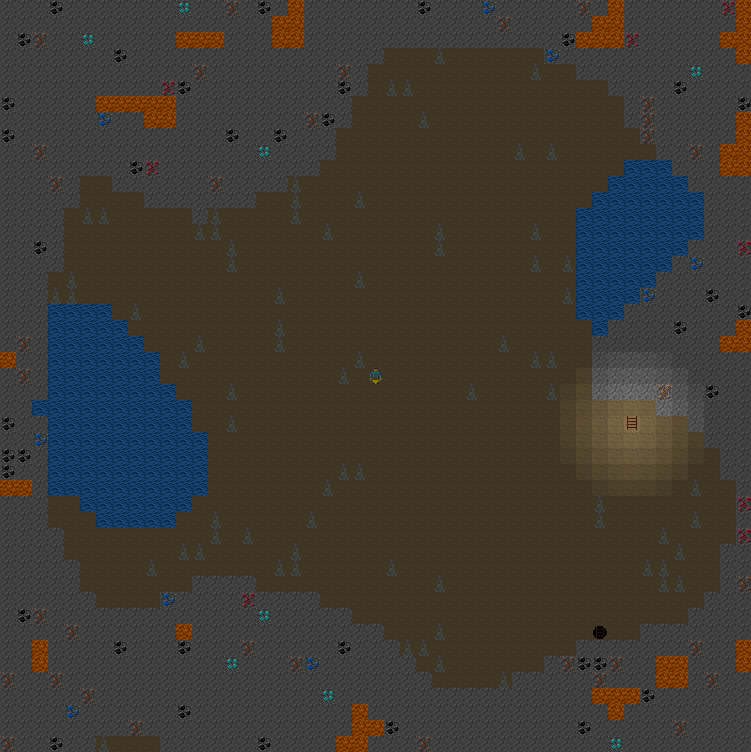}
    \end{subfigure}\hfill%
    \begin{subfigure}{0.28\linewidth}
        \includegraphics[width=1\linewidth]{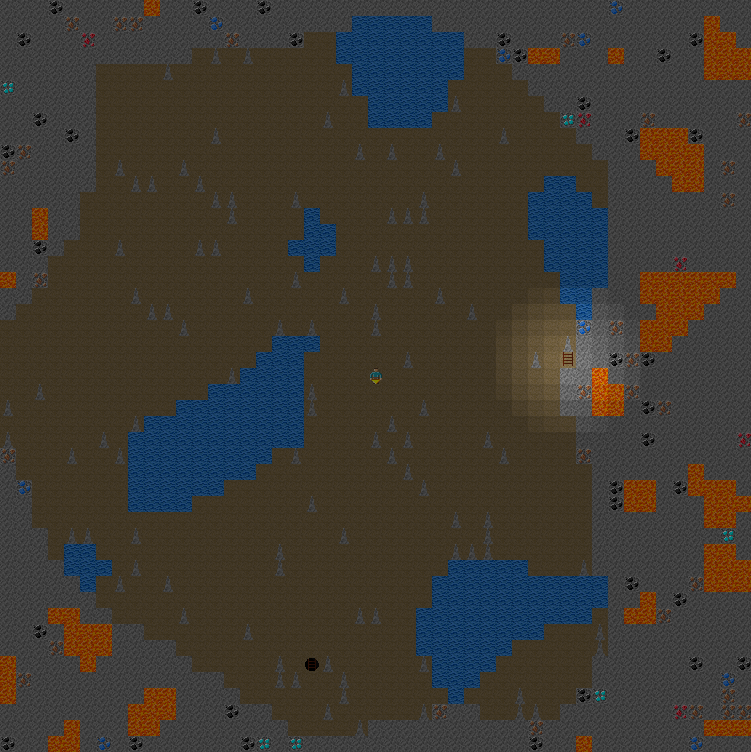}
    \end{subfigure}\hfill%
    \begin{subfigure}{0.28\linewidth}
        \includegraphics[width=1\linewidth]{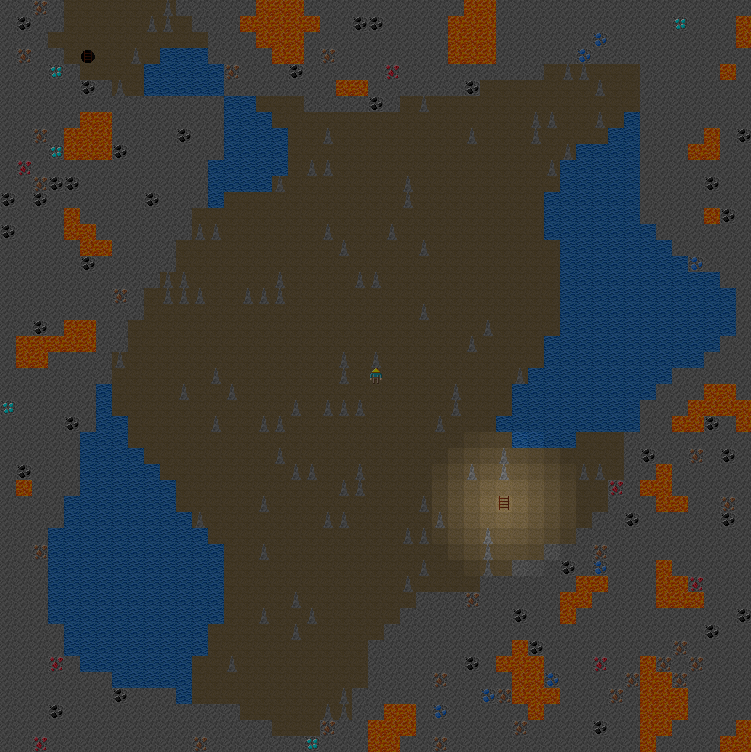}
    \end{subfigure}\hfill%
    \caption{\texttt{Craftax}: Floor 5 - Troll Mines}
    \label{fig:app_craftax_classic_level_5}

    \centering
    \begin{subfigure}{0.28\linewidth}
        \includegraphics[width=1\linewidth]{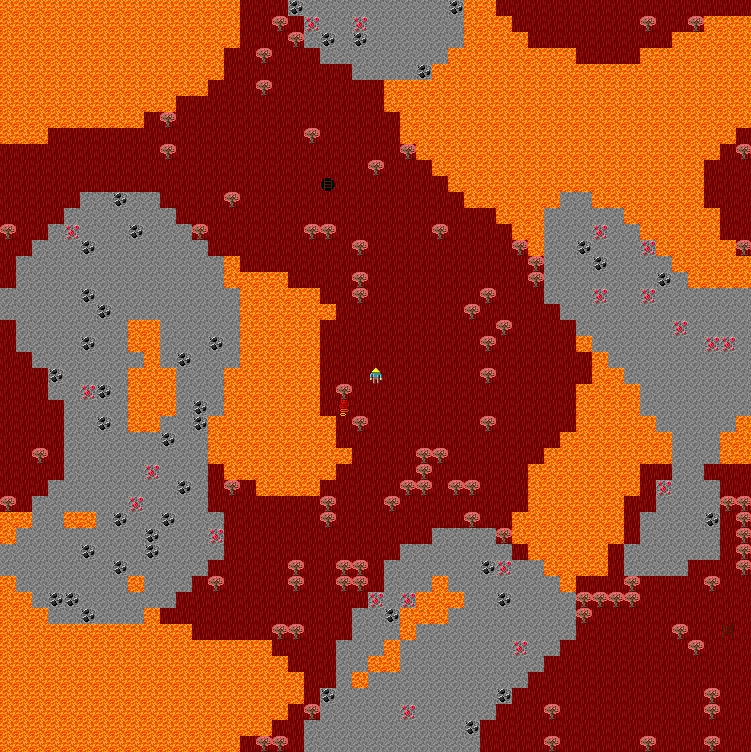}
    \end{subfigure}\hfill%
    \begin{subfigure}{0.28\linewidth}
        \includegraphics[width=1\linewidth]{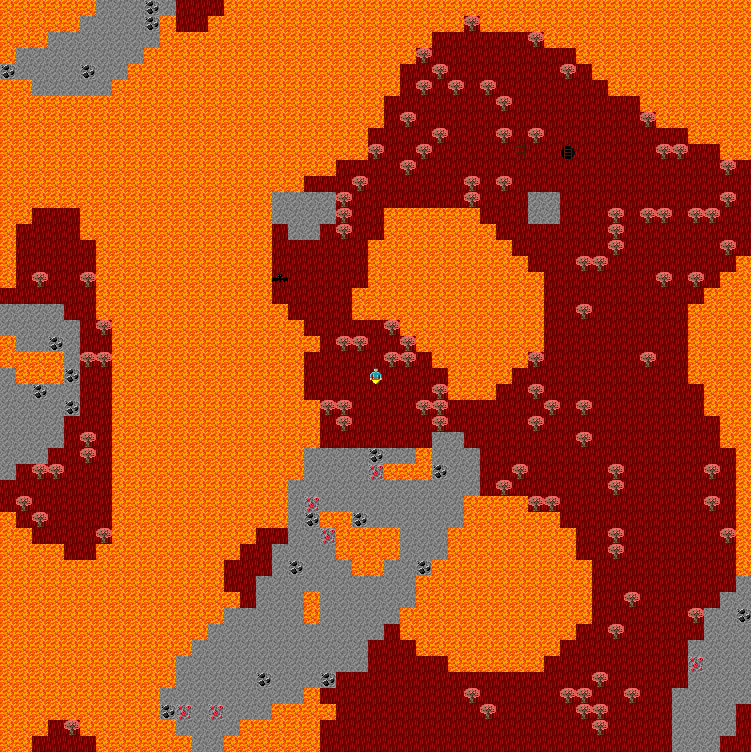}
    \end{subfigure}\hfill%
    \begin{subfigure}{0.28\linewidth}
        \includegraphics[width=1\linewidth]{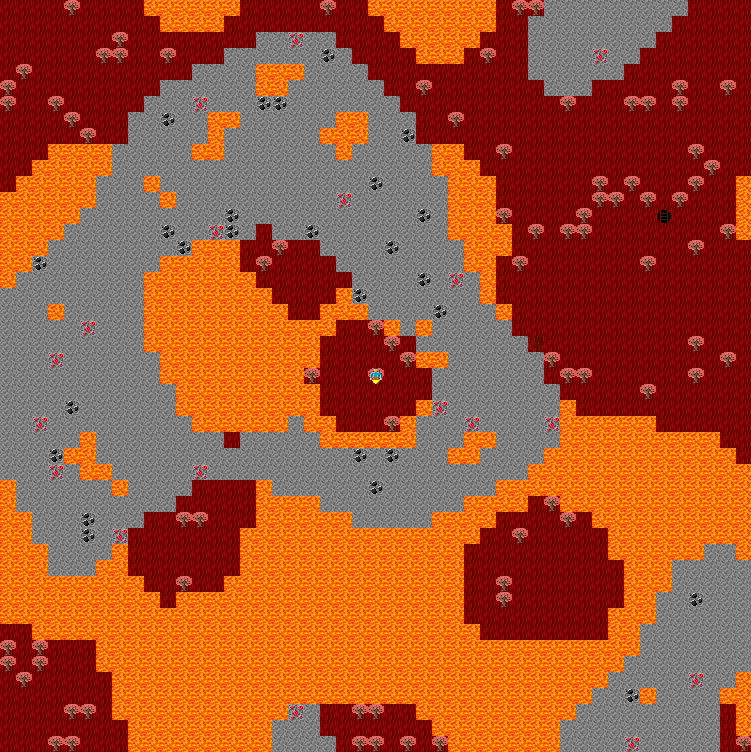}
    \end{subfigure}\hfill%
    \caption{\texttt{Craftax}: Floor 6 - Fire Realm}
    \label{fig:app_craftax_classic_level_6}

    \centering
    \begin{subfigure}{0.28\linewidth}
        \includegraphics[width=1\linewidth]{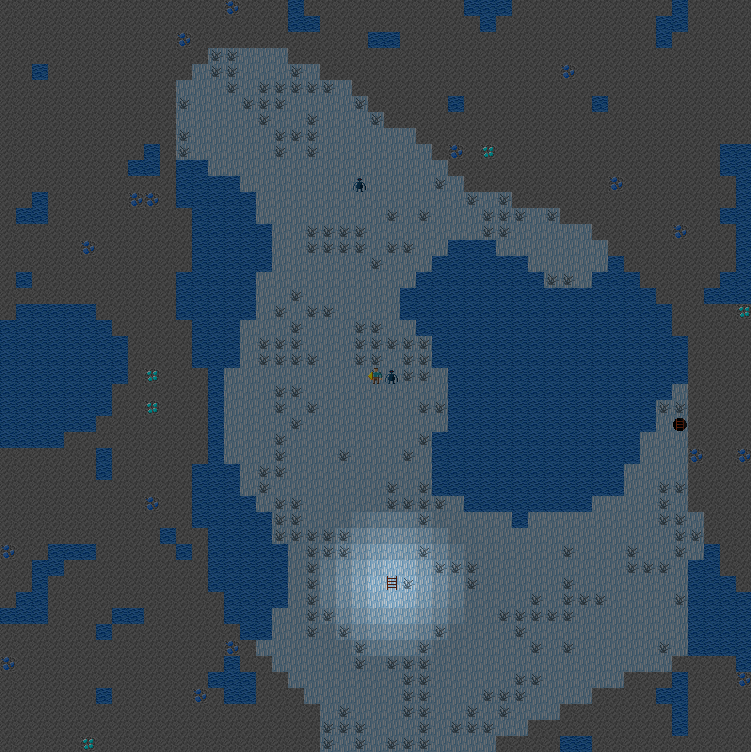}
    \end{subfigure}\hfill%
    \begin{subfigure}{0.28\linewidth}
        \includegraphics[width=1\linewidth]{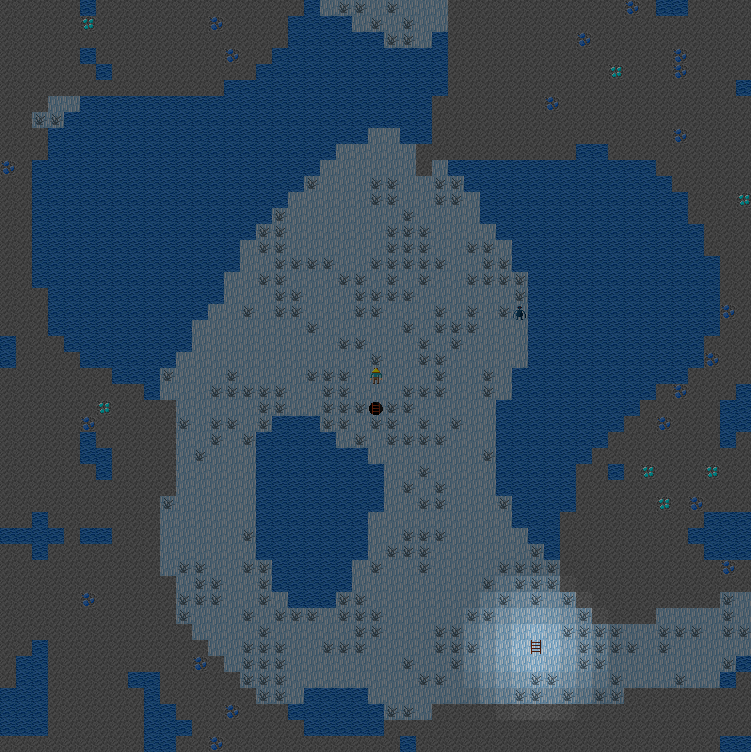}
    \end{subfigure}\hfill%
    \begin{subfigure}{0.28\linewidth}
        \includegraphics[width=1\linewidth]{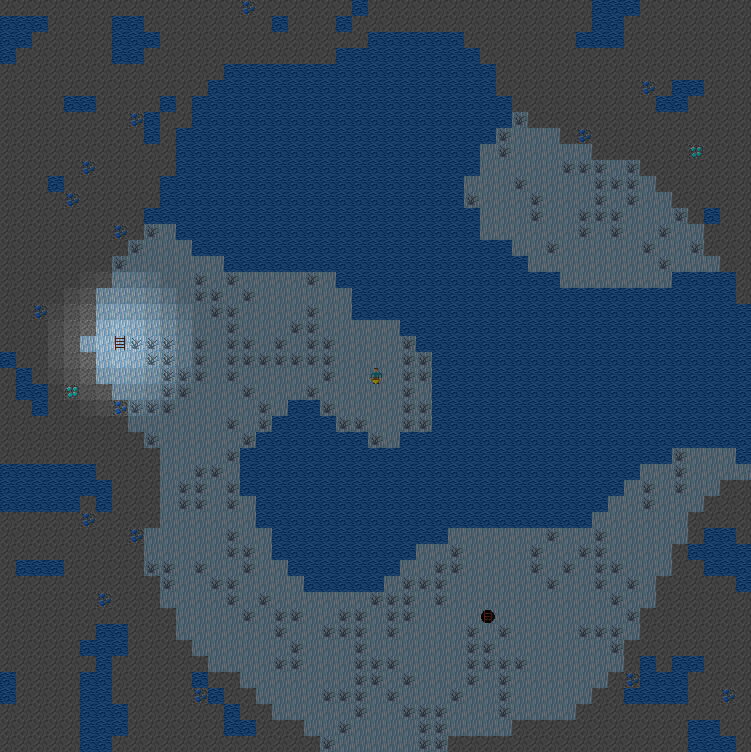}
    \end{subfigure}\hfill%
    \caption{\texttt{Craftax}: Floor 7 - Ice Realm}
    \label{fig:app_craftax_classic_level_7}

    \centering
    \begin{subfigure}{0.28\linewidth}
        \includegraphics[width=1\linewidth]{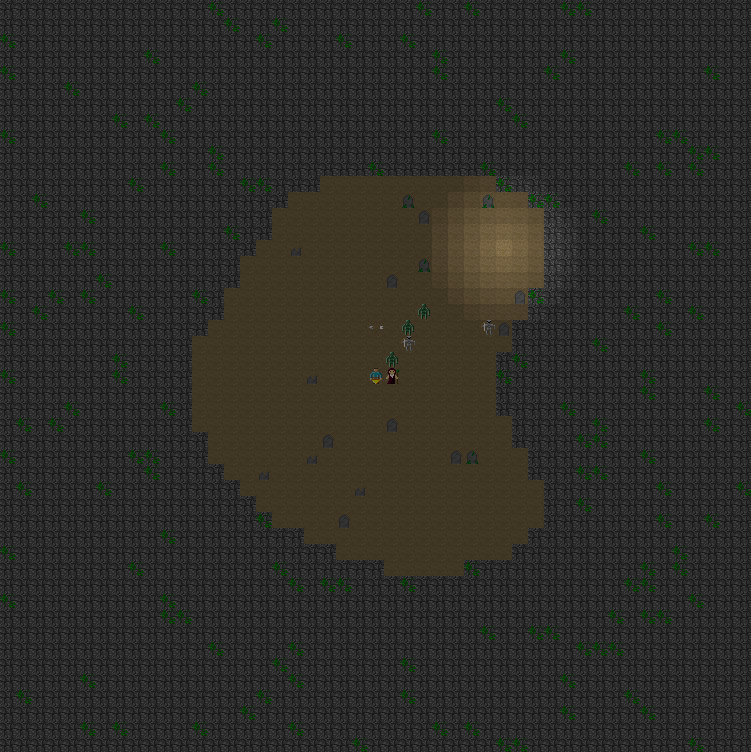}
    \end{subfigure}\hfill%
    \begin{subfigure}{0.28\linewidth}
        \includegraphics[width=1\linewidth]{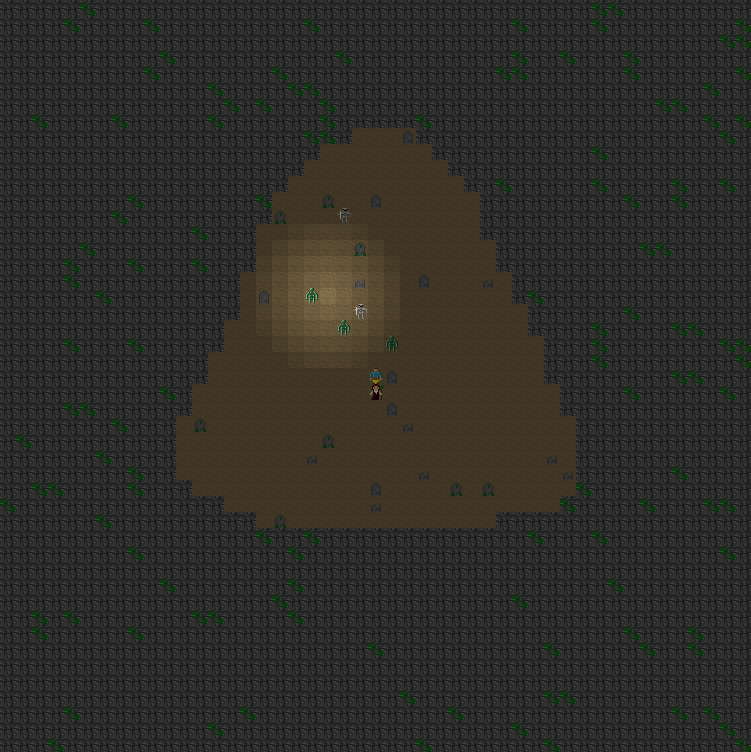}
    \end{subfigure}\hfill%
    \begin{subfigure}{0.28\linewidth}
        \includegraphics[width=1\linewidth]{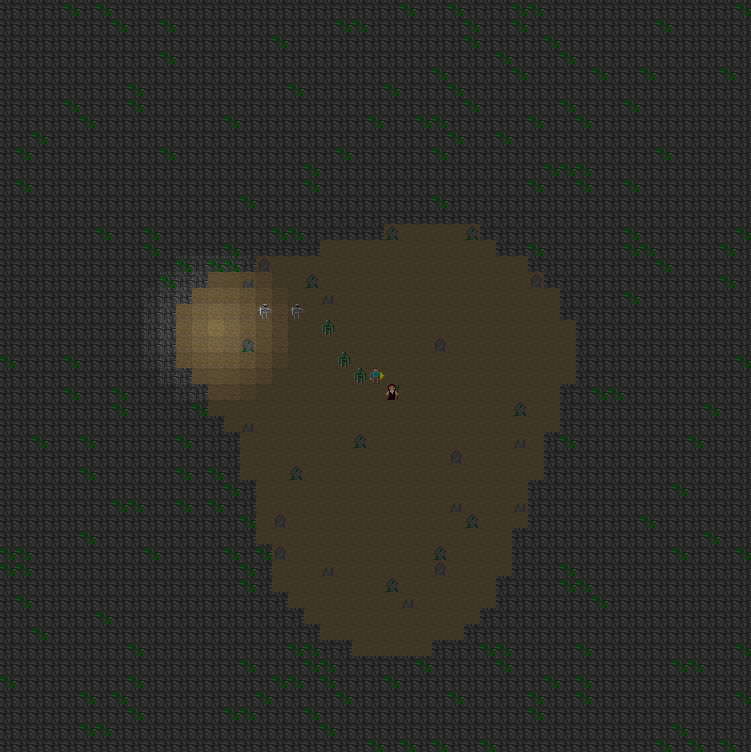}
    \end{subfigure}\hfill%
    \caption{\texttt{Craftax}: Floor 8 - Graveyard}
    \label{fig:app_craftax_classic_level_8}

\end{figure*}

Examples of procedurally generated floors are shown in Figures \ref{fig:app_craftax_classic_level_1}, \ref{fig:app_craftax_classic_level_2}, \ref{fig:app_craftax_classic_level_3}, \ref{fig:app_craftax_classic_level_4}, \ref{fig:app_craftax_classic_level_5}, \ref{fig:app_craftax_classic_level_6}, \ref{fig:app_craftax_classic_level_7} and \ref{fig:app_craftax_classic_level_8}. Floors 2, 5, 7 and 8 have been artificially lit up to make them visible, whereas in gameplay torches need to be placed to reveal them.  Floor 0 is the overworld and is essentially equivalent to the world from \texttt{Craftax-Classic}.

\subsection{Observation Space} \label{app:obs_space}

Both \texttt{Craftax} and \texttt{Craftax-Classic} have options for pixel-based and symbolic observations.  Each consist of a view of the map and the players stats and inventory.  The map view in \texttt{Craftax-Classic} is 7x9 squares, while \texttt{Craftax} uses 9x11.

\begin{figure*}
    \centering
    \begin{subfigure}{0.28\linewidth}
        \includegraphics[width=1\linewidth]{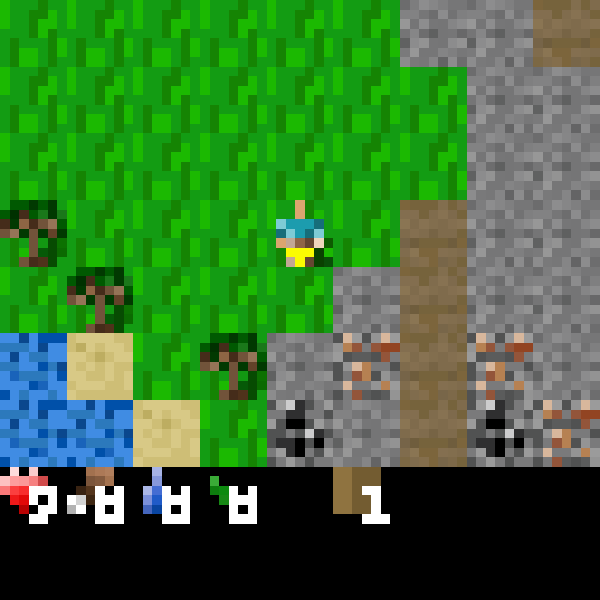}
        \caption{Crafter}
    \end{subfigure}\hfill%
    \begin{subfigure}{0.28\linewidth}
        \includegraphics[width=1\linewidth]{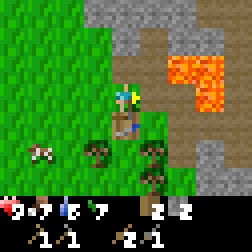}
        \caption{\texttt{Craftax-Classic}}
    \end{subfigure}\hfill%
    \begin{subfigure}{0.28\linewidth}
        \includegraphics[width=1\linewidth]{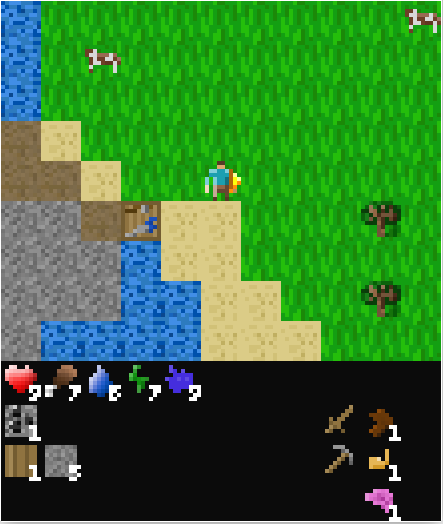}
        \caption{\texttt{Craftax}}
    \end{subfigure}
    \caption{Comparison of downscaled pixel views for agent observations.}
    \label{fig:app_pixels_view}
\end{figure*}

\textbf{Pixels} The pixel-based observations for \texttt{Craftax-Classic} take the same form as those in Crafter, with each 16x16 square downscaled to 7x7.  For \texttt{Craftax}, we downscale only to 10x10.  This is because we deal with numbers greater than 9, meaning that the digits had to be rendered in a smaller font size.  At 7x7 downscaling many of these digits were indistinguishable.  Pixel observations from Crafter, \texttt{Craftax-Classic} and \texttt{Craftax} are shown in Figure \ref{fig:app_pixels_view}.

\textbf{Symbolic}  The primary observation space we experiment on is symbolic, as it allows for significantly faster runtime.  Each square on the map encodes three one-hot vectors and a light level.  The first one-hot vector is of size 37 and encodes the block type, the second is of size 5 and encodes the item type, while the third is of size 36 and encodes the creature type.  The light level is a float between 0 and 1.  If the light level for a block is $<0.05$ then the block, item and creature are masked out.

The inventory/stats bar is represented as an array with each element representing the amount corresponding to a particular item or attribute.  All values are scaled to be roughly in the range 0-1.  All inventory items (which the player can carry up to 99 of) are represented as $\frac{\sqrt{n}}{10}$.  Table \ref{tab:symbolic_obs_inv} lists the symbolic inventory observation for \texttt{Craftax}.  The observation for \texttt{Craftax-Classic} is the subset of these elements that exist in the classic version.

\textbf{Textual} Although we don't make use of it for our experiments, we also provide a textual `renderer'.  This encodes the grid in the agents view as ``\texttt{<CREATURE>} on \texttt{<ITEM>} on \texttt{<BLOCK>}" for each viewable block, so a troll standing on a path block with a torch on it would be encoded as ``troll on torch on path".  We encode the players inventory and intrinsics as ``\texttt{<INVENTORY\_ITEM>}: X", so an agent with 2 diamonds in its inventory would see ``\texttt{DIAMOND}: 2".  Use in conjunction with the text tutorial found in the codebase, this could be use for language-conditioned learning.

\begin{table*}[t]
\centering
\begin{tabular}{@{}l c c c@{}} 
    \toprule
    \textbf{Item} & \textbf{Representation} & \textbf{Min} & \textbf{Max} \\
    \midrule
    Wood & $\frac{\sqrt{n}}{10}$ & $0$ & $1$ \\
    Stone & $\frac{\sqrt{n}}{10}$ & $0$ & $1$ \\
    Coal & $\frac{\sqrt{n}}{10}$ & $0$ & $1$ \\
    Iron & $\frac{\sqrt{n}}{10}$ & $0$ & $1$ \\
    Diamond & $\frac{\sqrt{n}}{10}$ & $0$ & $1$ \\
    Sapphire & $\frac{\sqrt{n}}{10}$ & $0$ & $1$ \\
    Ruby & $\frac{\sqrt{n}}{10}$ & $0$ & $1$ \\
    Sapling & $\frac{\sqrt{n}}{10}$ & $0$ & $1$ \\
    Torch & $\frac{\sqrt{n}}{10}$ & $0$ & $1$ \\
    Arrow & $\frac{\sqrt{n}}{10}$ & $0$ & $1$ \\
    Potion (x6) & $\frac{\sqrt{n}}{10}$ & $0$ & $1$ \\
    Book & $\frac{n}{2}$ & $0$ & $1$ \\
    Pickaxe & $\frac{\text{level}}{4}$ & $0$ & $1$ \\
    Sword & $\frac{\text{level}}{4}$ & $0$ & $1$ \\
    Sword Enchantment & $\mathbb{I}_{\text{fire}} + 2 \cdot \mathbb{I}_{\text{ice}} $ & $0$ & $2$ \\
    Bow & $\mathbb{I}_{\text{bow}} $ & $0$ & $1$ \\
    Armour (x4) & $\frac{\text{level}}{2}$ & $0$ & $1$ \\
    Armour Enchantment (x4) & $\mathbb{I}_{\text{fire}} + 2 \cdot \mathbb{I}_{\text{ice}} $ & $0$ & $2$ \\
    Health & $\frac{x}{10}$ & $0$ & $1.4$ \\
    Food & $\frac{x}{10}$ & $0$ & $1.7$ \\
    Drink & $\frac{x}{10}$ & $0$ & $1.7$ \\
    Energy & $\frac{x}{10}$ & $0$ & $1.7$ \\
    Mana & $\frac{x}{10}$ & $0$ & $2.1$ \\
    XP & $\frac{x}{10}$ & $0$ & $0.8$ \\
    Dexterity & $\frac{x}{10}$ & $0.1$ & $0.5$ \\
    Strength & $\frac{x}{10}$ & $0.1$ & $0.5$ \\
    Intelligence & $\frac{x}{10}$ & $0.1$ & $0.5$ \\
    Direction & $\text{onehot}(n)$ & $-$ & $-$ \\
    Day/Night & $x$ & $0$ & $1$ \\
    Sleeping & $\mathbb{I}_{\text{sleep}}$ & $0$ & $1$ \\
    Resting & $\mathbb{I}_{\text{rest}}$ & $0$ & $1$ \\
    Learned Fireball & $\mathbb{I}_{\text{fireball}}$ & $0$ & $1$ \\
    Learned Iceball & $\mathbb{I}_{\text{iceball}}$ & $0$ & $1$ \\
    Floor & $\frac{n}{10}$ & $0$ & $0.9$ \\
    Floor Cleared & $\mathbb{I}_{\text{cleared floor}}$ & $0$ & $1$ \\
    Boss Vulnerable & $\mathbb{I}_{\text{boss vulnerable}}$ & $0$ & $1$ \\
    \bottomrule
\end{tabular}
\caption{Symbolic observations for player inventory.}
\label{tab:symbolic_obs_inv}
\end{table*}

\subsection{Action Space} \label{app:action_space}

The action space for \texttt{Craftax} is shown in Table \ref{tab:action_mapping}.

\begin{table*}[t]
\centering
\begin{tabular}{@{}l l c@{}} 
    \toprule
    \textbf{Action ID} & \textbf{Action Name} & \textbf{Key} \\
    \midrule
    0 & NOOP & \keystroke{Q} \\
    1 & LEFT & \keystroke{A} \\
    2 & RIGHT & \keystroke{D} \\
    3 & UP & \keystroke{W} \\
    4 & DOWN & \keystroke{S} \\
    5 & DO & \keystroke{Space} \\
    6 & SLEEP & \keystroke{Tab} \\
    7 & PLACE\_STONE & \keystroke{R} \\
    8 & PLACE\_TABLE & \keystroke{T} \\
    9 & PLACE\_FURNACE & \keystroke{F} \\
    10 & PLACE\_PLANT & \keystroke{P} \\
    11 & MAKE\_WOOD\_PICKAXE & \keystroke{1} \\
    12 & MAKE\_STONE\_PICKAXE & \keystroke{2} \\
    13 & MAKE\_IRON\_PICKAXE & \keystroke{3} \\
    14 & MAKE\_WOOD\_SWORD & \keystroke{5} \\
    15 & MAKE\_STONE\_SWORD & \keystroke{6} \\
    16 & MAKE\_IRON\_SWORD & \keystroke{7} \\
    17 & REST & \keystroke{e} \\
    18 & DESCEND & \keystroke{ . } \\
    19 & ASCEND & \keystroke{ , } \\
    20 & MAKE\_DIAMOND\_PICKAXE & \keystroke{4} \\
    \bottomrule
\end{tabular}
\quad
\begin{tabular}{@{}l l c@{}} 
    \toprule
    \textbf{Action ID} & \textbf{Action Name} & \textbf{Key} \\
    \midrule
    21 & MAKE\_DIAMOND\_SWORD & \keystroke{8} \\
    22 & MAKE\_IRON\_ARMOUR & \keystroke{Y} \\
    23 & MAKE\_DIAMOND\_ARMOUR & \keystroke{U} \\
    24 & SHOOT\_ARROW & \keystroke{ I } \\
    25 & MAKE\_ARROW & \keystroke{O} \\
    26 & CAST\_FIREBALL & \keystroke{G} \\
    27 & CAST\_ICEBALL & \keystroke{H} \\
    28 & PLACE\_TORCH & \keystroke{J} \\
    29 & DRINK\_POTION\_RED & \keystroke{Z} \\
    30 & DRINK\_POTION\_GREEN & \keystroke{X} \\
    31 & DRINK\_POTION\_BLUE & \keystroke{C} \\
    32 & DRINK\_POTION\_PINK & \keystroke{V} \\
    33 & DRINK\_POTION\_CYAN & \keystroke{B} \\
    34 & DRINK\_POTION\_YELLOW & \keystroke{N} \\
    35 & READ\_BOOK & \keystroke{M} \\
    36 & ENCHANT\_SWORD & \keystroke{K} \\
    37 & ENCHANT\_ARMOUR & \keystroke{L} \\
    38 & MAKE\_TORCH & \keystroke{ [ } \\
    39 & LEVEL\_UP\_DEXTERITY & \keystroke{ ] } \\
    40 & LEVEL\_UP\_STRENGTH & \keystroke{ - } \\
    41 & LEVEL\_UP\_INTELLIGENCE & \keystroke{=} \\
    42 & ENCHANT\_BOW & \keystroke{ ; } \\ 
    \bottomrule
\end{tabular}
\caption{Actions for \texttt{Craftax}.}
\label{tab:action_mapping}
\end{table*}

\subsection{Achievements} \label{app:achievements}

The achievements for \texttt{Craftax} are listed in Table \ref{tab:achievements_listing}.  Achievements for \texttt{Craftax-Classic} are those with an ID of up to and including 21.

\begin{table*}[t]
\centering
\begin{tabular}{@{}l l l@{}} 
    \toprule
    \textbf{ID} & \textbf{Name} & \textbf{Category} \\
    \midrule
    0 & COLLECT\_WOOD & Basic (1)\\
    1 & PLACE\_TABLE & Basic (1)\\
    2 & EAT\_COW & Basic (1)\\
    3 & COLLECT\_SAPLING & Basic (1)\\
    4 & COLLECT\_DRINK & Basic (1)\\
    5 & MAKE\_WOOD\_PICKAXE & Basic (1)\\
    6 & MAKE\_WOOD\_SWORD & Basic (1)\\
    7 & PLACE\_PLANT & Basic (1)\\
    8 & DEFEAT\_ZOMBIE & Basic (1)\\
    9 & COLLECT\_STONE & Basic (1)\\
    10 & PLACE\_STONE & Basic (1)\\
    11 & EAT\_PLANT & Basic (1)\\
    12 & DEFEAT\_SKELETON & Basic (1)\\
    13 & MAKE\_STONE\_PICKAXE & Basic (1)\\
    14 & MAKE\_STONE\_SWORD & Basic (1)\\
    15 & WAKE\_UP & Basic (1)\\
    16 & PLACE\_FURNACE & Basic (1)\\
    17 & COLLECT\_COAL & Basic (1)\\
    18 & COLLECT\_IRON & Basic (1)\\
    19 & COLLECT\_DIAMOND & Basic (1)\\
    20 & MAKE\_IRON\_PICKAXE & Basic (1)\\
    21 & MAKE\_IRON\_SWORD & Basic (1)\\
    22 & MAKE\_ARROW & Basic (1)\\
    23 & MAKE\_TORCH & Basic (1)\\
    24 & PLACE\_TORCH & Basic (1)\\
    25 & MAKE\_DIAMOND\_SWORD & Intermediate (3)\\
    26 & MAKE\_IRON\_ARMOUR & Intermediate (3)\\
    27 & MAKE\_DIAMOND\_ARMOUR & Intermediate (3)\\
    28 & ENTER\_GNOMISH\_MINES & Intermediate (3)\\
    29 & ENTER\_DUNGEON & Intermediate (3)\\
    30 & ENTER\_SEWERS & Advanced (5)\\
    31 & ENTER\_VAULT & Advanced (5)\\
    32 & ENTER\_TROLL\_MINES & Advanced (5)\\
    \bottomrule
\end{tabular}
\quad
\begin{tabular}{@{}l l l@{}} 
    \toprule
    \textbf{ID} & \textbf{Name} & \textbf{Category} \\
    \midrule
    33 & ENTER\_FIRE\_REALM & Very Advanced (8)\\
    34 & ENTER\_ICE\_REALM & Very Advanced (8)\\
    35 & ENTER\_GRAVEYARD & Very Advanced (8)\\
    36 & DEFEAT\_GNOME\_WARRIOR & Intermediate (3)\\
    37 & DEFEAT\_GNOME\_ARCHER & Intermediate (3)\\
    38 & DEFEAT\_ORC\_SOLIDER & Intermediate (3)\\
    39 & DEFEAT\_ORC\_MAGE & Intermediate (3)\\
    40 & DEFEAT\_LIZARD & Advanced (5)\\
    41 & DEFEAT\_KOBOLD & Advanced (5)\\
    42 & DEFEAT\_TROLL & Advanced (5)\\
    43 & DEFEAT\_DEEP\_THING & Advanced (5)\\
    44 & DEFEAT\_PIGMAN & Very Advanced (8)\\
    45 & DEFEAT\_FIRE\_ELEMENTAL & Very Advanced (8)\\
    46 & DEFEAT\_FROST\_TROLL & Very Advanced (8)\\
    47 & DEFEAT\_ICE\_ELEMENTAL & Very Advanced (8)\\
    48 & DAMAGE\_NECROMANCER & Very Advanced (8)\\
    49 & DEFEAT\_NECROMANCER & Very Advanced (8)\\
    50 & EAT\_BAT & Intermediate (3)\\
    51 & EAT\_SNAIL & Intermediate (3)\\
    52 & FIND\_BOW & Intermediate (3)\\
    53 & FIRE\_BOW & Intermediate (3)\\
    54 & COLLECT\_SAPPHIRE & Intermediate (3)\\
    55 & LEARN\_FIREBALL & Advanced (5)\\
    56 & CAST\_FIREBALL & Advanced (5)\\
    57 & LEARN\_ICEBALL & Advanced (5)\\
    58 & CAST\_ICEBALL & Advanced (5)\\
    59 & COLLECT\_RUBY & Intermediate (3)\\
    60 & MAKE\_DIAMOND\_PICKAXE & Intermediate (3)\\
    61 & OPEN\_CHEST & Intermediate (3)\\
    62 & DRINK\_POTION & Intermediate (3)\\
    63 & ENCHANT\_SWORD & Advanced (5)\\
    64 & ENCHANT\_ARMOUR & Advanced (5)\\
    65 & DEFEAT\_KNIGHT & Advanced (5)\\
    66 & DEFEAT\_ARCHER & Advanced (5)\\
    \bottomrule
\end{tabular}
\caption{Achievements listing for \texttt{Craftax}.  The category column indicates the difficulty classification of the achievement and the associated reward.}
\label{tab:achievements_listing}
\end{table*}

\section{Hyperparameter Tuning} \label{app:hyp}

\subsection{Craftax-1B}
The hyperparameters and considered values for PPO on Craftax-1B are shown in Table \ref{tab:craftax_1b_hyp_ppo}.  Each run was performed for 1 billion timesteps with 1 seed.  We started from a set of values picked informally and then tuned each hyperparameter individually.  These hyperparameters were then taken as the baseline for the other methods, which all build off of the PPO implementation.  For PPO-RNN, we kept the same hyperparameters as PPO, with the hidden size being equal to the MLP layer size.  The hyperparameters for ICM, E3B and RND are shown in Tables \ref{tab:craftax_1b_hyp_icm}, \ref{tab:craftax_1b_hyp_e3b} and \ref{tab:craftax_hyp_rnd} respectively.

\begin{table*}[t]
\centering
\begin{tabular}{@{}l l l@{}} 
    \toprule
    \textbf{Hyperparameter} & \textbf{Considered Values} & \textbf{Value} \\
    \midrule
    Entropy Coefficient & $\{0, 0.01\}$ & $0.01$ \\
    GAE $\lambda$ & $\{0.8, 0.9, 0.95\}$ & $0.8$ \\
    $\gamma$ & $\{0.98, 0.99, 0.995, 0.999, 0.9995\}$ & $0.99$ \\
    MLP Layer Size & $\{256, 512, 1024\}$ & $512$ \\
    Learning Rate & $\{0.0002, 0.0003, 0.0004\}$ & $0.0002$ \\
    Environment Workers & $\{256, 512, 1024\}$ & $1024$ \\
    \# Minibatches & $\{4, 8, 16\}$ & $8$ \\
    \# Steps & $\{64\}$ & $64$ \\
    \# Update Epochs & $\{4\}$ & $4$ \\
    Clip $\epsilon$ & $\{0.2\}$ & $0.2$ \\
    Value Function Coefficient & $\{0.5\}$ & $0.5$ \\
    Activation Function & $\{\text{tanh}\}$ & tanh \\
    Anneal Learning Rate & $\{\text{True}\}$ & True \\
    \bottomrule
\end{tabular}
\caption{Hyperparameters for PPO for Craftax-1B.}
\label{tab:craftax_1b_hyp_ppo}
\end{table*}

\begin{table*}[t]
\centering
\begin{tabular}{@{}l l l@{}} 
    \toprule
    \textbf{Hyperparameter} & \textbf{Considered Values} & \textbf{Value} \\
    \midrule
    Intrinsic Reward Coefficient $\beta$ & $\{0.1, 0.5, 1.0\}$ & $1.0$ \\
    World Models Learning Rate & $\{0.0003\}$ & $0.0003$ \\
    Forward Loss Coefficient & $\{1.0\}$ & $1.0$ \\
    Inverse Loss Coefficient & $\{1.0\}$ & $1.0$ \\
    World Models Layer Size & $\{256\}$ & $256$ \\
    Latent Size & $\{32, 64, 128\}$ & $32$ \\
    \bottomrule
\end{tabular}
\caption{Hyperparameters for ICM for Craftax-1B.}
\label{tab:craftax_1b_hyp_icm}
\end{table*}

\begin{table*}[t]
\centering
\begin{tabular}{@{}l l l@{}} 
    \toprule
    \textbf{Hyperparameter} & \textbf{Considered Values} & \textbf{Value} \\
    \midrule
    Intrinsic Reward Coefficient $\beta$ & $\{0.0005, 0.001, 0.005\}$ & $0.001$ \\
    Ridge Regulariser $\lambda$ & $\{0.005, 0.1, 0.5\}$ & $0.1$ \\
    \bottomrule
\end{tabular}
\caption{Hyperparameters for E3B for Craftax-1B.}
\label{tab:craftax_1b_hyp_e3b}
\end{table*}

\begin{table*}[t]
\centering
\begin{tabular}{@{}l l l@{}} 
    \toprule
    \textbf{Hyperparameter} & \textbf{Considered Values} & \textbf{Value} \\
    \midrule
    Intrinsic Reward Coefficient $\beta$ & $\{0.01, 0.1, 1.0\}$ & $1.0$ \\
    RND Output Dimension & $\{4, 32, 256, 2048\}$ & $256$ \\
    Distillation Network Learning Rate & $\{0.00003, 0.0003, 0.003\}$ & $0.0003$ \\
    RND Loss Coefficient & $\{0.0001, 0.001, 0.01\}$ & $0.01$ \\
    RND GAE Coefficient & $\{0.01, 0.1, 1.0\}$ & $0.01$ \\
    \bottomrule
\end{tabular}
\caption{Hyperparameters for RND for Craftax-1B and Craftax-1M.}
\label{tab:craftax_hyp_rnd}
\end{table*}

\subsection{Craftax-1M}
The hyperparameters for Craftax-1M were tuned more thoroughly, as the experiments were very fast to run.  We performed a random search over the PPO hyperparameters to obtain a baseline PPO implementation, before then running a random search for ICM and E3B.  The hyperparameters for PPO, ICM and E3B are shown in Tables \ref{tab:craftax_1m_hyp_ppo}, \ref{tab:craftax_1m_hyp_icm} and \ref{tab:craftax_1m_hyp_e3b} respectively.

\begin{table*}[t]
\centering
\begin{tabular}{@{}l l l@{}} 
    \toprule
    \textbf{Hyperparameter} & \textbf{Considered Values} & \textbf{Value} \\
    \midrule
    Entropy Coefficient & $\{0.01\}$ & $0.01$ \\
    GAE $\lambda$ & $\{0.7, 0.8, 0.9, 0.95\}$ & $0.8$ \\
    $\gamma$ & $\{0.99\}$ & $0.99$ \\
    MLP Layer Size & $\{128, 256, 512, 1024\}$ & $512$ \\
    Learning Rate & $\{0.0002, 0.0003, 0.0004\}$ & $0.0003$ \\
    Environment Workers & $\{16, 64, 256\}$ & $256$ \\
    \# Minibatches & $\{2, 4, 8, 16\}$ & $8$ \\
    \# Steps & $\{16, 64, 256\}$ & $16$ \\
    \# Update Epochs & $\{1, 2, 4, 8, 16\}$ & $4$ \\
    Clip $\epsilon$ & $\{0.2\}$ & $0.2$ \\
    Value Function Coefficient & $\{0.5\}$ & $0.5$ \\
    Activation Function & $\{\text{tanh}, \text{relu}\}$ & tanh \\
    Anneal Learning Rate & $\{\text{True}\}$ & True \\
    \bottomrule
\end{tabular}
\caption{Hyperparameters for PPO for Craftax-1M.}
\label{tab:craftax_1m_hyp_ppo}
\end{table*}

\begin{table*}[t]
\centering
\begin{tabular}{@{}l l l@{}} 
    \toprule
    \textbf{Hyperparameter} & \textbf{Considered Values} & \textbf{Value} \\
    \midrule
    Intrinsic Reward Coefficient $\beta$ & $\{0.01, 0.1, 1.0, 10.0, 100.0\}$ & $10.0$ \\
    World Models Learning Rate & $\{0.0003\}$ & $0.0003$ \\
    Forward Loss Coefficient & $\{0.1, 0.5, 1.0\}$ & $0.1$ \\
    Inverse Loss Coefficient & $\{0.1, 0.5, 1.0\}$ & $0.5$ \\
    World Models Layer Size & $\{256\}$ & $256$ \\
    Latent Size & $\{16, 32, 64, 128\}$ & $16$ \\
    \bottomrule
\end{tabular}
\caption{Hyperparameters for ICM for Craftax-1M.}
\label{tab:craftax_1m_hyp_icm}
\end{table*}

\begin{table*}[t]
\centering
\begin{tabular}{@{}l l l@{}} 
    \toprule
    \textbf{Hyperparameter} & \textbf{Considered Values} & \textbf{Value} \\
    \midrule
    Intrinsic Reward Coefficient $\beta$ & $\{0.0001, 0.001, 0.01, 0.1, 1.0\}$ & $0.01$ \\
    Ridge Regulariser $\lambda$ & $\{0.001, 0.01, 0.1, 0.5\}$ & $0.5$ \\
    \bottomrule
\end{tabular}
\caption{Hyperparameters for E3B for Craftax-1M.}
\label{tab:craftax_1m_hyp_e3b}
\end{table*}

\section{Further Results}
\subsection{Individual Achievements} \label{app:achievement_results}

\begin{figure}[H]
    \centering
    \includegraphics[width=\linewidth]{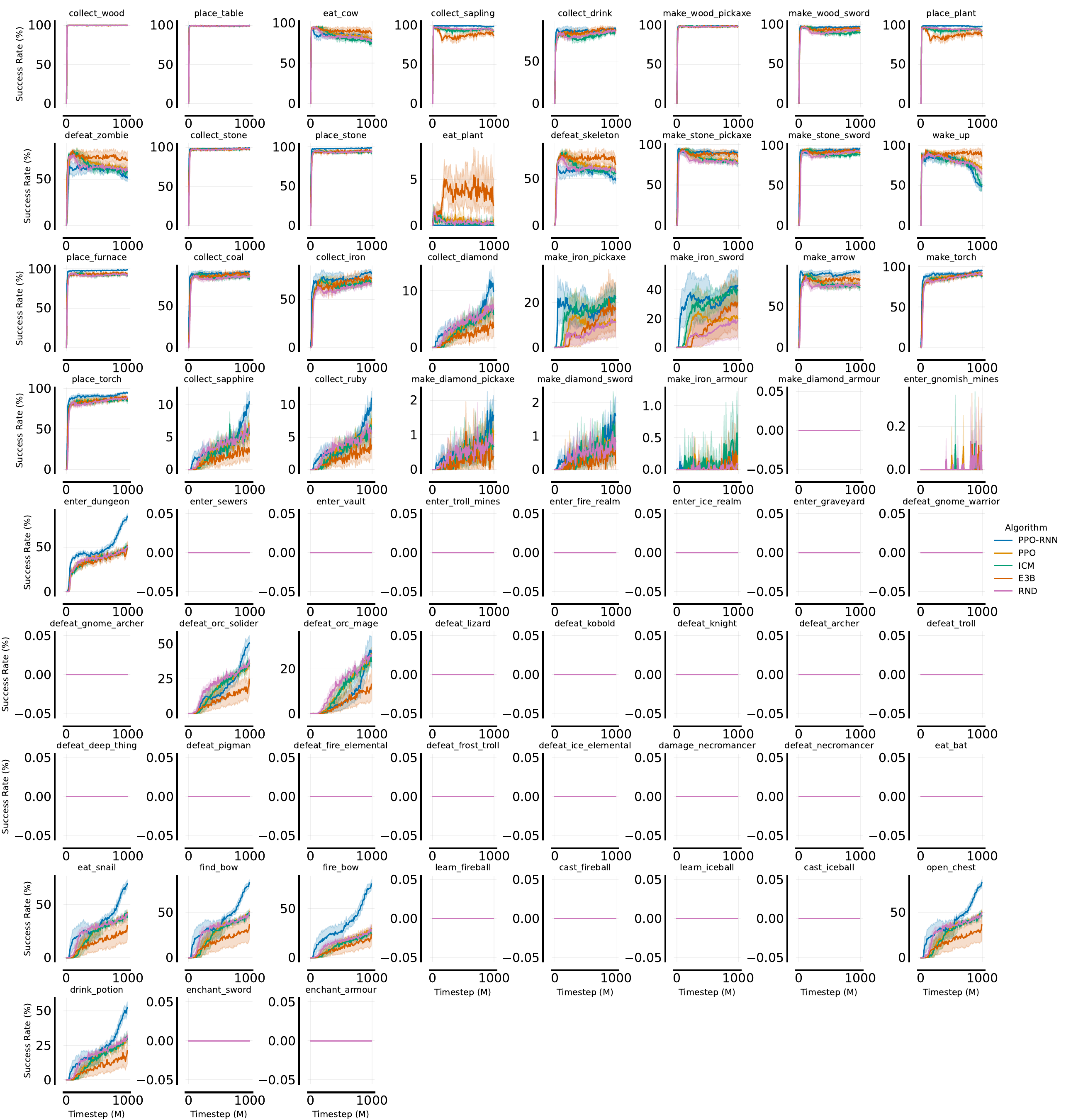}
    \caption{Success rate curves for all achievements on Craftax-1B.  Each algorithm was run on 10 seeds for 1 billion timesteps.  The shaded area denotes 1 standard error.}
    \label{fig:all_achievements_1b}
\end{figure}

\begin{figure}[H]
    \centering
    \includegraphics[width=\linewidth]{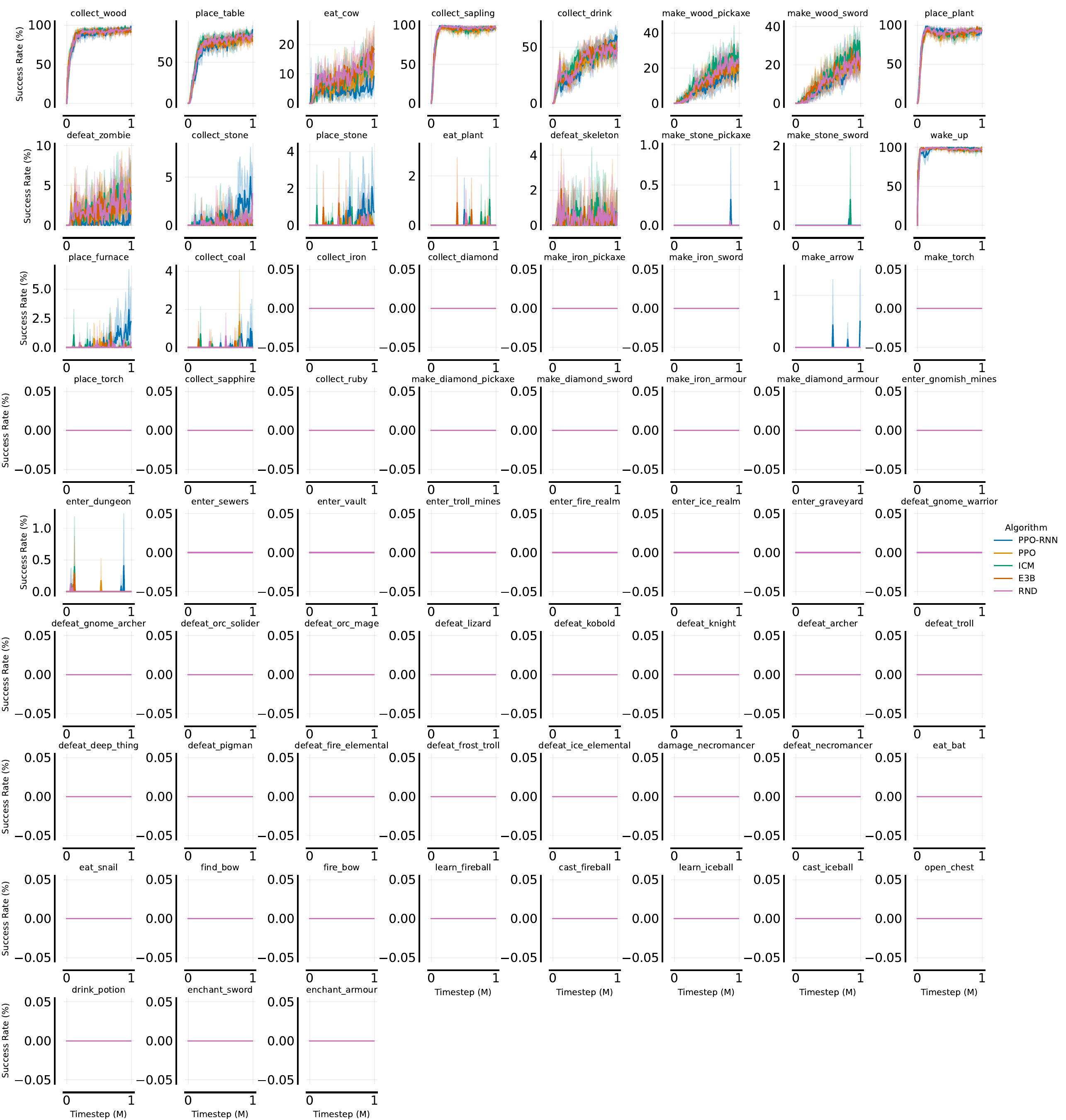}
    \caption{Success rate curves for all achievements on Craftax-1M.  Each algorithm was run on 10 seeds for 1 million timesteps.  The shaded area denotes 1 standard error.}
    \label{fig:all_achievements_1m}
\end{figure}

The success rates through training for Craftax-1B and Craftax-1M are shown in Figures \ref{fig:all_achievements_1b} and \ref{fig:all_achievements_1m} respectively.

\subsection{Craftax-Extended with 10B Environment Interactions} \label{app:craftax_10b}

\begin{figure}[H]
    \centering
    \includegraphics[width=\linewidth]{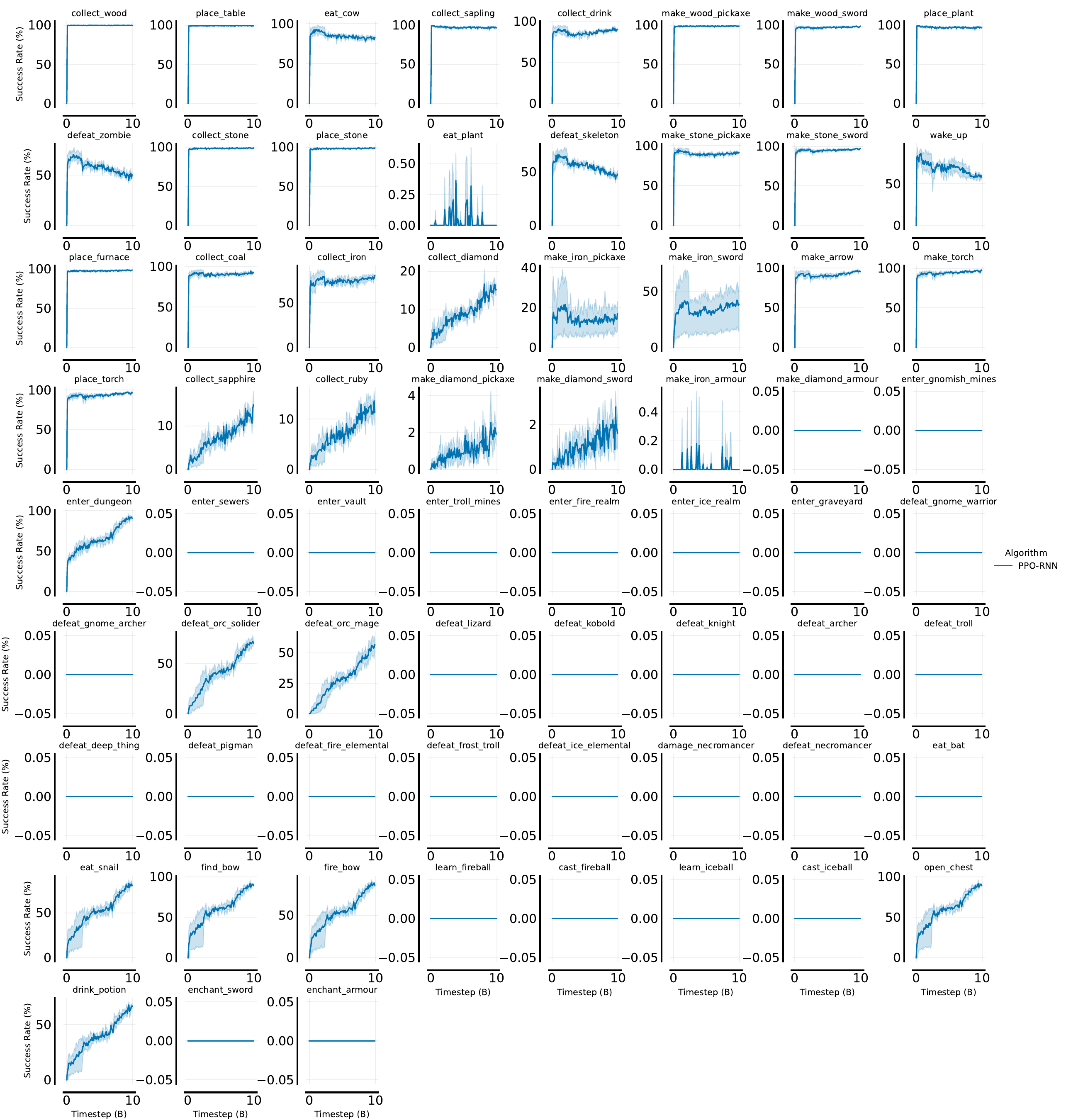}
    \caption{Achievement yields for PPO-RNN on \texttt{Craftax-Extended} given 10 billion environment interactions.  The experiment was repeated with 4 seeds and the shaded areas shows 1 standard error.}
    \label{fig:craftax_10b_achievements}
\end{figure}

\begin{figure}[H]
    \centering
    \includegraphics[width=0.5\linewidth]{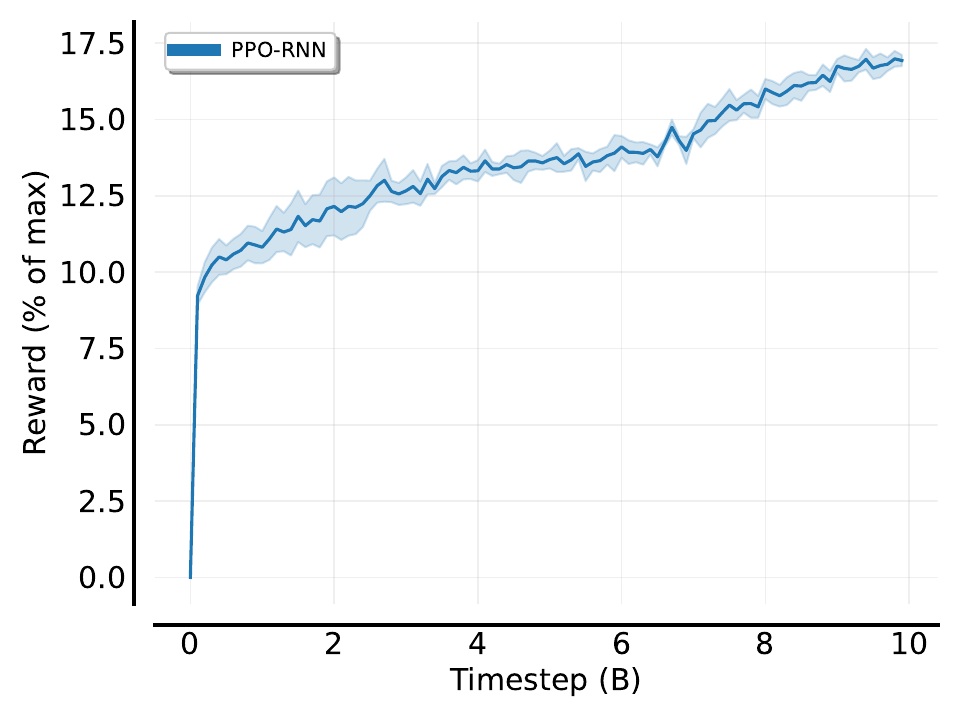}
    \caption{Reward for PPO-RNN on \texttt{Craftax-Extended} given 10 billion environment interactions.  The experiment was repeated with 4 seeds and the shaded areas shows 1 standard error.}
    \label{fig:craftax_10b}
\end{figure}

To really push the limits of the environment, we ran PPO-RNN on 10B environment interactions, with the results shown in Figures \ref{fig:craftax_10b} and \ref{fig:craftax_10b_achievements}.  The results show that learning for 10B iterations barely affects performance, with the agent still not making it into the gnomish mines.  Interestingly, as in the main results on Craftax-1B, the reward increases more quickly towards the end of the experiment.  The fact that this occurs at roughly the same value for both 1B and 10B implies that this is an artifact of the learning rate schedule.

\subsection{Craftax-Classic} \label{app:classic_experiments}

\begin{figure}[H]
    \centering
    \includegraphics[width=0.5\linewidth]{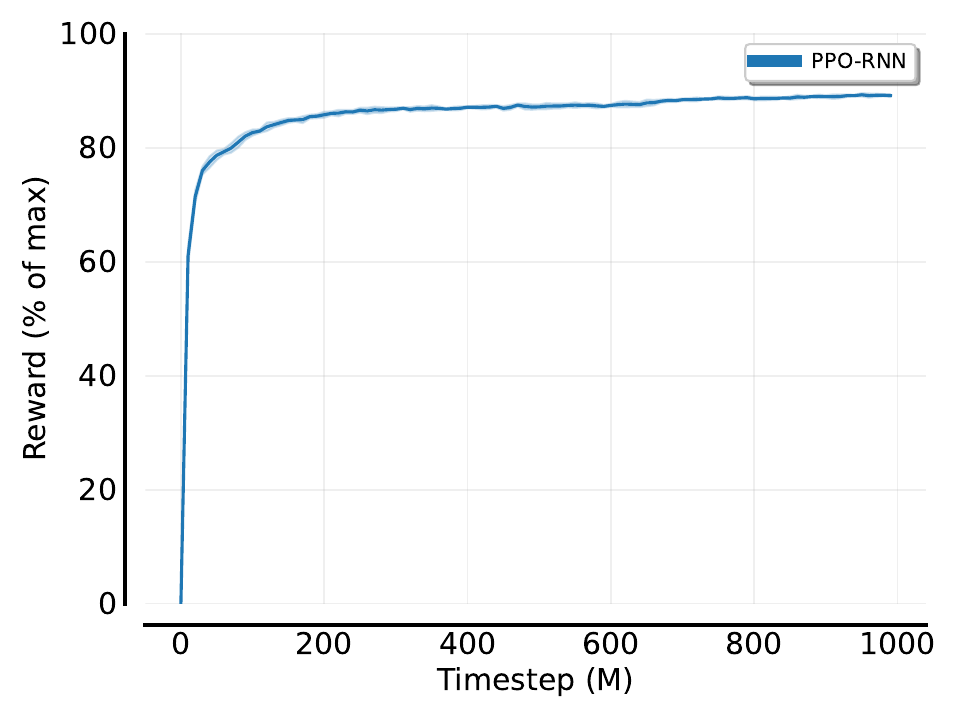}
    \caption{Reward for PPO-RNN on \texttt{Craftax-Classic} given 1 billion environment interactions.  The experiment was repeated with 10 seeds and the shaded areas shows 1 standard error.}
    \label{fig:classic_1b_reward}
\end{figure}

\begin{figure}[H]
    \centering
    \includegraphics[width=\linewidth]{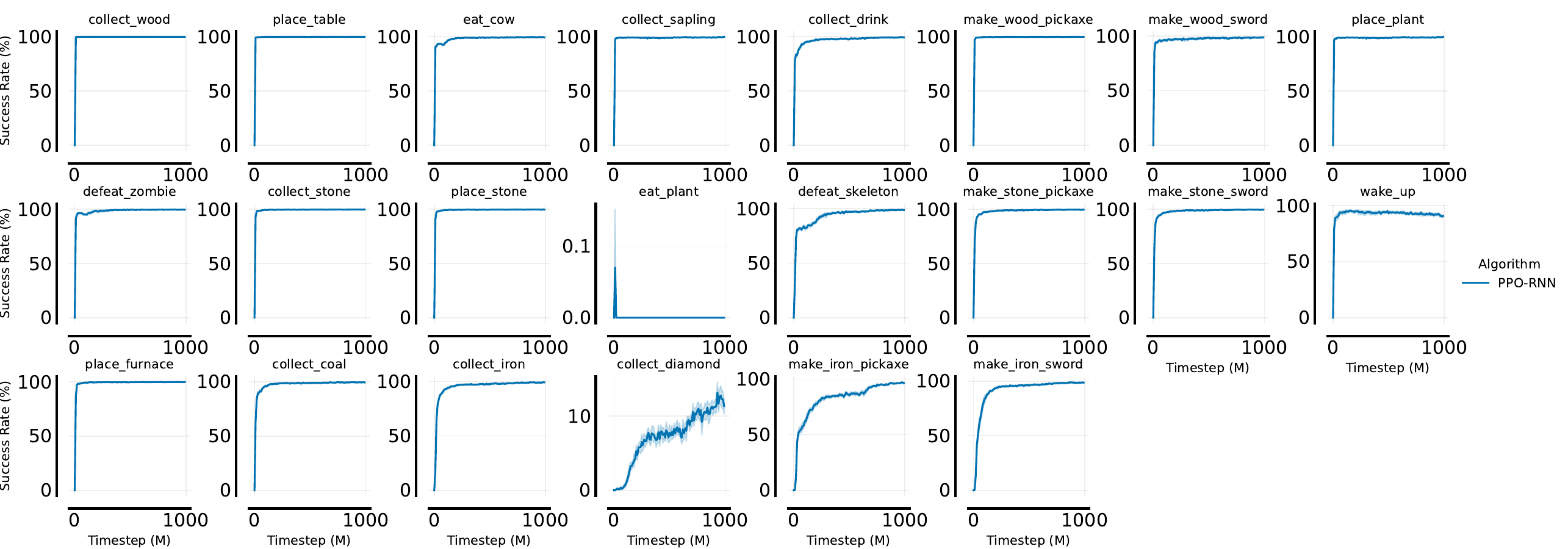}
    \caption{Achievement yields for PPO-RNN on \texttt{Craftax-Classic} given 1 billion environment interactions.  The experiment was repeated with 10 seeds and the shaded areas shows 1 standard error.}
    \label{fig:classic_1b_all_achievements}
\end{figure}

Figures \ref{fig:classic_1b_reward} and \ref{fig:classic_1b_all_achievements} show the results of running PPO-RNN on \texttt{Craftax-Classic} for 1 billion timesteps.  As can be seen, all achievements excepting \texttt{COLLECT\_DIAMOND} and \texttt{EAT\_PLANT} are reliably fulfilled.  Collecting diamonds is achieved somewhat reliably while eating plants is only achieved very rarely.  While the agent hasn't `solved' the benchmark by reliably completing every achievement, it is so close to doing so that releasing only the \texttt{Craftax-Classic} would be of little use, as it would likely be solved very quickly. 

\section{UED} \label{app:ued}
\subsection{Implementation Changes}
In standard implementations of UED~\citep{jiang2023Minimax}, the PPO rollout length is determined by the episode's length. This is because, whenever new levels are sampled, one rollout is performed. This is used to both (a) learn and (b) to approximate a regret value for this level. 
This does not pose a problem if the maximum episode length is short (such as in Minigrid); however, with \texttt{Craftax} and its long episodes, this approach is limited, as there must either be a long time between subsequent agent updates, or the score would only take into account part of an episode.

We modify the implementation to have multiple short inner rollouts, which the agent learns on as normal. After a certain number of these, we use the entire trajectory information to approximate regret, which then updates the adversary. This allows us to completely decouple these two aspects. Although this induces some non-stationarity---as the agent is learning on the trajectories that the adversary uses for scores---we found this does not significantly hamper performance.
Finally, since the agent rarely runs out of time, we restrict the episode length to be 4096 instead of the full 10000.
\subsection{Additional UED Results}

\begin{figure}[H]
    \begin{minipage}{0.47\linewidth}
        \centering
        \begin{subfigure}[t]{0.3\linewidth}
            \includegraphics[width=\linewidth]{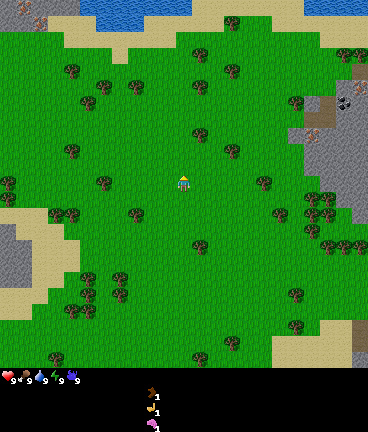}
        \end{subfigure}\hfill%
        \begin{subfigure}[t]{0.3\linewidth}
            \includegraphics[width=\linewidth]{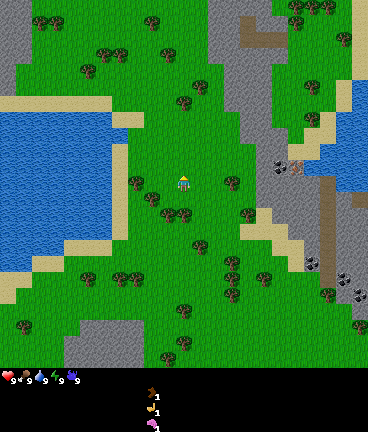}
        \end{subfigure}\hfill%
        \begin{subfigure}[t]{0.3\linewidth}
            \includegraphics[width=\linewidth]{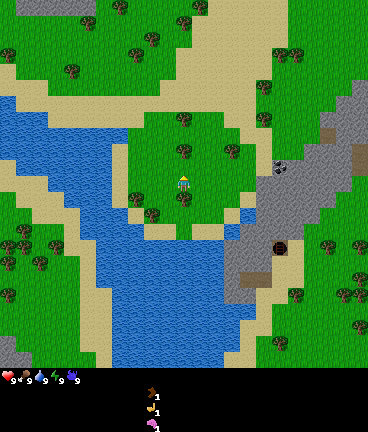}
        \end{subfigure}%
        \caption{ACCEL Noise replay levels at the start, middle and end of training}
        \label{fig:examples_accel_noise}
    \end{minipage}\hfill
    \begin{minipage}{0.47\linewidth}
        \centering
        \begin{subfigure}[t]{0.3\linewidth}
            \includegraphics[width=\linewidth]{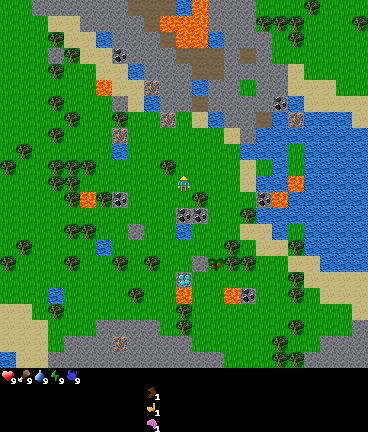}
        \end{subfigure}\hfill%
        \begin{subfigure}[t]{0.3\linewidth}
            \includegraphics[width=\linewidth]{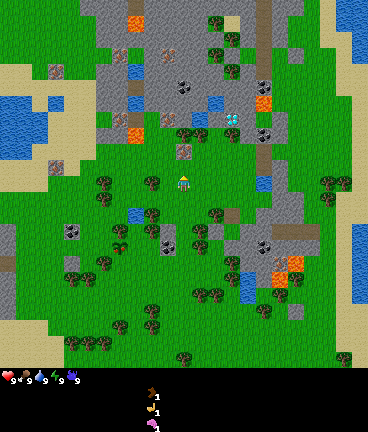}
        \end{subfigure}\hfill%
        \begin{subfigure}[t]{0.3\linewidth}
            \includegraphics[width=\linewidth]{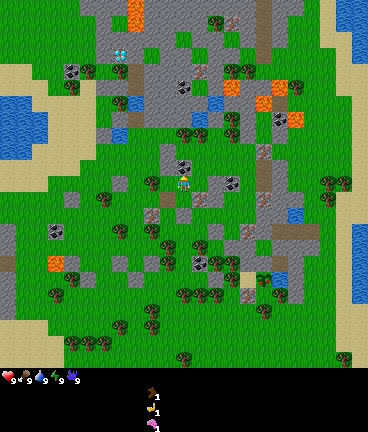}
        \end{subfigure}%
        \caption{ACCEL Swap replay levels at the start, middle and end of training}
        \label{fig:examples_accel_swap}
    \end{minipage}
\end{figure}

\begin{figure}[H]
\begin{minipage}{0.47\linewidth}
    \centering
    \begin{subfigure}[t]{0.3\linewidth}
        \includegraphics[width=\linewidth]{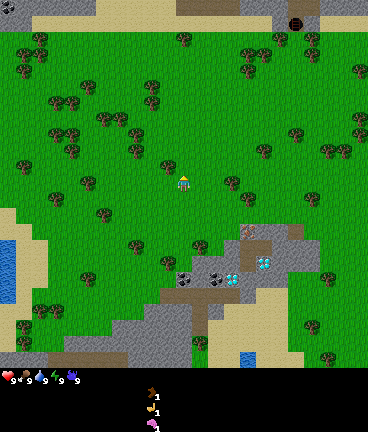}
    \end{subfigure}\hfill%
    \begin{subfigure}[t]{0.3\linewidth}
        \includegraphics[width=\linewidth]{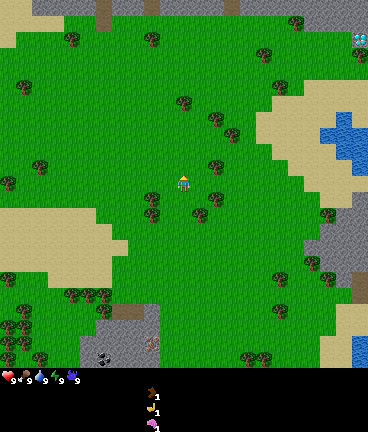}
    \end{subfigure}\hfill%
    \begin{subfigure}[t]{0.3\linewidth}
        \includegraphics[width=\linewidth]{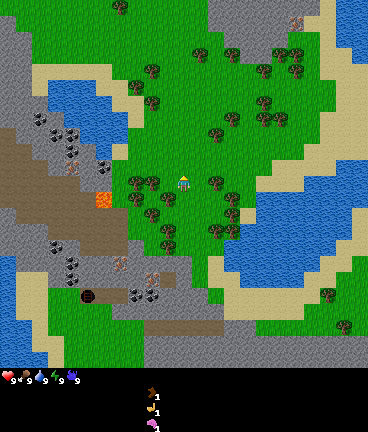}
    \end{subfigure}%
    \caption{ACCEL Swap (Restrict) replay levels at the start, middle and end of training}
    \label{fig:examples_accel_swap_restrict}
    \end{minipage}\hfill
    \begin{minipage}{0.47\linewidth}
    \centering
    \begin{subfigure}[t]{0.3\linewidth}
        \includegraphics[width=\linewidth]{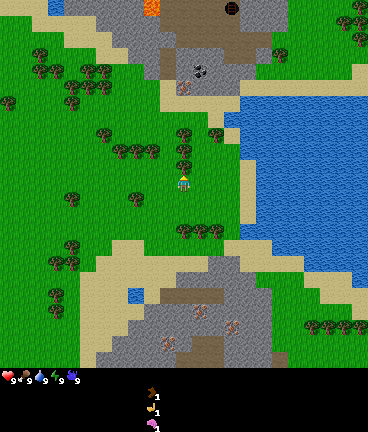}
    \end{subfigure}\hfill%
    \begin{subfigure}[t]{0.3\linewidth}
        \includegraphics[width=\linewidth]{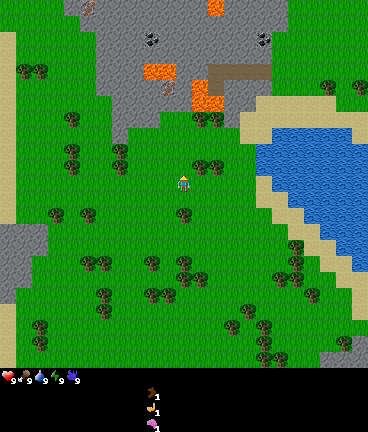}
    \end{subfigure}\hfill%
    \begin{subfigure}[t]{0.3\linewidth}
        \includegraphics[width=\linewidth]{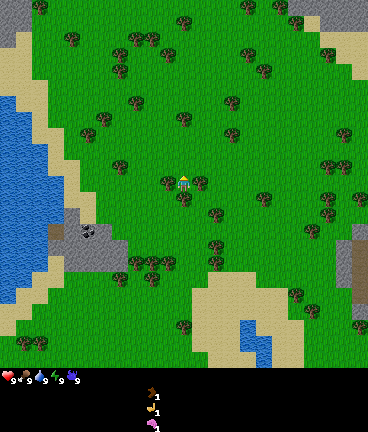}
    \end{subfigure}%
    \caption{Normal PLR replay levels at the start, middle and end of training}
    \label{fig:examples_normal_plr}
    \end{minipage}
\end{figure}

\begin{figure}[H]
    \begin{minipage}{0.47\linewidth}
    \centering
    \begin{subfigure}[t]{0.3\linewidth}
        \includegraphics[width=\linewidth]{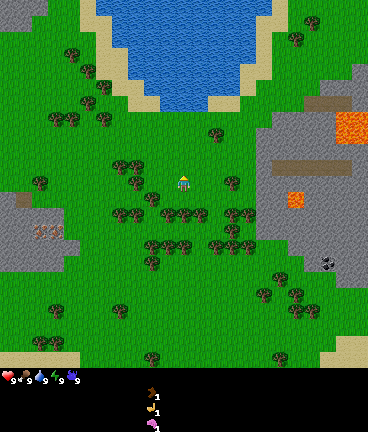}
    \end{subfigure}\hfill%
    \begin{subfigure}[t]{0.3\linewidth}
        \includegraphics[width=\linewidth]{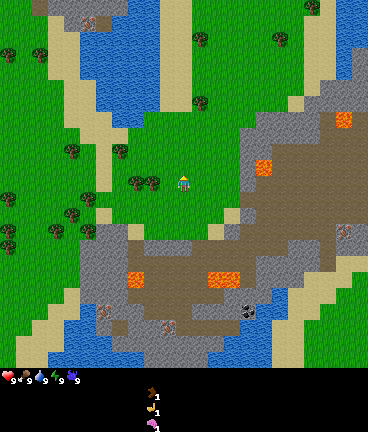}
    \end{subfigure}\hfill%
    \begin{subfigure}[t]{0.3\linewidth}
        \includegraphics[width=\linewidth]{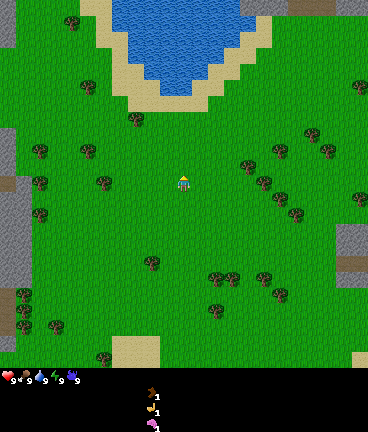}
    \end{subfigure}%
    \caption{Robust PLR replay levels at the start, middle and end of training}
    \label{fig:examples_robust_plr}
    \end{minipage}\hfill
    \begin{minipage}{0.47\linewidth}
    \centering
    \begin{subfigure}[t]{0.3\linewidth}
        \includegraphics[width=\linewidth]{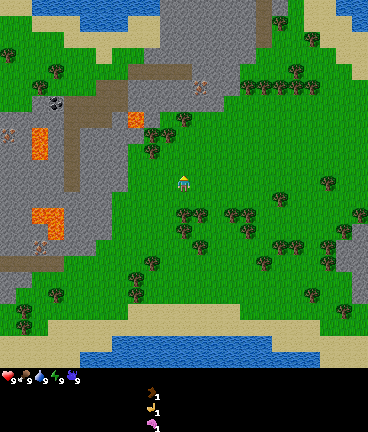}
    \end{subfigure}\hfill%
    \begin{subfigure}[t]{0.3\linewidth}
        \includegraphics[width=\linewidth]{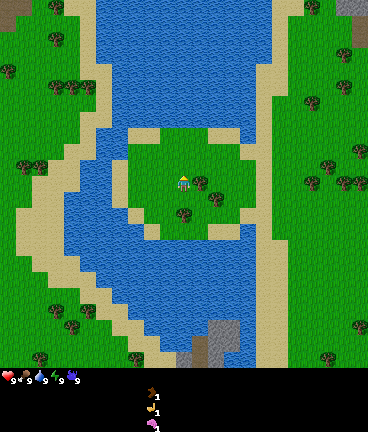}
    \end{subfigure}\hfill%
    \begin{subfigure}[t]{0.3\linewidth}
        \includegraphics[width=\linewidth]{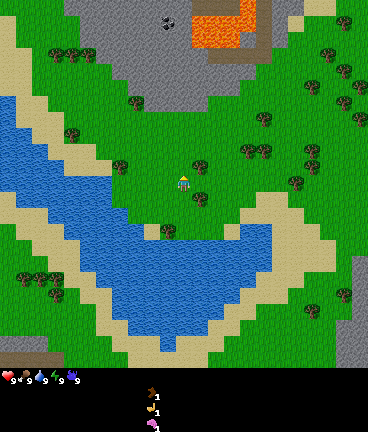}
    \end{subfigure}%
    \caption{DR levels at the start, middle and end of training}
    \label{fig:examples_dr}
    \end{minipage}
\end{figure}

\textbf{Qualitative Results}
Figures~\ref{fig:examples_accel_noise} to \ref{fig:examples_dr} in show several levels curated by each method. 
The curation-based methods generally cause resources to be more centrally-located, near where the player spawns. The unrestricted ACCEL swapping method generates levels that are visually very different to the normally generated ones, as there is no restriction to where it can place tiles. The other mutations and PLR generate more plausible levels.

\begin{figure*}
    \centering
        \includegraphics[width=\linewidth]{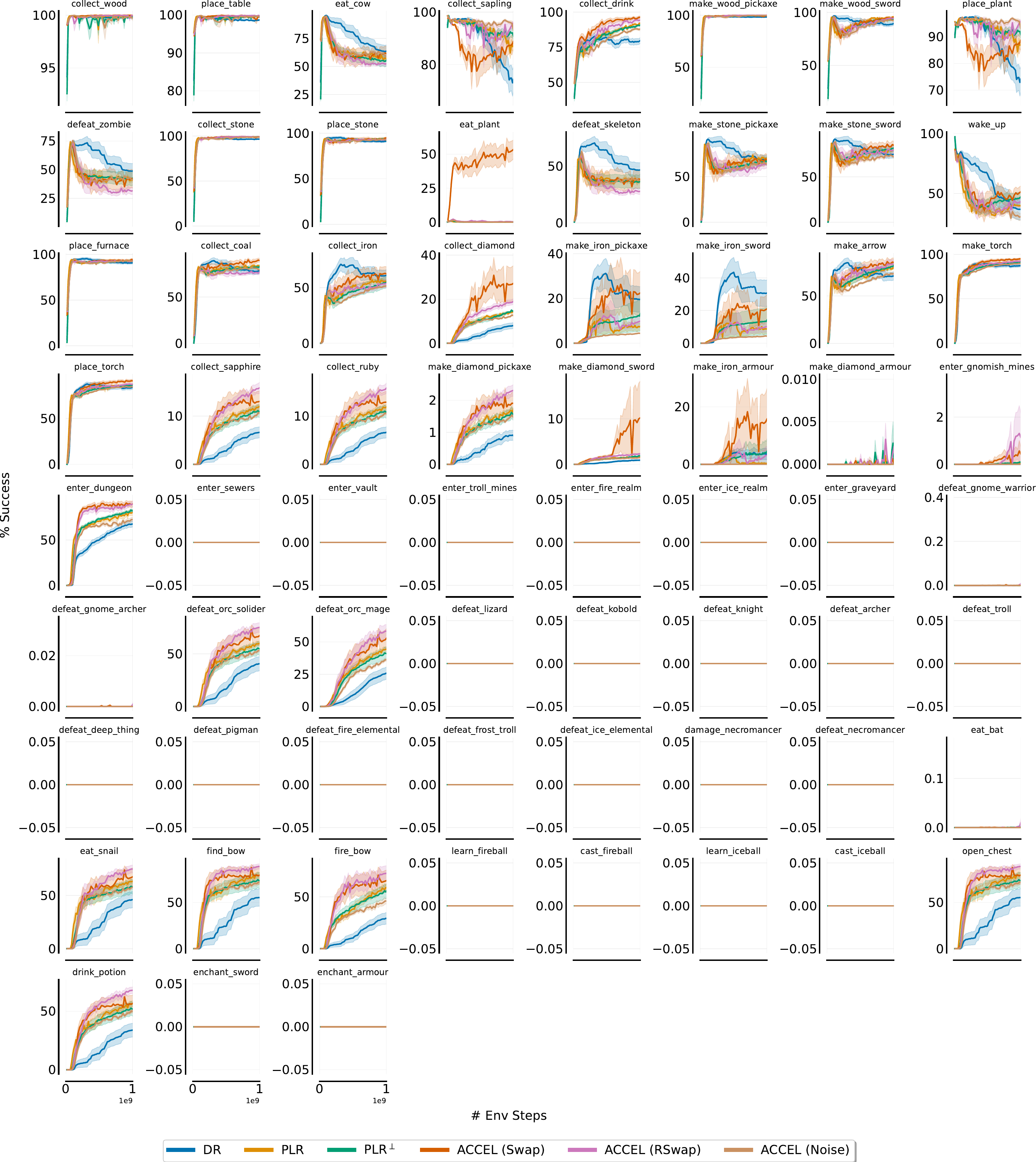}
        \caption{Replay achievement success rates for UED}
        \label{fig:ued:replay_vs_dr:replay}
\end{figure*}
\begin{figure*}
        \includegraphics[width=\linewidth]{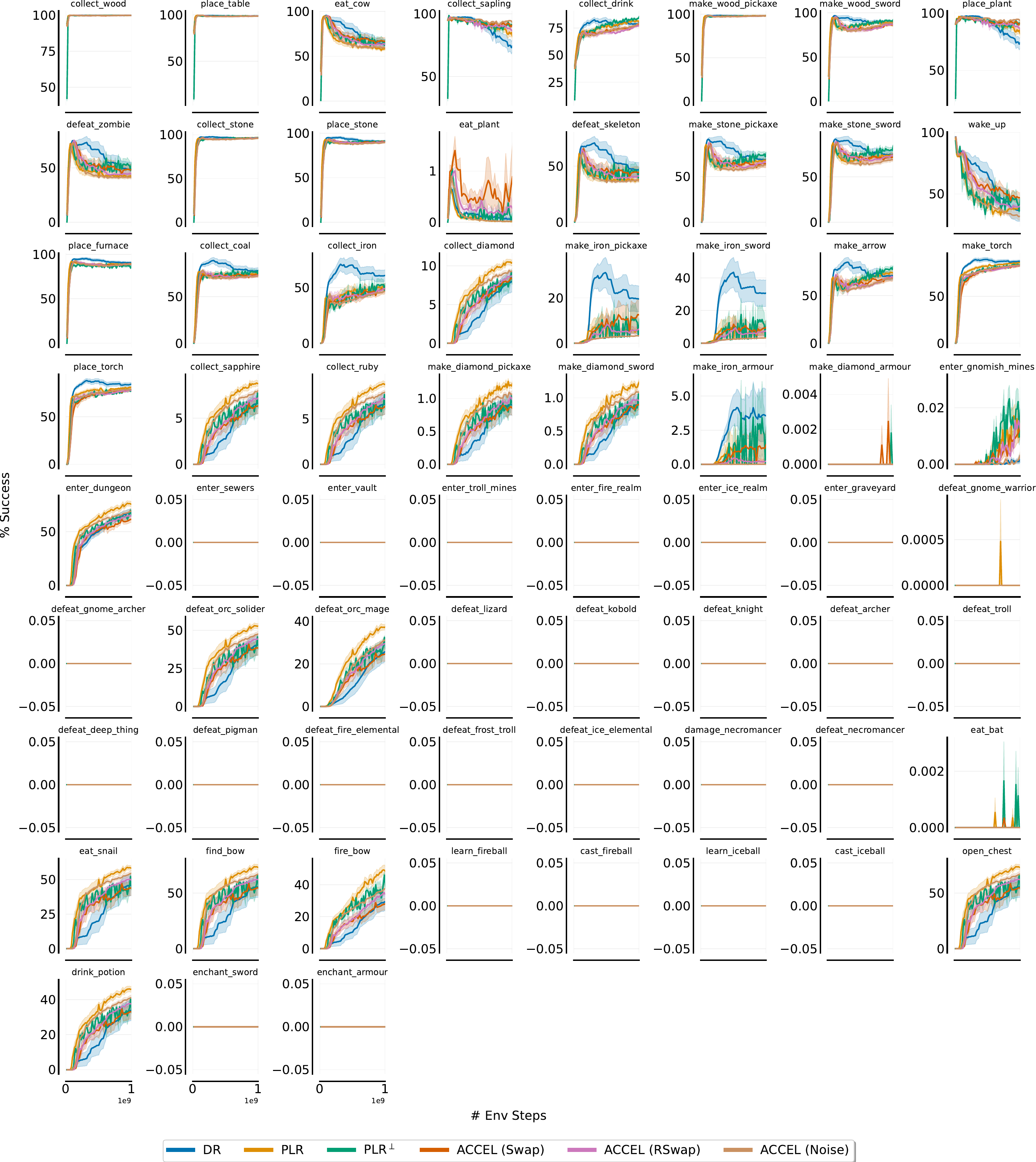}
        \caption{DR achievement success rates for UED}
        \label{fig:ued:replay_vs_dr:dr}
\end{figure*}
\begin{figure*}
        \includegraphics[width=\linewidth]{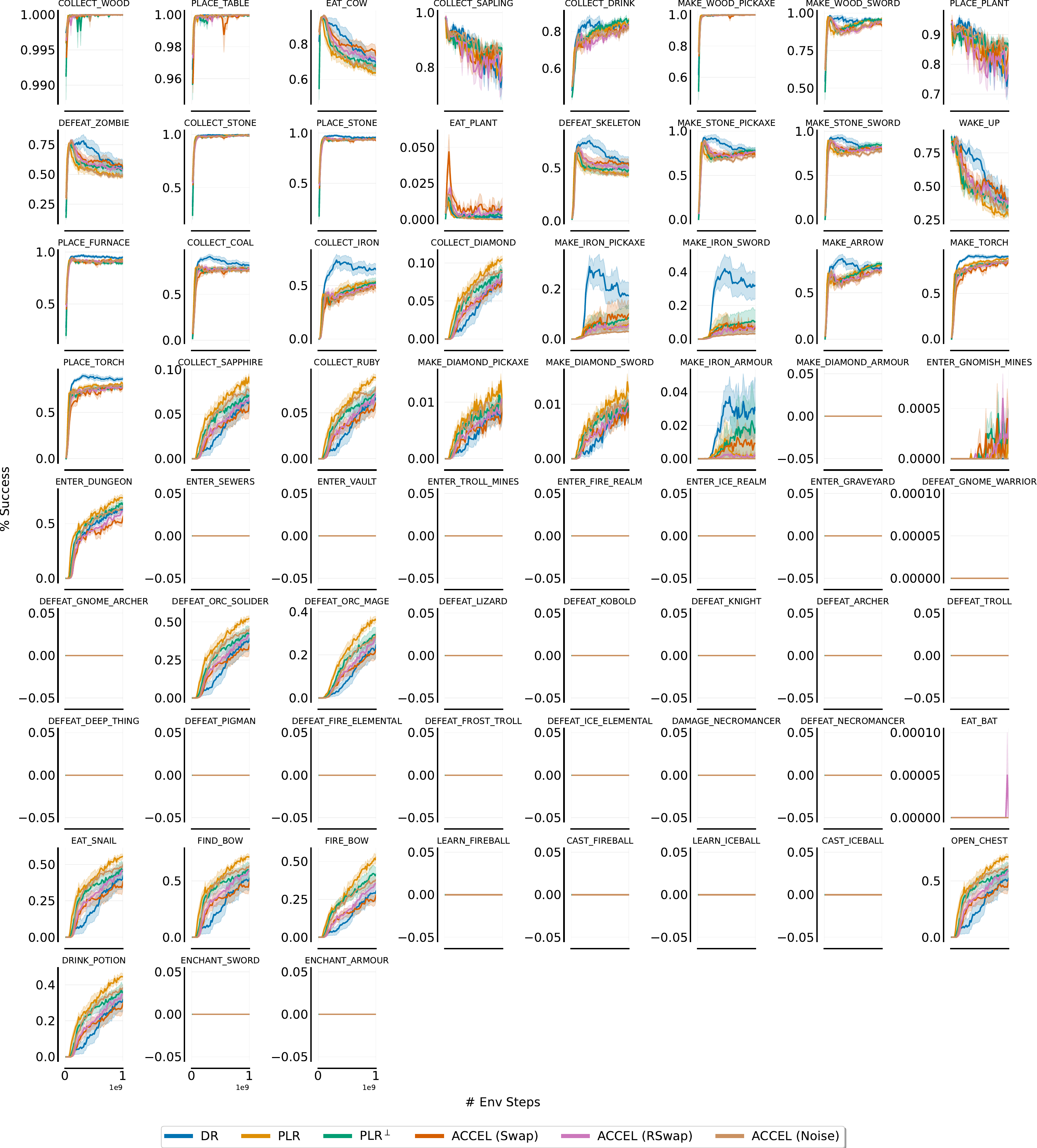}
        \caption{Evaluation achievement success rates for UED}
        \label{fig:ued:eval:all_achievements}
\end{figure*}

\subsection{Hyperparameters}
\cref{table:ued:hyperparams} contains the UED hyperparameters used. We tuned these as follows.
\begin{enumerate}
    \item We used PPO hyperparameters found in an earlier hyperparameter sweep and did a small search over learning rate $\{ 0.0002, 0.0003 \}$.
    \item We then performed a grid search over the following parameters for both PLR variations: temperature $\{ 0.3, 1.0 \}$, replay probability $\{ 0.5, 0.8 \}$, score function $\{ \text{MaxMC}, \text{PVL} \}$, prioritisation $\{ \text{rank}, \text{topk} \}$ and topk's k in $\{32, 256 \}$.
    \item We then used the best parameters for the ACCEL-based methods, and further tuned over the number of mutations in $\{10, 100, 200 \}$. In this case, PLR outperformed robust PLR, so all of our ACCEL methods also perform gradient updates on DR levels, but not on mutated levels.
\end{enumerate}

All ACCEL-methods use exploratory gradient updates.
\begin{table}[H]
    \caption{{UED Hyperparameters. We use a replay rate of 0.8 for $\text{PLR}^\perp$ and a replay rate of 0.5 for PLR and ACCEL.}}
    \label{table:ued:hyperparams}
    \begin{center}
    \scalebox{1.0}{
        \begin{tabular}{lrr}
        \toprule
        \textbf{Parameter}              & DR        & PLR/$\text{PLR}^\perp$/ACCEL\\
        \midrule
        \emph{PPO}                      &\\
        Network Width                   & 512       &\\
        \# Environments                 & 1024      &           \\
        $\gamma$                        & 0.99      &           \\
        $\lambda_{\text{GAE}}$          & 0.9       &           \\
        PPO inner rollout               & 64        &           \\
        PPO outer rollout               & 64        &           \\
        PPO epochs                      & 5         &           \\
        PPO minibatches per epoch       & 2         &           \\
        PPO clip range                  & 0.2       &           \\
        Adam learning rate              & 0.0002    & 0.0003           \\
        Anneal LR                       & yes       &           \\
        Adam $\epsilon$                 & 1e-5      &           \\
        PPO max gradient norm           & 1.0       &           \\
        PPO value clipping              & yes       &           \\
        return normalization            & no        &           \\
        value loss coefficient          & 0.5       &           \\
        student entropy coefficient     & 0.01      &           \\
        \emph{PLR}                      &           &           \\
        Replay rate, $p$                & -         & 0.8 ($\text{PLR}^\perp$), 0.5 (PLR, ACCEL)       \\
        Buffer size, $K$                & -         & 4000      \\
        Scoring function                & -         & MaxMC     \\
        Prioritization                  & -         & Rank      \\
        Temperature, $\beta$            & -         & 1.0       \\
        Staleness coefficient           & -         & 0.3       \\
        \emph{ACCEL}                    &           & -         \\
        Number of Mutations (Swap)      & -         & 10         \\
        Number of Mutations (RSwap)     & -         & 200         \\
        Number of Mutations (Noise)     & -         & 100         \\
        \bottomrule 
        \end{tabular}}
    \end{center}
\end{table}


\end{document}